\documentclass[final]{clv2025}
\usepackage{hyperref}
\usepackage{xcolor}
\definecolor{darkblue}{rgb}{0, 0, 0.5}
\hypersetup{colorlinks=true,citecolor=darkblue, linkcolor=darkblue, urlcolor=darkblue}
\usepackage{longtable}
\usepackage{booktabs}
\usepackage{subcaption}
\usepackage{algorithm}
\usepackage{algpseudocode}
\usepackage{pdflscape}
\usepackage{booktabs}
\usepackage{multirow}
\usepackage{tabularx}
\usepackage{comment}
\usepackage{amsmath}
\usepackage{pgfplots}
\pgfplotsset{compat=1.18}
\usepgfplotslibrary{fillbetween}

\usepackage{tikz}
\usetikzlibrary{positioning, decorations.pathmorphing, shapes, arrows}

\pgfplotsset{vasymptote/.style={
		before end axis/.append code={
			\draw[densely dashed] ({rel axis cs:0,0} -| {axis cs:#1,0})
			-- ({rel axis cs:0,1} -| {axis cs:#1,0});
		}
},
width=6.5cm
}

\newcommand{\modelname}{Psychology-based Unified Dynamic Framework for Curriculum Learning}
\newcommand{\modelabbr}{PUDF}

\usepackage{booktabs}
\usepackage{subcaption}
\usepackage{tabularx}
\usepackage{array}

\begin{document}
\issue{}{}{}
\dochead{Long Paper}

\runningtitle{Psychology-Inspired Curriculum Learning}

\runningauthor{Meng, Zeng, Lalor, Yu}

\pageonefooter{Action editor: Afra Alishahi. Submission received: 13 December 2024; revised version received: 6 November 2025; accepted for publication: 17 November 2025.}

\title{A Psychology-based Unified Dynamic Framework for Curriculum Learning}

\author{Guangyu Meng$^{1}$, Qinkai Zeng$^{2}$, John P. Lalor$^{1,3}$\thanks{Corresponding authors}, Hong Yu$^{4,5,6*}$}

\affilblock{
    \affil{University of Notre Dame, Department of Computer Science and Engineering\\\quad  \email{gmeng@nd.edu}}
    \affil{Nankai University, College of Computer Science\\ \quad \email{qzengnkcs@gmail.com}}
    \affil{University of Notre Dame, Department of IT, Analytics, and Operations\\
    \quad \email{john.lalor@nd.edu}}
    \affil{VA Bedford Healthcare System, Center for Health Optimization and Implementation Research}
    \affil{University of Massachusetts Amherst, Manning College of Information and Computer Sciences}
    \affil{University of Massachusetts Lowell, Miner School of Computer and Information Sciences\\ \quad
    \email{hong\_yu@uml.edu}}
}

\maketitle 

\begin{abstract}

Directly learning from examples of varying difficulty levels is often challenging for both humans and machine learning models. 
A more effective strategy involves exposing learners to examples in a progressive order from easy to difficult. 
Curriculum Learning (CL) has been proposed to implement this strategy in machine learning model training. 
However, two key challenges persist in CL framework design: defining the difficulty of training data and determining the appropriate amount of data to input at each training step.
Drawing inspiration from psychometrics, this paper presents a \modelname~(\modelabbr). 
We quantify the difficulty of training data by applying Item Response Theory (IRT) to responses from Artificial Crowds (AC). 
This theory-driven IRT-AC approach leads to global (i.e., model-independent) and interpretable difficulty values. 
Leveraging IRT, we propose a training strategy, Dynamic Data Selection via Model Ability Estimation (DDS-MAE), to schedule the appropriate amount of data during model training. 
Since our difficulty labeling and model ability estimation are based on a consistent theory, namely IRT, their values are comparable within the same scope, potentially leading to aligned training data selection and faster convergence compared to the other CL methods.
Experimental results demonstrate that fine-tuning pre-trained large language models with \modelabbr{} leads to higher accuracy and faster convergence on a suite of benchmark datasets compared to standard fine-tuning and state-of-the-art CL methods. 
Ablation studies and downstream analyses further validate the impact of \modelabbr{}~for CL.


\end{abstract}

\newcommand{\hl}[2]{{\color{#1}\bfseries [[#2]]}} 
\newcommand{\todo}[1]{\hl{red}{#1}}

\section{Introduction}
\label{sec:introduction}

Curriculum learning (CL) is a machine learning framework that trains models by gradually introducing examples of increasing difficulty~\citep{bengio_curriculum_2009}. 
CL can effectively improve the generalization capacity and convergence rate of various models in a wide range of scenarios, such as computer vision~\citep{soviany2021curriculum,zhang2021flexmatch}, natural language processing~\citep[NLP,][]{zhan2021meta,zhao2021automatic}, robotics \citep{milano2021automated,manela2022curriculum}, and medical applications~\citep{liu2022competence,burduja2021unsupervised}. 
In NLP in particular, CL has been shown to improve performance in applications such as machine translation~\citep{zhan2021meta,mohiuddin-etal-2022-data}, sentiment analysis~\citep{cirik2016visualizing,tsvetkov2016learning}, and natural language understanding~\citep{xu2020curriculum}. 
A key benefit of CL is its ability to guide the training process towards optimal regions in the parameter space, thus reducing time spent on noisy and difficult samples in early training stages~\citep{wang2021survey}.
Recent work applying CL to pre-trained large language models (LLMs) has shown it to be effective for fine-tuning~\citep{lee2022efficient, nagatsuka2021pre, platanios_competence-based_2019,xu2020curriculum}. 
While these CL methods show promise in improving accuracy, they also introduce increased complexity and longer training times, which can offset some of the benefits and hinder widespread adoption.

In this work, 
we propose a novel CL framework, \modelname{}~(\modelabbr{}). 
With \modelabbr{}, we introduce novel approaches to two key CL components \citep{wang2021survey}: difficulty measurement (DM) and training scheduling (TS).
Specifically, we propose Item Response Theory-based Artificial Crowds (IRT-AC) as DM and Dynamic Data Selection via Model Ability Estimation (DDS-MAE) as TS.
Both components are based on Item Response Theory \citep[IRT,][]{baker_item_2004,de2013theory}, a well-established methodology in psychometrics for test construction and subject evaluation. 
IRT assumes that a latent difficulty value for items (which we refer to as ``examples'') can be estimated from responses to the examples from a population of test-takers (``subjects'').
Each subject is assumed to have a latent ability value corresponding to their proficiency on a task, as evaluated by performance on the items.
We use a one-parameter IRT model, which assumes that all examples are equally discriminative and only vary in their difficulty. 
IRT-AC estimates latent difficulty parameters for each example from an artificial crowd of NLP models. 
When fine-tuning a new model, DDS-MAE uses the IRT-AC output to estimate a latent ability score for the model at each training epoch,  
then dynamically selects training data based on the model's current ability.
This approach is similar to existing CL methods that use reinforcement learning, but is more efficient as it does not require a carefully designed reward function. 

A key benefit of using IRT is that example difficulty is global and model-independent. 
Other methods for DM, such as loss, are model- or training epoch-dependent, which means that difficulty can vary between models and training runs; IRT estimated difficulty estimates are fixed a priori.
As a result, using IRT-AC as the DM allows for estimating example difficulty offline for efficient use during training (TS).
What's more, in a one-parameter IRT model, there is an interpretable relationship between example difficulty and model ability.
Specifically, an example's difficulty can be interpreted as the model ability value needed to have a 50\% chance of labeling that example correctly. 
This provides a theoretically grounded and interpretable way to relate example difficulty to model ability.

Traditionally, fitting IRT models required extensive human-annotated data. 
However, recent work has shown that IRT models can be fit using machine-generated data instead of human-generated data \citep{lalor_learning_2019}. 
Building on this, we propose the use of artificial crowds (AC) composed of multiple high-performing LLMs to obtain predicted results of the training data used to estimate an IRT model. 
Generating responses from multiple LLMs for the AC can be done offline; responses can be reused to reduce the computational cost. 
Furthermore, traditional IRT models do not scale well to large numbers of subjects and examples. 
Therefore, we leverage a variational inference (VI) method \citep{hoffman2013stochastic, jordan1999introduction} to fit a large-scale IRT model. 
VI estimates a variational distribution that approximates the true posterior.
Learning involves minimizing the KL Divergence between the variational distribution and the true distribution via batched stochastic optimization, allowing for efficient estimation and scalability to larger datasets. 

To test its effectiveness, we evaluate \modelabbr{} with a comprehensive suite of benchmarking datasets and existing CL methods.
We find that \modelabbr{} improves training efficiency and predictive performance across our benchmarking models and datasets.
For example, on the AG News dataset, consisting of almost 1 million training examples, fine-tuning Llama3.1-8B with \modelabbr{} leads to a 4.13\% relative improvement in accuracy and a 69.68\% relative improvement in training time compared to traditional fine-tuning.
\modelabbr{} also improves over a robust suite of other curriculum learning methods.
For example, fine-tuning Llama3.1-8B on AG News with \modelabbr{} leads to relative improvements of 0.42\% and 75.48\% in accuracy and training time, respectively, compared to a state-of-the-art (SOTA) reinforcement learning CL framework~\citep{senguptagood}.

Our contributions are as follows: (1) We propose \modelabbr{}, an innovative approach to implementing an effective CL strategy for fine-tuning LLMs; (2) Compared to existing CL methods, our DM (IRT-AC) and TS (DDS-MAE) automatically define model-independent data difficulty and achieve dynamic data selection without significant time penalties; (3) Experimental results demonstrate \modelabbr{}'s faster convergence and higher accuracy for fine-tuning LLMs, particularly with difficult and large-scale datasets, highlighting its scalability and efficiency.
Overall, our results demonstrate that using \modelabbr{} leads to more efficient training and better performance. What's more, \modelabbr{} allows for scaling curriculum learning to much larger datasets, as demonstrated by our results on AG News, which contains almost 1 million training examples.

In prior work~\cite{lalor-yu-2020-dynamic}, we proposed Dynamic Data Selection for Curriculum Learning via Ability Estimation (DDaCLAE), a preliminary framework for CL using learned difficulty and ability parameters.
We benchmarked DDaCLAE against a single curriculum learning baseline using BERT and LSTM models, and demonstrated the potential of learned difficulties over heuristics such as sentence length to validate our approach.
To the best of our knowledge, this was the first work to learn model competence during training that is directly comparable to the difficulty of the examples.
In this work, we significantly enhance the scope and depth of our preliminary results with DDaCLAE.
We have generalized and modularized the previous model to facilitate its adaptation to emerging curriculum learning paradigms~\citep{wang2021survey} and its extension for future research.
Specifically, in this manuscript we extend and enhance \citet{lalor-yu-2020-dynamic} in the following ways: 

\begin{enumerate}
    \item \textbf{Enhanced Difficulty Assessment in IRT-AC:} For IRT-AC, we now employ 13 distinct pre-trained models to learn example difficulties (\S\ref{ssec:acModels}). This contrasts with the previous methodology, which utilized only a single model type (LSTM or BERT) to estimate difficulty under varying training data noise levels. IRT-AC thus offers a more robust and comprehensive evaluation, capable of encompassing a broader range of conditions. 

    \item \textbf{Integration of SOTA Language Models:} To evaluate \modelabbr{}, we have replaced BERT \cite{kenton2019bert} and LSTM \cite{hochreiter_long_1997} with larger and more advanced architectures (\S\ref{ssec:benchmarkModels}). Specifically, here we conduct our experiments using DeBERTaV3 \cite{he2020deberta}, GPT-2 \cite{radford2019language}, Llama3.1-8B \cite{grattafiori2024llama}, and Qwen2.5-7B \cite{yang2024qwen2}. This allows for a more rigorous validation of our approach using recent large language models.

    \item \textbf{Expanded Benchmarking Comparisons:} We extend our benchmarking experiments from our previous work, which were limited to comparisons with heuristic-based CL methods.
    Here, we add several advanced CL benchmark methods (\S\ref{ssec:benchmarkMethods}), including techniques based on reinforcement learning teachers, transfer teachers, and self-paced learning.

    \item \textbf{Inclusion of New Diverse Datasets:} We have augmented our benchmarking suite with two new datasets (\S\ref{ssec:data_description}): AG News, a multi-class classification dataset comprising 14 classes and almost 1 million training examples \cite{zhang2015character}, and MedQA-UMLS, a challenging medical question-answering dataset \cite{jin2020disease}. These additions facilitate a more thorough evaluation of \modelabbr{}'s performance across diverse and complex tasks. 
    
    \item \textbf{Extension to Text Generation}
    We include an example of how \modelabbr{} can be used for generative tasks with experiments on the GSM8K dataset (\S \ref{ssec:gsm8k}), a text generation math problem dataset used for LLM benchmarking.
    Our results indicate that \modelabbr{} improves performance and runtime for GSM8K over the existing benchmark methods.

    \item \textbf{Robust Downstream Analyses: } We have also added a robust downstream analysis of \modelabbr{} that extends beyond improvements to performance and training time to provide a more detailed examination of \modelabbr{}'s characteristics and efficacy. 
    Specifically, we have conducted the following new analyses: A \textit{theoretical analysis} (\S\ref{sssec:timeCompl}) and a \textit{per-component runtime analysis} (\S\ref{sssec:runtime}) of \modelabbr{} to reinforce and validate our claims regarding its efficiency; a systematic \textit{ablation study} (\S\ref{ssec:ablation}) to isolate and quantify the contribution of each component to \modelabbr{}'s overall performance; and a \textit{convergence analysis} (\S\ref{sssec:convergence}) to highlight its advantages in convergence behavior, data efficiency, information utilization, and stability, and provide detailed insights into \modelabbr{}'s faster and more efficient training dynamics compared to baseline models.
    \item \textbf{Analysis of Learned Difficulties:}
    We have also added a robust analysis of the difficulty values learned from IRT-AC across three dimensions to further validate the approach: the distribution of data difficulty (\S\ref{sssec:diffdist}), artificial crowd prediction accuracy across difficulty bins (\S\ref{sssec:accDiffBins}), and crowd models' confidence scores relative to estimated difficulty (\S\ref{sssec:confGenDiff}). 
\end{enumerate}

Overall, our enhanced benchmarking results with newer models and datasets validate and extend our initial results~\citep{lalor-yu-2020-dynamic} and demonstrate the applicability of \modelabbr{} to recent LLMs and large-scale datasets. Our newly-added downstream analyses provide further insights into the mechanisms by which \modelabbr{} improves training performance and efficiency. Our dataset-related results concerning example difficulty can aid researchers in better understanding the intricacies of specific dataset examples.

\noindent \textbf{Significance to the Research Community}: This manuscript situates \modelabbr{} within the ongoing curriculum learning discourse by aligning it with the existing understanding of CL frameworks \citep{wang2021survey} and benchmarking it against SOTA methods. The additional analyses and broader benchmarking provide robust evidence of \modelabbr{}'s utility, offering new insights into instance-level metrics like difficulty and confidence, which are becoming increasingly relevant in NLP research~\citep{swayamdipta2020dataset,rodriguez2021evaluation,cook2025no}. By addressing scalability and interpretability, this work serves as a foundation for future innovations in dynamic curriculum learning for NLP.
To facilitate future work, we make our code available\footnote{\url{https://github.com/nd-ball/cl-irt/}} and also release the data collected from our artificial crowds.\footnote{\url{https://huggingface.co/datasets/nd-ball/response-patterns}}

The rest of this paper is organized as follows.
Section \ref{sec:relatedwork} reviews the related work in curriculum learning and Section \ref{sec:background} presents background information on Item Response Theory.
In Section \ref{sec:methodology}, we present \modelabbr{} and describe its key components.
Section \ref{sec:experiments} describes our main experiments, results, and detailed analyses.
In Section \ref{sec:exploration}, we discuss our analyses of example difficulties learned from IRT-AC.
Section \ref{sec:limitations} discusses the limitations of this work and Section \ref{sec:conclusion} concludes.

\section{Related Work}
\label{sec:relatedwork}

\subsection{Curriculum Learning}


CL methods implement model training strategies by progressively moving from easier to harder training data. 
The concept of training neural networks in a progressively easy-to-difficult manner can be traced back to the work of \citet{elman1993learning}. 
Building on these foundations, CL was formally proposed by \citet{bengio_curriculum_2009}; there, the authors evaluated pre-designed CL methods on toy datasets with heuristic measures of difficulty.
CL has since been studied in machine learning broadly \citep[e.g., ][]{soviany2021curriculum,zhang2021flexmatch,milano2021automated,manela2022curriculum,liu2022competence,burduja2021unsupervised} and NLP specifically \citep[e.g., ][]{zhan2021meta,mohiuddin-etal-2022-data,cirik2016visualizing,tsvetkov2016learning,xu2020curriculum} and has been shown to improve learning across a variety of tasks and domains. 
There has also been a stream of research investigating the theory behind CL \citep{weinshall2018curriculum,hacohen2019power}, particularly with regard to defining an ideal curriculum. 
CL theoretically leads to a steeper optimization landscape (i.e., faster learning) than standard training while keeping the same global minimum of the task.
These theoretical results also highlight a key distinction between CL and similar guided training methods such as self-paced learning \citep{kumar2010self}, hard example mining \citep{shrivastava_training_2016}, and boosting \citep{freund1997decision}: namely that CL considers difficulty with respect to the final hypothesis space (i.e., a model trained on the full dataset), while the other methods consider ranking examples according to how difficult the current model determines them to be \citep{weinshall2018curriculum,hacohen2019power}. 
Our proposed \modelabbr~bridges a gap between these methods by probing model ability at the current point in training and using this ability to identify appropriate training examples in terms of difficulty that is independent of a specific model or training epoch.

\subsection{A General CL Framework}

In a recent survey, \citet{wang2021survey} categorized CL methods in the literature based on two key components: a \textbf{difficulty measurer (DM)}, which provides a score indicating the relative easiness of each data example, and a \textbf{training scheduler (TS)}, which decides the sequence of data subsets to use throughout the training process. 
The general workflow for CL involves first ordering all training examples from easiest to hardest according to the DM. 
Subsequently, at each training epoch, the TS selects the appropriate subset of training data and presents it to the model for learning.
CL methods can be categorized into two types based on how the DM and TS are implemented: Predefined and Automated~\citep{wang2021survey}.
In Predefined CL, both DM and TS are designed using prior human knowledge and without data-driven methods. 
In Automatic CL, one or both of the DM and TS are learned by data-driven models or algorithms. 
In their review, \citet{wang2021survey} identified three variations of Automatic CL: self-paced learning CL, transfer teacher CL, and reinforcement learning (RL) CL. 
We summarize each CL type in Table \ref{tab:cl_methods} and discuss the characteristics of each approach below.

\begin{table}[t]
\centering
\footnotesize
\begin{tabular}{p{2.55cm}p{1.5cm}p{1.5cm}cp{4.25cm}}
\toprule
\textbf{CL Category}  & \textbf{Difficulty\newline Measurer} & \textbf{Training\newline Scheduler} & \textbf{Complexity}  & \textbf{References}\\ \midrule
Predefined & Predefined & Predefined & Low & \citet{bengio_curriculum_2009} \newline\citet{platanios_competence-based_2019}\\
Self-paced learning & Automatic & Predefined & Low &\citet{kumar2010self} \newline\citet{ouyang2023unsupervised}\\
Transfer teacher & Automatic & Predefined & High & \citet{xu2020curriculum}\newline \citet{maharana2022curriculum}\\
RL teacher  & Automatic & Automatic & High & \citet{zhao2020reinforced}\newline \citet{kumar2019reinforcement}\\
\modelabbr{} & Automatic & Automatic & Low & This work\\ \bottomrule
\end{tabular}
\caption{Summary of curriculum learning methods in the literature.}
\label{tab:cl_methods}
\end{table}

\textit{Predefined CL} relies on heuristics (e.g., sentence length or word rarity) for DM and a predetermined scheduling function (e.g., a linear function or root function) for TS based on task-specific data characteristics \citep{platanios_competence-based_2019,spitkovsky2010baby,wei2016stc,tsvetkov2016learning}. 
Examples of difficult text for the DM include longer sentences, the presence of rare words~\citep{platanios_competence-based_2019}, the number of coordinating conjunctions \citep[e.g., ``and'', ``or'',][]{kocmi2017curriculum}, and the number of phrases \citep[e.g., prepositional phrases,][]{tsvetkov2016learning}. 
Common training schedulers include linear and root functions \citep{platanios_competence-based_2019}, which increase the number of training samples at a linear or exponential pace. 
While simple and often effective, finding the optimal combination of DM and TS for specific tasks and datasets often requires expert domain knowledge. 
Moreover, examples that are easy for humans may not be easy for models due to different decision boundaries \citep{yuan2019adversarial}.

\textit{Self-paced learning CL} (SPL) allows the model itself to be the DM based on some model-dependent metric \citep{kumar2010self,jiang2015self,wan-etal-2020-self,mohiuddin-etal-2022-data,ouyang2023unsupervised}. 
For example, prior work has used model training loss~\citep{kumar2010self} and pseudo-label predictions~\citep{ouyang2023unsupervised} as inputs to the DM to measure learning difficulty. 
SPL is more automatic and more aligned with the model's learning process; however, early training may incur high uncertainty when the model is not yet sufficiently trained. 
Moreover, SPL still uses a predetermined function as TS; model competence is not typically considered.
Instead, it is assumed that competence improves monotonically as more difficult examples are added.

\textit{Transfer teacher CL} \citep{weinshall2018curriculum,xu2020curriculum,maharana2022curriculum,hacohen2019power} employs a pre-trained, ``stronger'' model as a teacher to be the DM according to its accuracy. 
For instance, prior work has used RoBERTa-{large} \citep{liu2019roberta} as the teacher model; its output probabilities for training examples were used as difficulty estimates~\citep{maharana2022curriculum}. 
Similarly, a ``cross-review'' strategy was proposed where a teacher model with the same structure as the student model labels difficulty~\citep{xu2020curriculum}.
TS in this transfer teacher-based CL approach leverages predefined functions, such as an annealing method \citep{xu2020curriculum} or an adaptive function \citep{maharana2022curriculum}.
However, such methods are costly due to the additional fine-tuning and still rely on a predefined TS.


\textit{RL teacher CL} \citep{zhao2020reinforced,kumar2019reinforcement} methods adopt RL models as the teacher to perform TS according to the feedback from the model. 
Examples of RL teacher methods include a multi-armed bandit RL method \citep{graves2017automated}, a Q-Learning strategy \citep{zhao2020reinforced}, and a deterministic Actor-Critic RL model \citep{kumar2019reinforcement}. 
This dynamic approximates the learning process in human education, where the teacher and student improve together through interactions: the student makes progress based on the tailored learning materials selected by the teacher, while the teacher adjusts teaching strategy based on student performance. 
However, the RL model method is costly; we not only need to train the original model but also fine-tune the RL model based on a carefully-designed reward function during training.

\section{Background: Item Response Theory}
\label{sec:background}

In this section, we first introduce IRT, in particular the one-parameter logistic (1PL) model \citep{rasch_studies_1960,baker_item_2004}, and describe learning IRT models for machine-learning scale datasets with variational inference methods. 
As discussed in the introduction, evaluating the difficulty of data examples while considering a model's capabilities allows for an interpretable comparison between data difficulty and model ability. 
This aligns with an intuitive understanding of human learning, namely that a good student answering a question correctly does not necessarily imply that the question is easy. 
To achieve this mutual evaluation, we employ IRT methods, which learn latent parameters of dataset examples (called ``items'' in the IRT literature) and latent ability parameters of individual ``subjects.''
We refer to ``items'' as ``examples'' and ``subjects'' as ``models,'' respectively, for clarity and consistency with the curriculum learning literature.

For a model $j$ and an example $i$, the probability that $j$ labels $i$ correctly ($z_{ij}=1$) is a function of the latent parameters of $j$ and $i$.
The one-parameter logistic (1PL) model, or Rasch model, assumes that the probability of labeling an example correctly is a function of a single latent difficulty parameter of the example, $b_i$, and a latent ability parameter of the model, $\theta_j$ \citep{rasch_studies_1960,baker_item_2004}: 

\begin{equation} 
	\label{eq:irt}
	p(z_{ij} = 1 \vert \theta_j, b_i) = \frac{1}{1 + e^{-(\theta_j - b_i)}}
\end{equation} 



When plotted, $p(z_{ij} = 1 \vert \theta_j, b_i)$ is known as an item characteristic curve (ICC). 
The ICC is a visual representation of the example with regard to how a subject is expected to perform (Figure \ref{fig:irtexamplegood}).
With a 1PL model, there is an intuitive relationship between difficulty and ability.
An example's difficulty value $b_i$ can be thought of as the ability value for a model that has a 50\% chance of labeling that example correctly.
Put another way, model $j$ has a 50\% chance of labeling example $i$ correctly when $j$'s ability is equal to $i$'s difficulty ($\theta_j = b_i$, see Figure \ref{fig:irtexamplegood}).

Fitting an IRT model requires a set of $I$ examples $\{i_0, i_1, \dots, i_I\}$, a set of $J$ models $\{j_0, j_1, \dots, j_J\}$, and the binary graded responses of the models to each of the examples, $Z = \{\forall_{i \in I} \forall_{j \in J}: z_{ij}\}$. 
The log likelihood of a dataset of response patterns $Z$ given the parameters $\Theta$ and $B$ is:

\begin{align}
	\log \mathcal{L} &= \sum_{j=1}^J \sum_{i=1}^I \log p(Z_{ij}=z_{ij} \vert \theta_j, b_i)
\end{align}

where $z_{ij} = 1$ if model $j$ answers example $i$ correctly and $z_{ij} = 0$ otherwise.

\begin{figure}[hbt]
	\centering
	\begin{subfigure}{0.45\textwidth}
		\centering 
		\begin{tikzpicture}
			\begin{axis}[xlabel=$\theta_j$,ylabel={$p(z=1\vert \theta_j,b_i)$}, ymin=0, ymax = 1, xmin=-4, xmax=4, vasymptote=0]
				\addplot[draw=blue, fill=gray!0] {((1.0) / (1 + exp(-1*(x ))))};
			\end{axis}
      \node[above] at (current bounding box.north) {Example where $b_i=0$};
		\end{tikzpicture}
		\caption{\label{fig:irtA}}
	\end{subfigure}
    ~~~~~~~
	\begin{subfigure}{0.45\textwidth}
		\centering 
		\begin{tikzpicture}
			\begin{axis}[xlabel=$\theta_j$,ylabel={$p(z=1\vert \theta_j,b_i)$}, ymin=0, ymax = 1, xmin=-4, xmax=4, vasymptote=2]
				\addplot[draw=blue, fill=gray!0] {((1.0) / (1 + exp(-1*(x - 2))))};
			\end{axis}
      \node[above] at (current bounding box.north) {Example where $b_i=2$};
		\end{tikzpicture}
		\caption{\label{fig:irtB}}
	\end{subfigure}
	
	\caption{Plot of $p(z_{ij} = 1 | \theta_j, b_i)$ as a function of $\theta_j$ for two examples: (\ref{fig:irtA}) an example with difficulty $b_i=0$, and (\ref{fig:irtB}) a more difficult example ($b_i=2$). Models with ability $\theta_j > b_i$ (right of dashed line) have greater than 50\% chance of labeling the example correctly.}
	\label{fig:irtexamplegood}
\end{figure}

For a given dataset of response patterns $Z$, item parameters are traditionally estimated using a marginal maximum likelihood expectation-maximization algorithm \citep{bock_marginal_1981}, where the latent ability parameters ($\theta$) are assumed to be random effects and are integrated out to define the marginal probability.
Once item parameters are estimated, model ability is scored via maximum likelihood estimation.
However, traditional IRT model fitting does not scale to large datasets. 
Therefore, prior work proposed the use of variational inference \citep[VI, ][]{natesan_bayesian_2016,jordan1999introduction} to estimate latent IRT parameters. 
VI-IRT approximates the joint posterior distribution $p(\Theta,B|Z)$ by a variational distribution $q(\Theta, B)$:

\begin{align} 
	q(\Theta, B) &=  \prod_{j=1}^J \pi^\theta_j(\theta_j) \prod_{i=1}^I \pi^b_i(b_i)
\end{align} 
where $\pi^\theta_j()$ and $\pi^b_i()$ denote Gaussian densities for different parameters.
Parameter means and variances are determined by minimizing the KL-Divergence between $q(\Theta,B)$ and $p(\Theta,B \vert Z)$: 

\begin{equation}
\arg \min_{q} D_{\text{KL}}(q(\Theta, B) \vert \vert p(\Theta, B \vert Z))
\end{equation}

Optimization is typically performed via batched stochastic gradient descent, which scales to larger datasets and can leverage GPUs for faster training \citep{lalor2023py}.
In selecting priors for VI-IRT, we follow the results of prior work and use hierarchical priors \citep{natesan_bayesian_2016,lalor_learning_2019}.
The hierarchical model assumes that ability and difficulty means are sampled from a vague Gaussian prior (Equation \ref{eqn:mean}), and ability and difficulty variances are sampled from an inverse Gamma distribution (Equation \ref{eqn:variance}):

\begin{align}
	\theta_j\ |\ m_\theta, u_\theta &\sim N(m_{\theta}, u^{-1}_{\theta}) \\
	b_i\ |\ m_b, u_b &\sim N(m_b, u^{-1}_b) \\
	m_{\theta}, m_{b} &\sim N(0, 10^6) \label{eqn:mean} \\
	u_{\theta}, u_b &\sim \Gamma(1, 1) \label{eqn:variance}
\end{align}

\section{Methodology}
\label{sec:methodology}

In this section, we first present the workflow for \modelabbr{} in Section \ref{ssec:PUDF}. 
We then discuss key insights and challenges for the DM and TS components of \modelabbr{} in Sections \ref{ssec:IRT-AC} and \ref{ssec:DDS-MAE}, respectively. 
We discuss model training and provide the pseudo-algorithm for DDS-MAE in Section \ref{ssec:training} and perform theoretical time complexity analysis in Section \ref{sssec:timeCompl}. 
For clarity, notations and their descriptions are listed in Table \ref{tab:notations}.

\begin{table}[!htbp]
\centering
\begin{tabular}{c l}
\hline
\textbf{Notation} & \textbf{Description} \\ \hline
$J$ & Artificial crowds\\
$M$ & Model \\
$\text{Val}(X,Y)$ & Validation set \\ 
$\text{Train}(X,Y)$ & Training set \\ 
$b$ & Difficulty parameter of the training data \\ 
$\theta$ & Ability parameter of the model \\ 
$e$ & Training epoch \\ 
$Z$ & Response patterns of the models \\ 
$\mathcal{L}$ & Likelihood \\ 
$q(\Theta, B)$ & Variational distribution \\ 
$\pi^\theta_j$ & Gaussian density for ability parameters \\ 
$\pi^b_i$ & Gaussian density for difficulty parameters \\ 
$D_{\text{KL}}$ & KL-Divergence \\ 
$m_\theta, m_b$ & Means of ability and difficulty parameters \\ 
$u_\theta, u_b$ & Variances of ability and difficulty parameters \\ \hline
\end{tabular}
\caption{The descriptions of the notations in our model.}
\label{tab:notations}
\end{table}

\subsection{\modelabbr{} Workflow}
\label{ssec:PUDF}

\begin{figure*}[ht]
    \centering
    \includegraphics[width=\textwidth] {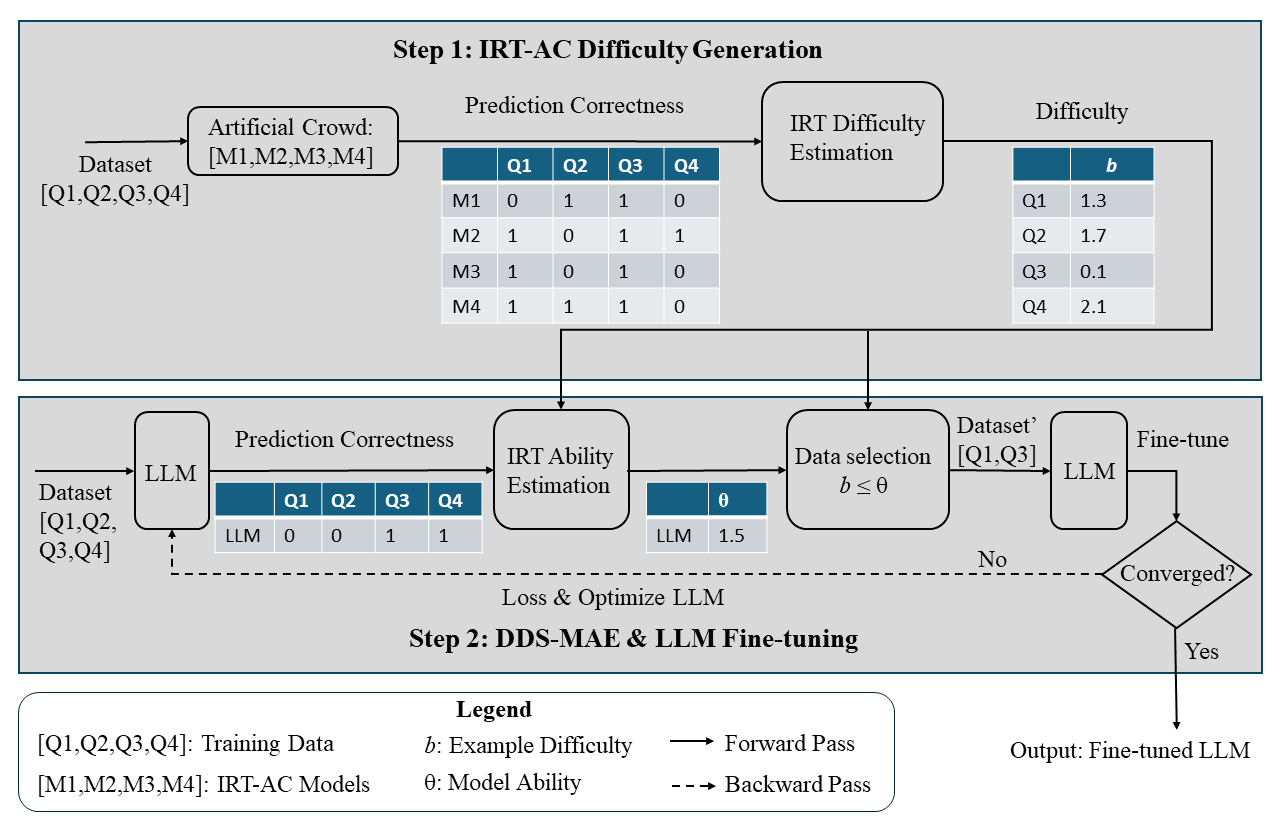}
    \caption{Workflow of \modelabbr{}. The process consists of two main steps: 1) IRT-AC for the DM, 2) DDS-MAE and LLM Fine-tuning for the TS.}
    \label{fig:pudf_workflow}
\end{figure*}

We first introduce the \modelabbr{} workflow, as illustrated in Figure \ref{fig:pudf_workflow}. 
Consistent with \citet{wang2021survey}, \modelabbr{} includes two steps: 1) IRT-AC for the DM and 2) DDS-MAE for the TS. 
To estimate difficulty, the training dataset is the input for the artificial crowd (AC). 
The AC consists of multiple models that generate predictions for each training data example.
These predictions are evaluated against the true labels and converted to binary outcomes (0 or 1) for each model-example pair. 
The AC predictions are then used to estimate example difficulty with an IRT 1PL model, which generates difficulty scores for each training data example, where higher scores indicate greater difficulty.
The TS evaluates the current LLM's ability based on LLM-generated predictions for the training dataset, which are converted to correct/incorrect responses. 
The LLM's responses are used in conjunction with the training data examples' difficulty from the DM to estimate the LLM's ability. 
Then, a subset of the data where difficulty ($b$) is less than or equal to ability ($\theta$), is selected: $b \leq \theta$. 
This workflow combines the strengths of IRT-AC for difficulty estimation, IRT for ability assessment, and dynamic data selection for efficient fine-tuning, resulting in a comprehensive approach to improving LLM performance via CL.

\subsection{IRT-AC}
\label{ssec:IRT-AC}

A bottleneck of using IRT methods on machine learning datasets is the fact that each human subject would have to label all (or most) of the examples in the dataset in order to have enough response patterns to estimate the latent parameters. 
Gathering enough labels for each example to fit an IRT model would be prohibitively expensive for human annotators and would require significant effort to ensure annotation quality.
Therefore, we use artificial crowds \citep{lalor_learning_2019} to generate our response patterns in our IRT-AC module.
IRT-AC consists of two parts: training artificial crowd models to generate responses and using IRT to learn the difficulty of examples. 

Our prior work \citep{lalor-yu-2020-dynamic} used a single neural network architecture (either LSTM or BERT) with training data modifications (training sub-sampling and label flipping) to construct an artificial crowd where performance across models in the crowd was varied (Figure \ref{fig:artificialcrowd} in Appendix \ref{ssec:acOld}).
Because of the single-model design, this yielded monotonous variations based only on the training data manipulations.
Here, we leverage multiple LLMs as artificial crowd sources to improve performance. 
Specifically, our proposed AC incorporates a range of advanced pre-trained LLMs, including encoder-based, decoder-based, and encoder-decoder-based transformer architectures \citep{vaswani2017attention, kenton2019bert, brown2020language}.
This variety in artificial crowd models can increase the diversity of predicted results on our datasets while maintaining high confidence in their outputs \citep{bai2022exploiting}. 

This approach allows us to leverage the predictive performance of LLMs, potentially leading to more robust and diverse difficulty assessments for the IRT-AC method.
To further enhance difficulty diversity and increase the credibility of the evaluated difficulty, we perform fine-tuning on the AC LLMs using the validation dataset for 1, 3, 5, and 10 epochs and include these fine-tuned models in the crowd. 
We then use these fine-tuned AC LLMs to predict labels for the training dataset, thereby obtaining the response patterns for difficulty estimation via IRT model fitting.
Specifically, we fit the IRT model using variational inference (VI) \citep{natesan_bayesian_2016,lalor_learning_2019,DBLP:conf/edm/WuDDPG20} in order to account for the large scale of machine learning datasets. 
IRT-AC can incur a high cost, especially for large, complex models. 
However, that cost is a one-time cost, since the response patterns can be stored for future use. 
If new IRT difficulty estimates are needed, for example, when new models are added to the AC, then only those new models need to be fine-tuned. 
Once those response patterns are added, re-running IRT with VI is relatively low-cost compared to the fine-tuning of the AC models. 
In Section \ref{sssec:compTimeIRT}, we analyze the relative impact of each component on IRT difficulty estimation.

\subsection{Dynamic Data Selection via Model Ability Estimation}
\label{ssec:DDS-MAE}

For the TS component in \modelabbr{}, we propose Dynamic Data Selection via Model Ability Estimation (DDS-MAE). 
DDS-MAE trains the LLM with examples where difficulty is less than or equal to the model's ability. 
The estimated ability of the model at a given epoch $e$, $\hat{\theta}_e$, is on the same scale as the difficulty parameters of the data.
This establishes a principled approach for selecting data at any given training epoch, namely those examples where $b_i \leq \hat{\theta}_e$.
This results in a sample of training data for which the model has at least a 50\% probability of labeling the example correctly.
All that is required is a single forward pass of the model on the training data to generate a response pattern (Equation \ref{eqn:RP}).
When example difficulties are known, model ability is estimated by maximizing the likelihood of the data given the response patterns and the example difficulties (Equation \ref{eqn:MML}).
Estimation is typically done via maximum-likelihood estimation using an existing estimation function \citep[e.g., the Nelder-Mead solver, ][]{lagarias1998convergence}:

\begin{align}
	Z_j &= \forall_{y \in Y} \mathbf{I}[y_i = \hat{y_i}] \label{eqn:RP} \\ 
	\hat{\theta}_e &= \operatorname*{arg\,max}_{\theta_e}  \prod_{i=1}^I p(z_{ij}=y_{ij} \vert b_i) \label{eqn:MML}
\end{align}

Subsequently, the selected data subset is used to fine-tune the LLM. 
This process is iterative: after each epoch, the LLM's convergence is checked. 
If fine-tuning has not converged, the process repeats to re-estimate the LLM's ability and select new data for further fine-tuning. 

When implementing DDS-MAE, we encountered two significant challenges.
First, the initial evaluated ability of the model is often low, and in some cases, the available training data is limited. 
This scenario results in insufficient data utilization for model training in the initial epoch, potentially causing the model's ability to stagnate and impeding further data selection. 
To mitigate this issue, we implemented an adaptive solution: if the model's ability fails to improve over two consecutive epochs, we incrementally increase the ability parameter by 0.1. 
This adjustment enables the training process to overcome initial saddle points and facilitates continued model improvement.
Second, when utilizing the entire training dataset to evaluate the model's ability, the IRT calculation time becomes prohibitively long, particularly for large-scale datasets. 
For instance, the AG News dataset includes almost 1 million training examples \cite{zhang2015character}, and the ability estimation at each epoch consumes hundreds of minutes, comparable to the entire model training duration. 
To address this computational bottleneck, we propose a sampling-based method to evaluate the model's ability on a randomly selected subset of one thousand data points. 
This approach effectively balances estimation efficiency and accuracy, significantly reducing computational overhead while maintaining robust ability estimates.
These refinements serve to enhance the scalability and efficiency of our framework, enabling its application across a diverse range of tasks and dataset sizes.


\subsection{DDS-MAE Training Process}
\label{ssec:training}

Algorithm \ref{alg:dcl} describes the training procedure in detail.
Note that we assume that example difficulties have been learned offline using IRT-AC (see \S \ref{ssec:IRT-AC}).
Each example in the training set has an estimated difficulty parameter ($b_i$).
The first step of DDS-MAE is to estimate the ability of the model using the estimation function (\S \ref{ssec:DDS-MAE}, Alg. \ref{alg:dcl} line \ref{alg:estimation}). 
To do this, we use part of the training set, but crucially, only to get response data, not to update parameters (i.e., no backward pass). 
We do not use a held-out validation set for estimating ability because we do not want the validation set to influence training.
In our experiments, the validation set is only used for early stopping.
Model outputs are obtained for the training set, and graded as correct or incorrect as compared to the gold standard label (Alg. \ref{alg:dcl} line \ref{alg:identity}). 
This response pattern is then used to estimate model ability at the current epoch ($\hat{\theta}_e$, Alg. \ref{alg:dcl} line \ref{alg:mle}).
Once ability is estimated, data selection is done by comparing estimated ability to the examples' difficulty parameters.
If the difficulty of an example is less than or equal to the estimated ability, then the example is included in training for this epoch (Alg. \ref{alg:dcl} line \ref{alg:filter}).
The model is then trained with the training data subset (Alg. \ref{alg:dcl} line \ref{alg:training}).

\begin{algorithm}[!t]
	\small 
	\caption{Training process with DDS-MAE}
	\hspace*{\algorithmicindent}\textbf{Input:} Data ($X$, $Y$), model $M$, difficulties $B$, $\text{num\_epochs}$ 
	\hspace*{\algorithmicindent}
	\textbf{Output:} Learned model $M^{\prime}$ 
	\begin{algorithmic}[1]
		\Procedure{ability\_est}{$Y, \hat{Y}, B$}
		\State $Z = \forall_{y \in Y} \mathbf{I}[y_i = \hat{y_i}]$ \label{alg:identity}
		\State $\hat{\theta}_e = \operatorname*{arg\,max}_\theta p(Z \vert \theta, b)$ \label{alg:mle} 
		\State return $\hat{\theta_e}$
		\EndProcedure
		\State $M^{\prime} = M$
		\For{$e$ in $\text{num\_epochs}$}
		\State $\hat{Y} = M^{\prime}(X)$
		\State $\hat{\theta}_e = \texttt{ABILITY\_EST}(Y, \hat{Y}, D)$  \label{alg:estimation}
		\State $X_e, Y_e = \{(x,y): b_i \leq \hat{\theta}_e\}$  \label{alg:filter} 
		\State $M^{\prime} = \texttt{train}(M^{\prime}, X_e, Y_e)$ \label{alg:training}
		\EndFor 
		\State return $M^{\prime}$
	\end{algorithmic} 
	\label{alg:dcl}
\end{algorithm}

In contrast to other TS methods in the literature, the training data size does not have to be monotonically increasing with DDS-MAE. 
\modelabbr~adds or removes training data based not on a fixed step schedule but rather by probing the model at each epoch and using the estimated ability to match data to the model.
This way, if a model has a high estimated ability early in training, then more data can be added to the training set more quickly, and learning is not artificially slowed down due to the curriculum schedule.
If a model's performance suffers when adding data too quickly, then this will be reflected in lower ability estimates, which leads to less data selected in the next epoch.

\subsection{Time Complexity Analysis} 
\label{sssec:timeCompl}

In the DM (IRT-AC), we have two components: training the models in the AC to generate response patterns and estimating difficulty via VI-IRT. 
Assuming transformer-based architecture models in the AC, the time complexity to fine-tune the models is $O(K_{\text{FT}}MN_{\text{val}}/BL(n^2d+nd^2))$, where $K_{\text{FT}}$ is the number of optimization iterations, $B$ is the batch size, $L$ is the number of layers, $n$ is the sequence length, $d$ is the data dimension, $M$ is the number of models in the artificial crowd, and $N_{\text{val}}$ is the size of the validation set \citep{khan2022transformers,efficient_transformer}. 
For VI-IRT, the time complexity is $O(K_{\text{VI}}N_{\text{train}})$. 
We note here that the complexity associated with fine-tuning IRT-AC models is an offline cost that, once run, generates response patterns and difficulty estimates that can be reused. 
In particular, estimating IRT models with VI has been shown to reduce runtime, in particular when leveraging GPUs \cite{lalor2023py}. 
Once the difficulty parameters of the training data are estimated, they can be used for multiple training runs.
If new models are added to the AC, then only those new models must be fine-tuned, and VI-IRT is then rerun on the entire response pattern pool.
We empirically assess this cost in Section \ref{sssec:compTimeIRT}.

Our DDS-MAE approach introduces two additional steps to traditional fine-tuning: model ability estimation and training data filtering (Alg. \ref{alg:dcl}, lines \ref{alg:estimation} and \ref{alg:filter}). 
The model ability estimation procedure consists of (i) comparing each prediction with its corresponding true label, with a time complexity of $O(N_{\theta})$, where $N_{\theta}$ is the number of estimated training examples; and (ii) Maximum Likelihood Estimation (MLE) using the Nelder-Mead method, as described in Section \ref{ssec:DDS-MAE}, with a time complexity of $O(K_{\theta}N_{\theta})$, where $K_{\theta}$ is the number of optimization iterations (typically, $K_{\theta}=10$). 
The training data filtering step also exhibits linear complexity, $O(N_{\text{train}})$. 
Consequently, the total time complexity for DDS-MAE is $O(N_{\theta}) + O(K_{\theta}N_{\theta}) + O(N_{\text{train}})$. 
For conventional transformer-based model training in each epoch, the time complexity is $O(K_{\text{FT}}N_{\text{train}}/BL(n^2d+nd^2))$, where $K_{\text{FT}}$ is the number of optimization iterations, $N_{\text{train}}$ is the number of training data, $B$ is the batch size, $L$ is the number of layers, $n$ is the sequence length, and $d$ is the data dimension \citep{khan2022transformers,efficient_transformer}. 
Theoretically, our proposed method is asymptotically equivalent to the conventional training process; we experimentally verify that the additional time required is low and is typically offset by overall faster convergence (\S\ref{sssec:runtime}).

\section{Experiments}
\label{sec:experiments}

In this section, we first introduce the experimental setup in Section \ref{ssec:setup}. 
Then, we validate \modelabbr{}'s performance and compatibility across different LLM models and compare \modelabbr{} with other advanced CL methods in Sections \ref{ssec:LM models}. 
We conduct multiple analyses in Section \ref{ssec:moreAnalyses} to demonstrate the contribution of each component to \modelabbr{}'s overall performance.
In Section \ref{ssec:gsm8k}, we present results applying \modelabbr{} to a math problem text generation task to demonstrate its use beyond traditional classification tasks.

\subsection{Experimental Setup}
\label{ssec:setup}

\phantomsection
\subsubsection{Datasets}
\label{ssec:data_description}

We conduct our experiments with eight datasets. 
MedQA-UMLS \cite{jin2020disease} is a recent multiple-choice QA dataset consisting of medical exam questions.
The AG News dataset \cite{zhang2015character} includes almost 1 million training examples across 14 classes.
Lastly, we evaluate \modelabbr{} on natural language understanding tasks from the GLUE \citep{wang2019glue} benchmark for consistency with and comparability to prior CL research \citep{senguptagood,maharana2022curriculum,wan-etal-2020-self,xu2020curriculum,lalor-yu-2020-dynamic}. 
We specifically consider the six GLUE classification tasks\footnote{We exclude the WNLI dataset due to dataset construction inconsistencies; see \url{https://gluebenchmark.com/faq} note 12.}
 which cover natural language inference (MNLI, RTE, QNLI), duplicate detection (MRPC, QQP), and sentiment analysis (SST-2).\footnote{Because test set labels for our tasks are only available via the GLUE evaluation server, we use the held-out validation sets to measure performance, consistent with prior work. For training, we use a $90\%$ - $10\%$ split of the training data and use the 10\% split as our held-out validation set for early stopping. We can then use the full validation set as our test set to evaluate performance across experiments without making multiple submissions to the GLUE server.}
Dataset details and summary statistics are provided in Table \ref{tab:glue_superglue_stats}.

\begin{table}[ht!]
    \centering
    \small
        \begin{tabular}[t]{llccccp{5.5cm}}
            \toprule
            \multicolumn{1}{c}{\bf Dataset} & \bf Train & \bf Validation & \bf Test & \bf Labels & \bf Reference  \\
            \midrule
            MedQA & 9.2k & 1.02k & 1.27k & 4 & \citet{jin2020disease} \\
            \midrule
            AG News & 995.3k & 124.4k & 124.4k & 14 & \citet{zhang2015character}\\
            \midrule 
            MNLI & 353k & 39k & 9.8k  & 3 & \citet{williams2018broad}\\
            MRPC & 3.3k & 366 & 409  & 2 & \citet{dolan2005automatically}\\
            QNLI & 94k & 10k & 5.5k & 2 & \citet{wang2019glue} \\
            QQP & 327k & 36k & 40k & 2 &\citet{Iyer_Dandekar_Csernai_2017}\\
            RTE & 2.2k & 249 & 278& 2 &\citet{bentivogli2009fifth} \\
            SST-2 & 61k & 6.7k & 873 & 2&\citet{socher_recursive_2013}\\
            \bottomrule
        \end{tabular}            
    \caption{Statistics of the datasets used in the experiments, including training, validation, and test set sizes, number of labels, and original references.}

    \label{tab:glue_superglue_stats}
\end{table}

\subsubsection{IRT-AC Models}
\label{ssec:acModels}

For the artificial crowd models, we include BERT~\citep{kenton2019bert}, DistillBERT~\citep{sanh2019distilbert}, RoBERTa~\citep{liu2019roberta}, DeBERTa~\citep{he2020deberta}, ALBERT~\citep{lan2019albert}, XLNet~\citep{yang2019xlnet}, ELECTRA~\citep{clark2020electra}, T5~\citep{2020t5}, BART~\citep{lewis2019bart}, Llama3.1-8B~\citep{grattafiori2024llama}, Qwen2.5-7B~\citep{yang2024qwen2}, and GPT-2~\citep{radford2019language}. 
We collect response patterns from each AC model with fine-tuning for 0, 1, 3, 5, and 10 epochs for a total of 60 AC models.

\subsubsection{Benchmarking Models}
\label{ssec:benchmarkModels}

We test the effectiveness and compatibility of \modelabbr{} by integrating it with different types of transformer architectures, including encoder-based and decoder-based models.
We include models of varying parameter size and complexity to demonstrate \modelabbr{}'s effectiveness across a variety of LLMs. 

DeBERTaV3 \citep[86M parameters,][]{he2022debertav3} is an encoder-based pre-trained language model developed by Microsoft. 
It uses a disentangled attention mechanism to better capture word dependencies and contextual information, enhancing performance on various natural language understanding tasks.
 
GPT-2 \citep[124M parameters,][]{radford2019language} is a decoder-based language model from OpenAI. 
Trained on diverse internet text, it excels in generating coherent, contextually relevant text for tasks like translation, summarization, and question-answering without task-specific training data.

Llama3.1-8B \citep[8B parameters,][]{grattafiori2024llama} is a decoder-based pre-trained language model developed by Meta. 
It is part of the Llama 3.1 collection of multilingual models, designed for both commercial and research use, and has been optimized for dialogue use cases with improved safety and helpfulness through supervised fine-tuning (SFT) and reinforcement learning with human feedback (RLHF). 
Trained on over 15 trillion tokens of publicly available data, Llama3.1-8B supports a 128K context length and demonstrates strong performance on various benchmarks including as text generation, coding, and multilingual conversation.

Qwen2.5-7B \citep[7.61B parameters,][]{yang2024qwen2} is a decoder-based language model from Alibaba Cloud as part of their Qwen2.5 series. 
It features a transformer architecture with enhancements including RoPE, SwiGLU, RMSNorm, and Attention QKV bias, and supports a context length of up to 131,072 tokens. 
Qwen2.5-7B shows significant improvements over previous Qwen models in knowledge-intensive tasks, coding, mathematics, instruction following, long-text generation, and multilingual capabilities, supporting over 29 languages.


\subsubsection{Benchmark CL Methods}
\label{ssec:benchmarkMethods}

To benchmark the performance of our proposed \modelabbr{} framework, we compare several CL methods that cover each of the four CL categories, i.e., predefined CL, self-paced learning, transfer teacher, and RL teacher \citep{wang2021survey}. 

\paragraph{Predefined CL} We evaluate predefined CL based on prior work that defines a predefined competence value~\citep{platanios_competence-based_2019}. 
For DM, we use sentence length ($d_{SL}$) and word rarity ($d_{WR}$). 
For TS, we use a linear or root function to adjust the training pace. 
We set the initial competence ($c_0$) to be $0.01$ and set the point where the model is fully competent ($T$) to be equal to $\text{total\_epochs}/2$; the predefined CL reaches competence halfway through training and trains with the full training set for the second half.

\paragraph{Self-paced learning (SPL)} We evaluate a novel self-paced learning algorithm \citep{zhang2024weighted} that incorporates belief functions to overcome limitations of traditional SPL approaches. Unlike traditional SPL methods, it mitigates the tendency to misjudge sample difficulty based solely on learning loss in early training stages, which often leads to the premature inclusion of hard-to-classify examples.
The DM combines evidential uncertainty and learning loss to characterize difficulty.
The TS adjusts the balance between evidential uncertainty and learning loss across training stages, allowing for more appropriate sample selection as the model improves.

\paragraph{Transfer teacher} We evaluate a transfer teacher-based CL from recent work \citep{maharana2022curriculum} that uses Question Answering Probability (QAP) as a DM scoring function and an adaptive function for TS. This method uses a pre-trained teacher model fine-tuned on the training data, and the QAP metric from the teacher's outputs serves as the DM to rank training examples by difficulty. 

\paragraph{RL Teacher} For RL teacher, we evaluate MPDistil \citep{senguptagood}. 
MPDistil is a meta-policy knowledge distillation framework with a reward-based policy learner as DM and a meta-reinforcement learning-based model and reward function as TS.

\subsubsection{Hyperparameter Tuning and Hardware Platform}

For all datasets (MedQA, AG News, and GLUE) the maximum input token length was set to the 95th percentile of token lengths within each respective dataset to balance representational capacity and computational load. The largest batch size that fit within the available GPU memory was employed to optimize throughput. Fine-tuning was run for 20 training epochs with early stopping based on validation set performance to mitigate overfitting and reduce training duration \citep{yao2007early}.
In order to validate our experimental results and report on the stability of each method, we ran the experiments five times and reported the mean and standard deviation of the results.

Hyperparameters for all the models were established via grid search. 
These optimized parameters were then consistently applied across all CL methodologies evaluated in this study. 
For the optimization method, we employed the AdamW optimizer \citep{loshchilov2017decoupled}. 
The learning rate was selected from the set $\{1 \times 10^{-5}, 2 \times 10^{-5}, 3 \times 10^{-5}\}$ based on preliminary experiments. 
A weight decay of $0.01$ was applied, excluding layer normalization and bias terms, while standard default values were maintained for AdamW ($\beta_1=0.9$, $\beta_2=0.999$, and $\epsilon=1 \times 10^{-8}$). 
The learning rate schedule incorporated a linear warm-up phase, typically accounting for 6-10\% of the total training steps, followed by a linear decay of the learning rate. 
Our batching strategy for these models involved adjusting the batch size (e.g., ranging from 2 to 64 depending on the specific model and dataset) and utilizing gradient accumulation (e.g., 1 to 16 steps) to achieve a larger effective batch size while adhering to GPU memory limitations.

Fine-tuning Llama3.1-8B and Qwen2.5-7B on the AG News dataset, which comprises almost one million training data points, presented significant GPU memory challenges. 
To address these constraints, 
we employed parameter-efficient fine-tuning (PEFT), specifically QLoRA \citep{dettmers2023qlora}.
QLoRA is an enhanced version of low-rank adaptation \citep[LoRA, ][]{hu2022lora} that incorporates aggressive quantization.
Specifically, our implementation involved 4-bit NormalFloat (NF4) quantization for the model parameters, a \texttt{bfloat16} compute data type within the quantization layers, and double quantization. 
QLoRA was further complemented by mixed-precision training via PyTorch AMP, which utilized bfloat16 numerical precision and a \texttt{GradScaler}. 
This combination of PEFT and mixed-precision training substantially decreased memory requirements and often accelerated training throughput.

To fit our IRT model, we use the py-irt Python package \citep{lalor2023py}, which is built on top of the Pyro probabilistic programming language \citep{bingham2018pyro}. 
All LLMs were implemented from Huggingface.\footnote{\url{https://github.com/huggingface/transformers}}
One NVIDIA H100 GPU was used to conduct all of the experiments.

\subsection{Incorporating \modelabbr{} in LLM Fine-tuning} 
\label{ssec:LM models}



\begin{figure}[hbt]
    \centering
    \includegraphics [width=\textwidth] {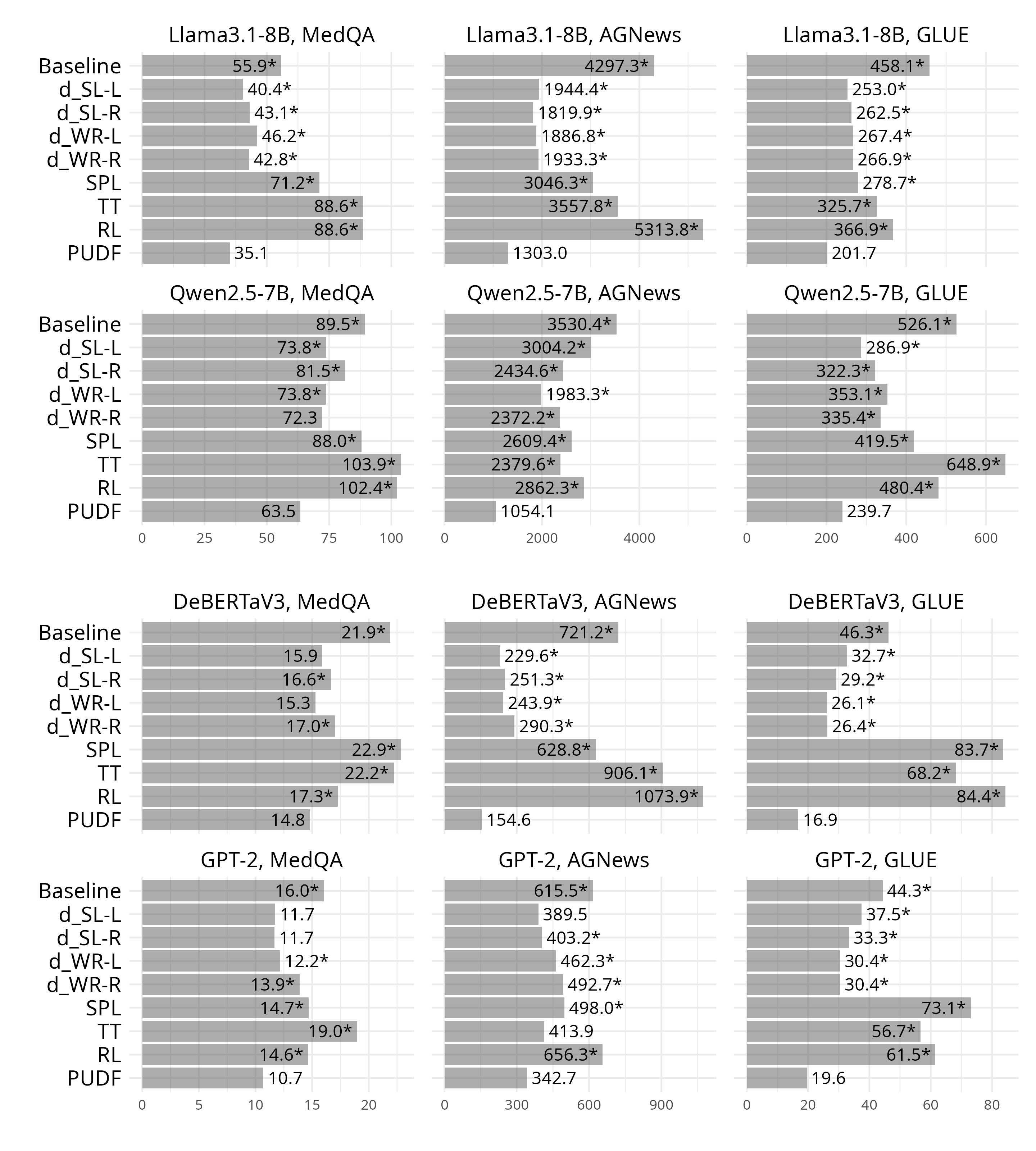}
    \caption{Comparing training time between \modelabbr{} and other CL methods. All runtimes reported in minutes. GLUE scores are reported as the mean across tasks, pooled by runs.\\
    $*$Indicates that the runtime is significantly longer than \modelabbr{} (Welch's single-tailed t-test with Benjamini-Hochberg correction, $\alpha < 0.05$).}
    \label{fig:traintimeCL}
\end{figure}

In this section, we report our main results comparing \modelabbr{} to a no-CL baseline as well as other CL frameworks. 
We report predictive performance via accuracy (Table \ref{tab:mainResult}) as well as runtime (Figure \ref{fig:traintimeCL}) results for MedQA, AG News, and GLUE. 
For GLUE, we averaged results across tasks; results for individual GLUE tasks are presented in the appendices (Table \ref{tab:appendixResult} and Figure \ref{fig:traintimeCLGLUE}) for space considerations.
We report the mean and standard deviation of five runs of each configuration.
In all cases, we conduct a one-tailed Welch's t-test to determine whether the performance of \modelabbr{} is significantly better than the benchmark methods. 
We used Benjamini-Hochberg correction ($\alpha < 0.05$) to control the false discovery rate across multiple comparisons \citep{benjamini1995controlling,ormerod2024kitchen}.

\begin{table}[!bht]
	\caption{\label{tab:mainResult} Accuracy results comparing PUDF with other CL Methods. Results are averaged over 5 runs with standard deviations as subscripts. The best performing method for each model is in \textbf{bold}; the second-best model is \underline{underlined}. For GLUE, we report the mean scores across tasks, pooled by runs.}
	\centering
	\begin{tabular}[t]{lllll}
		\toprule
		\textbf{Model} & \textbf{Method} & \textbf{MedQA}                     & \textbf{AGNews}                    & \textbf{GLUE}                      \\
		\midrule
		DeBERTaV3      & Baseline        & $32.69_{\pm 0.10}^{*}$             & $70.66_{\pm 0.35}^{*}$             & $89.65_{\pm 0.40}^{*}$             \\
		               & d\_SL-L         & $26.49_{\pm 1.43}^{*}$             & $65.62_{\pm 1.99}^{*}$             & $88.13_{\pm 0.35}^{*}$             \\
		               & d\_SL-R         & $27.25_{\pm 1.53}^{*}$             & $65.55_{\pm 2.47}^{*}$             & $88.58_{\pm 0.66}^{*}$             \\
		               & d\_WR-L         & $27.04_{\pm 1.80}^{*}$             & $64.02_{\pm 2.94}^{*}$             & $87.18_{\pm 0.72}^{*}$             \\
		               & d\_WR-R         & $27.95_{\pm 1.41}^{*}$             & $68.04_{\pm 1.87}^{*}$             & $88.97_{\pm 0.38}^{*}$             \\
		               & SPL             & $33.34_{\pm 0.33}^{*}$             & $70.54_{\pm 0.90}^{*}$             & $89.12_{\pm 0.86}^{*}$             \\
		               & TT              & $32.53_{\pm 0.52}^{*}$             & $70.93_{\pm 1.30}^{*}$             & $89.44_{\pm 1.35}$                 \\
		               & RL              & $\underline{33.35}_{\pm 0.39}^{*}$ & $\underline{71.87}_{\pm 0.56}^{*}$ & $\underline{89.99}_{\pm 0.69}$     \\
		               & PUDF            & $\mathbf{34.12}_{\pm 0.16}$        & $\mathbf{72.43}_{\pm 0.27}$        & $\mathbf{90.76}_{\pm 0.20}$        \\
		\cmidrule{1-5}
		GPT-2          & Baseline        & $27.97_{\pm 0.15}^{*}$             & $65.71_{\pm 1.13}^{*}$             & $\underline{82.34}_{\pm 0.30}^{*}$ \\
		               & d\_SL-L         & $26.09_{\pm 1.54}^{*}$             & $64.69_{\pm 1.34}^{*}$             & $79.13_{\pm 0.73}^{*}$             \\
		               & d\_SL-R         & $25.01_{\pm 1.99}^{*}$             & $63.15_{\pm 2.08}^{*}$             & $80.33_{\pm 0.57}^{*}$             \\
		               & d\_WR-L         & $26.09_{\pm 1.36}^{*}$             & $67.02_{\pm 0.17}^{*}$             & $79.44_{\pm 0.67}^{*}$             \\
		               & d\_WR-R         & $26.45_{\pm 1.22}^{*}$             & $\mathbf{67.77}_{\pm 0.23}$        & $80.54_{\pm 0.59}^{*}$             \\
		               & SPL             & $27.43_{\pm 0.86}^{*}$             & $66.01_{\pm 1.18}^{*}$             & $80.24_{\pm 0.43}^{*}$             \\
		               & TT              & $28.67_{\pm 0.50}^{*}$             & $67.05_{\pm 0.53}^{*}$             & $81.80_{\pm 0.61}^{*}$             \\
		               & RL              & $\underline{28.91}_{\pm 0.26}^{*}$ & $\underline{67.35}_{\pm 0.35}$     & $82.21_{\pm 0.55}^{*}$             \\
		               & PUDF            & $\mathbf{29.50}_{\pm 0.22}$        & $\mathbf{67.77}_{\pm 0.57}$        & $\mathbf{83.03}_{\pm 0.30}$        \\
		\cmidrule{1-5}
		Llama3.1-8B    & Baseline        & $55.15_{\pm 0.09}^{*}$             & $72.10_{\pm 0.15}^{*}$             & $\underline{90.59}_{\pm 0.42}^{*}$ \\
		               & d\_SL-L         & $45.89_{\pm 4.22}^{*}$             & $70.73_{\pm 2.06}^{*}$             & $87.78_{\pm 1.45}^{*}$             \\
		               & d\_SL-R         & $50.22_{\pm 3.68}^{*}$             & $71.20_{\pm 1.86}^{*}$             & $88.70_{\pm 0.87}^{*}$             \\
		               & d\_WR-L         & $57.09_{\pm 1.07}^{*}$             & $71.80_{\pm 1.85}^{*}$             & $89.27_{\pm 0.63}^{*}$             \\
		               & d\_WR-R         & $55.98_{\pm 1.49}^{*}$             & $71.16_{\pm 1.84}^{*}$             & $89.10_{\pm 0.76}^{*}$             \\
		               & SPL             & $\underline{58.60}_{\pm 0.04}^{*}$ & $73.87_{\pm 0.23}^{*}$             & $90.36_{\pm 0.36}^{*}$             \\
		               & TT              & $58.01_{\pm 0.03}^{*}$             & $72.67_{\pm 1.06}^{*}$             & $90.08_{\pm 0.50}^{*}$             \\
		               & RL              & $57.90_{\pm 0.24}^{*}$             & $\underline{74.77}_{\pm 0.21}^{*}$ & $90.50_{\pm 0.38}^{*}$             \\
		               & PUDF            & $\mathbf{58.82}_{\pm 0.09}$        & $\mathbf{75.08}_{\pm 0.11}$        & $\mathbf{91.26}_{\pm 0.35}$        \\
		\cmidrule{1-5}
		Qwen2.5-7B     & Baseline        & $63.52_{\pm 0.16}^{*}$             & $71.63_{\pm 0.20}^{*}$             & $89.82_{\pm 0.56}^{*}$             \\
		               & d\_SL-L         & $62.02_{\pm 1.18}^{*}$             & $69.97_{\pm 2.88}^{*}$             & $\underline{90.15}_{\pm 0.51}^{*}$ \\
		               & d\_SL-R         & $62.03_{\pm 0.97}^{*}$             & $69.43_{\pm 2.82}^{*}$             & $89.43_{\pm 0.63}^{*}$             \\
		               & d\_WR-L         & $63.47_{\pm 0.79}^{*}$             & $67.44_{\pm 3.15}^{*}$             & $89.59_{\pm 0.55}^{*}$             \\
		               & d\_WR-R         & $63.31_{\pm 1.10}^{*}$             & $69.89_{\pm 1.20}^{*}$             & $89.77_{\pm 0.50}^{*}$             \\
		               & SPL             & $64.17_{\pm 0.37}^{*}$             & $71.55_{\pm 0.46}^{*}$             & $88.96_{\pm 0.78}^{*}$             \\
		               & TT              & $64.13_{\pm 0.28}^{*}$             & $\underline{71.71}_{\pm 0.26}^{*}$ & $89.35_{\pm 0.53}^{*}$             \\
		               & RL              & $\underline{64.52}_{\pm 0.12}^{*}$ & $\mathbf{72.93}_{\pm 0.07}$        & $89.94_{\pm 0.60}^{*}$             \\
		               & PUDF            & $\mathbf{65.02}_{\pm 0.22}$        & $\mathbf{72.93}_{\pm 0.32}$        & $\mathbf{90.90}_{\pm 0.43}$        \\
		\bottomrule& & 
	\end{tabular}

	{\raggedright \footnotesize $^*$Indicates that the value is significantly lower than the \textbf{best accuracy} in the column (Welch's single-tailed t-test with Benjamini-Hochberg correction, $\alpha < 0.05$).}
\end{table}

Table~\ref{tab:mainResult} presents a comprehensive comparison of various CL methods across benchmark datasets. 
\modelabbr{} consistently outperforms traditional training (no CL) as well as the comparison CL methods in accuracy across datasets.
For example, training Llama3.1-8B with \modelabbr{} results in relative accuracy improvements over the no-CL baseline of 4.15\%, 6.65\%, and 0.74\% for AG News, MedQA, and GLUE, respectively. 
Relative improvements over the best performing benchmark CL method are 0.42\%, 0.38\%, and 0.74\%, respectively. 
In particular, when considering the larger LLM benchmark models (Llama3.1-8B and Qwen2.5-7B), \modelabbr{} accuracy is significantly higher than all benchmark methods with one exception: the RL benchmark with Qwen2.5-7B for AG News.  
In this case, our runtime results (Figure \ref{fig:traintimeCL}) show that \modelabbr{} is significantly faster compared to all benchmarks for these two models, including the RL method for Qwen2.5-7B on AGNews.
For GPT-2 and DeBERTa, \modelabbr{} improvements are significant in all cases except for two methods on AG News (d\_WR-R and RL for GPT-2) and two methods on GLUE (TT and RL for DeBERTaV3).
Runtime improvements for these two models are significant in most cases (Figure \ref{fig:traintimeCL}).

The performance advantages of PUDF correlate with dataset and model characteristics. Our analysis reveals two distinct poles of improvement. For the large-scale AG News dataset, with nearly one million examples, PUDF's primary benefit is a dramatic 69.68\% relative runtime reduction for Llama3.1-8B over the baseline. This efficiency stems from the DDS-MAE scheduler, which dynamically selects an optimal data subset, avoiding the high cost of training on the full dataset in early epochs. Conversely, for the MedQA dataset, a task identified as highly challenging by our IRT-AC analysis, PUDF shows its most substantial accuracy gains (a 6.65\% relative improvement for Llama3.1-8B). This suggests that in complex domains, the robust, global IRT-AC difficulty metric is more effective than simple heuristics or volatile early-stage metrics, guiding the model to a superior convergence.

Furthermore, the benefits of PUDF scale with model size, with Llama3.1-8B and Qwen2.5-7B deriving the most significant advantages. These large models are costly to fine-tune, and alternative CL frameworks (e.g., RL and Transfer Teachers) introduce substantial online computational overhead. PUDF's architecture relegates its main expense (IRT-AC) to a one-time, offline process. The online DDS-MAE component adds only minimal overhead, a single forward pass and lightweight estimation per epoch. This combination of a robust, pre-computed difficulty estimation and an efficient dynamic scheduler allows large models to converge faster to a better performance optimum. While smaller models such as DeBERTaV3 and GPT-2 also benefit, their lower intrinsic cost and capacity result in positive but less pronounced gains.




For training time, again looking at Llama3.1-8B, \modelabbr{} results in relative improvements of 69.68\%, 37.21\%, and 55.97\% over the no-CL baseline for AG News, MedQA, and GLUE, respectively. 
Relative improvements over the fastest CL alternative, which is usually not the highest accuracy option, are 28.40\%, 13.12\%, and 20.28\%, respectively. 
Comparing the training time between \modelabbr{} and the RL benchmark for Llama3.1-8B on AG News, the relative training time improvement is 75.48\%.

Across datasets and benchmark methods, \modelabbr{} is significantly faster when using DeBERTaV3, Qwen2.5-7B, and Llama3.1-8B as the fine-tuning model. 
For GPT-2, there are three cases where runtime performance is comparable, but in these cases predictive performance of the benchmark methods lags behind \modelabbr{} (AGNews, d\_SL-L, MedQA, d\_SL-L and d\_SL-R).

Notably, the best CL benchmark for training time is often less performant in terms of accuracy.
The benchmarking CL methods consistently underperform either on predictive performance or training time. 
Our results highlight \modelabbr{}'s ability to maintain competitive accuracy while significantly reducing training time compared to other CL approaches.

While results from our prior work were mixed in terms of performance and efficiency~\citep{lalor-yu-2020-dynamic}, these results indicate consistent improvements with \modelabbr{} and also provide new insights.
Specifically, \modelabbr{} outperforms both traditional training and benchmark CL methods on large (AG News) and difficult (MedQA) datasets, whereas prior work focused on the smaller, relatively easier GLUE datasets. 
In addition, \modelabbr{} outperforms advanced, automated CL methods as well as predefined CL methods. 
Lastly, \modelabbr{} improvements are consistent across smaller (DeBERTaV3, GPT-2) and larger (Llama3.1-8B, Qwen2.5-7B) LLMs, which demonstrates the consistency of the method above and beyond the smaller models (LSTM, BERT) evaluated in prior work. 
The consistent improvements and low standard deviations for \modelabbr{} provide strong evidence that \modelabbr{} can outperform existing CL techniques in terms of both accuracy and training efficiency across benchmark datasets. 
The observed improvements, especially for larger datasets such as AG News and more difficult datasets such as MedQA, highlight the robustness and effectiveness of \modelabbr{}.

\subsection{Further Analyses}
\label{ssec:moreAnalyses}

Here, we provide in-depth analyses of \modelabbr{} to better understand how and why our framework outperforms other methods.
We also analyze in detail the characteristics of \modelabbr{}'s DM and TS to evaluate the benefits of applying IRT to the problem of CL.

\subsubsection{Ablation Study} 
\label{ssec:ablation}

To clarify the factors contributing to \modelabbr{}'s accuracy and training time improvements, we conduct the following ablation study. 
For the ablation, we focus on Qwen2.5-7B based on its performance in our main experiments.
Specifically, we interchange the DM and TS components between \modelabbr{} and predefined CL methods.
Table~\ref{tab:ablationResult} presents the results of our ablation study, from which we can draw several conclusions.
In the first part of the experiment, we employed $d_{SL}$ or $d_{WR}$ as the DM in conjunction with DDS-MAE as the TS. Specifically, we first apply min–max normalization to the raw difficulty scores $d_{SL}$ and $d_{WR}$ so that they lie within the predefined IRT-AC difficulty interval. Then, at each epoch, DDS-MAE dynamically selects training samples according to the model’s current capability and the normalized difficulty values.
The results reveal that DDS-MAE significantly reduces training time compared to rule-based training schedulers (i.e., Linear and Root, Figure \ref{fig:traintimeCL}). 
Despite the additional time required for dynamic evaluation of model ability in each epoch, this overhead is negligible relative to the overall training time and contributes to improved model convergence.
The combination of $d_{SL}$ or $d_{WR}$ with DDS-MAE results in decreased accuracy compared to predefined methods (Table \ref{tab:mainResult}) and \modelabbr{}, while achieving training times that were significantly slower than \modelabbr{} on the AGNews and GLUE datasets. Further analysis of the difficulty distributions generated by $d_{SL}$ and $d_{WR}$ reveals a mismatch with the model ability evaluated by DDS-MAE.

Next, we utilize IRT-AC as the DM in combination with Linear or Root training scheduler functions. 
Our findings indicate that accuracy improves compared to predefined methods and approaches that of \modelabbr{}, suggesting that the difficulty generated by IRT-AC is more suitable for model training than $d_{SL}$ and $d_{WR}$. This underscores the efficacy of the IRT-AC method compared to the rule-based DMs. 
The training time remains similar to predefined methods but is slower than \modelabbr{}. This observation is consistent with expectations, as IRT-AC is responsible for labeling data difficulty, while the training scheduler controls the training pace.
While combining IRT-AC with Root or Linear schedulers improves accuracy but fails to improve training time, pairing it with DDS-MAE (i.e., \modelabbr{}) yields substantial enhancements, demonstrating the importance of complementarity between the DM and TS.
In conclusion, \modelabbr{} (IRT-AC + DDS-MAE) consistently outperforms other combinations across benchmark datasets, demonstrating the synergistic effect of its components. 
The IRT-AC component enhances model accuracy, while the DDS-MAE algorithm, guided by the IRT-AC difficulty scores, significantly reduces training time, resulting in an efficient and effective CL approach.

\begin{table}
	\caption{\label{tab:ablationResult}Results of ablation study for Qwen2.5-7B. The best performing method for each model is in \textbf{bold}; the second-best model is \underline{underlined}. For GLUE, we report the mean scores across tasks, pooled by runs.}
	\centering
	\small
	\begin{tabular}[t]{p{1.75cm}lllll}
		\toprule
		\textbf{Metric}                                                    & \textbf{DM} & \textbf{TS}    & \textbf{MedQA}      & \textbf{AGNews}        & \textbf{GLUE}   \\
		\midrule
		Accuracy                                                          & d\_SL       & DDS-MAE        & $\underline{64.55}_{\pm 2.71}$  & $70.70_{\pm 2.47}$     & $88.97_{\pm 0.47}^{*}$  \\
& d\_WR       & DDS-MAE        & $62.90_{\pm 1.38}^{*}$   & $71.22_{\pm 1.44}^{*}$     & $88.65_{\pm 0.93}^{*}$  \\
& IRT-AC      & Root           & $63.19_{\pm 0.79}^{*}$  & $\underline{72.19}_{\pm 0.46}^{*}$     & $\underline{89.30}_{\pm 0.59}^{*}$  \\
& IRT-AC      & Linear         & $63.02_{\pm 0.83}^{*}$  & $71.99_{\pm 0.45}^{*}$     & $89.19_{\pm 0.45}^{*}$  \\
& IRT-AC      & DDS-MAE (PUDF) & $\mathbf{65.02}_{\pm 0.22}$  & $\mathbf{72.93}_{\pm 0.32}$     & $\mathbf{90.90}_{\pm 0.43}$  \\
	\midrule
	\multirow{2}{*}{\parbox{1.75cm}{Training Time\newline (minutes)}} & d\_SL       & DDS-MAE        & $76.80_{\pm 7.18}^{*}$  & $\underline{1587.60}_{\pm 468.7}^{*}$   & $330.19_{\pm 45.4}^{*}$ \\
& d\_WR       & DDS-MAE        & $\underline{68.55}_{\pm 6.14}$  & $1625.64_{\pm 258.55}^{*}$ & $\underline{316.78}_{\pm 48.5}^{*}$ \\
& IRT-AC      & Root           & $87.31_{\pm 8.66}^{*}$  & $2371.32_{\pm 335.0}^{*}$  & $382.51_{\pm 43.9}^{*}$ \\
& IRT-AC      & Linear         & $91.77_{\pm 13.96}^{*}$ & $1981.68_{\pm 269.92}^{*}$ & $323.97_{\pm 33.7}^{*}$ \\
& IRT-AC      & DDS-MAE (PUDF) & $\mathbf{63.51}_{\pm 7.60}$  & $\mathbf{1054.08}_{\pm 131.1}$  & $\mathbf{239.71}_{\pm 28.9}$ \\
\bottomrule& & 
\end{tabular}

        {\raggedright \footnotesize $^*$Indicates that the value is significantly worse than the \textbf{best value} in the column (Welch's single-tailed t-test with Benjamini-Hochberg correction, $\alpha < 0.05$).}
\end{table}

\newpage
\subsubsection{Computational Cost and Efficiency of \modelabbr{}}
\label{sssec:compTimeIRT}

\paragraph{IRT-AC}
The IRT-AC difficulty generation process is designed as an offline, one-time procedure per dataset; that said, its computation cost may still be high. 
The specifics of this overhead and its impact on overall training efficiency are detailed in Table~\ref{tab:pudf_pipeline_time_comparison}. 
As illustrated in Figure~\ref{fig:pudf_workflow}, the IRT-AC framework comprises two primary stages: (i) obtaining Prediction Correctness from the AC and (ii)  IRT Difficulty Estimation. 
For the Prediction Correctness stage, we employ a diverse set of 13 LLMs (\S\ref{ssec:acModels}). 
To enhance the robustness and diversity of the AC, these models are fine-tuned for varying epochs (0, 1, 3, 5, and 10) on the respective validation sets before generating predictions on the training data. 
A key practical advantage of this stage is its parallelizability, as each LLM in the AC can be fine-tuned independently, potentially reducing the effective wall-clock time with concurrent computation. 
The second stage, IRT Difficulty Estimation, then utilizes a 1PL IRT model fit using VI~\cite{lalor_learning_2019,lalor2023py}, to efficiently estimate example difficulties from the response patterns.

The time invested in these IRT-AC stages (e.g., 3.94 hours for AG News, 9.89 minutes for MedQA, and an average of 127.3 minutes for GLUE tasks, as shown in Table~\ref{tab:pudf_pipeline_time_comparison}) constitutes a manageable, upfront computational cost. 
More importantly, this initial overhead is offset by substantial gains in overall training efficiency when using the \modelabbr{} framework. 
By comparing the Qwen2.5-7B baseline training time with the Total Time (\modelabbr{})—which includes both the IRT-AC processing and the subsequent DDS-MAE-guided training—we observe consistent net time savings. 
For instance, the total pipeline time for AG News with \modelabbr{} is 22.73 hours, a 63.8\% reduction from the 62.87-hour baseline. 
Similar efficiencies are evident for MedQA (77.70 minutes vs. 94.37 minutes) and the GLUE average (383.3 minutes vs. 568.00 minutes). 
These results underscore that the IRT-AC overhead is not only tolerable, due to its offline and parallelizable nature, but its integration into the \modelabbr{} framework ultimately leads to a more time-efficient training process.
In addition, the main IRT-AC cost, Prediction Correctness, can be reduced by reusing previously stored AC model response patterns. 

\begin{table}[htb]
    \centering
    \caption{Comparison of training times for the baseline model versus the \modelabbr{}-guided approach for Qwen2.5-7B, broken down by stage. Time units are specified per dataset.}
    \label{tab:pudf_pipeline_time_comparison}
    \begin{tabular}{llccc}
    \toprule
    \multicolumn{2}{l}{\bf Fine-Tuning Approach}  & \bf MedQA & \bf AG News& \bf GLUE \\
    && (minutes) &(hours) & (minutes) \\ 
    \midrule
    \multirow{7}{*}{PUDF} & Prediction Correctness & 7.91 & 3.15 & 102.32\\
    & IRT Difficulty & 1.98 & 0.79 & 24.98 \\
    \cmidrule{3-5}
    & \textit{IRT-AC Total} & 9.89 & 3.94 & 127.30\\
    \cmidrule{2-5}
    &Ability Estimation& 8.78 &2.56 & 35.97\\
    & Fine-Tuning & 59.03 & 16.23 & 220.03 \\
    \cmidrule{3-5}
    & \textit{DDS-MAE Total} & 67.81 & 18.79 & 256.00\\
    \cmidrule{2-5}
    & \textbf{\textit{PUDF Total}} & 77.70 & 22.73 & 383.30\\
    \midrule
    \multicolumn{2}{l}{No CL Benchmark} & 94.37 & 62.87 & 568.00 \\
    \bottomrule
    \end{tabular}
\end{table}

\paragraph{DDS-MAE}
\label{sssec:runtime}

The DDS-MAE component of \modelabbr{} introduces specific computational steps at each epoch for its decision-making process related to data selection and scheduling.
We analyze the runtime characteristics of this component and, more importantly, the overall impact of the \modelabbr{} method on the total training duration when applied to the Qwen2.5-7B model, compared to a standard Qwen2.5-7B baseline.
As detailed in Table~\ref{tab:pudf_pipeline_time_comparison}, the computational time attributed directly to Ability Estimation is modest, with its duration corresponding to a small percentage (ranging from $4.07\%$ to $9.30\%$) of the original Qwen2.5-7B baseline's runtime on the respective datasets. This highlights that the additional computational steps introduced by the DDS-MAE component are relatively lightweight.
Crucially, despite this inherent processing time from the DDS-MAE component, the integration of \modelabbr{} leads to a substantial reduction in the overall training duration compared to the Qwen2.5-7B baseline. Table~\ref{tab:pudf_pipeline_time_comparison} demonstrates these time savings across all evaluated datasets, with total training times for \modelabbr{} being considerably shorter; in some cases, such as AG News and the GLUE average, training time reduced by more than half. This demonstrates that the \modelabbr{} method, incorporating DDS-MAE, offers significant gains in computational efficiency for effectively training the Qwen2.5-7B model across diverse datasets.

\subsubsection{Convergence Analysis}
\label{sssec:convergence}

\begin{figure}[bt!]
    \centering
    \includegraphics[scale=0.3]{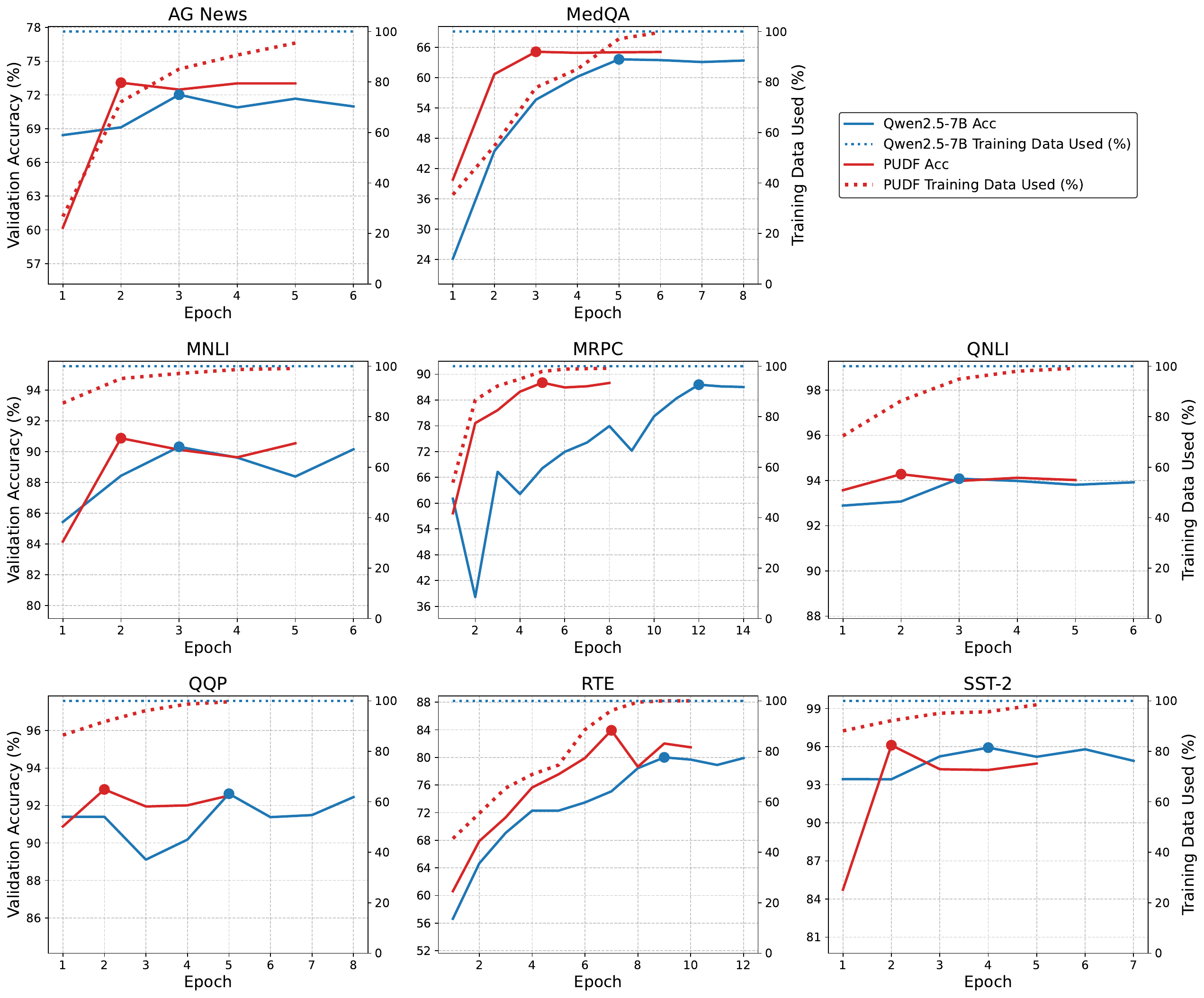} 
    \caption{Convergence analysis of the proposed \modelabbr{} against the Qwen2.5-7B baseline on AG News, MedQA, and the GLUE benchmark datasets. The solid lines represent validation accuracy, while the dotted lines indicate the percentage of training data utilized per epoch. Circular markers highlight the epoch with the best validation accuracy achieved by each model.}
    \label{fig:convergence}
\end{figure}

We next conduct a convergence analysis to evaluate DDS-MAE, our proposed TS component of \modelabbr{}, using the Qwen2.5-7B model as the foundational architecture.
The training dynamics of \modelabbr{} compared to a no-CL baseline are illustrated in Figure~\ref{fig:convergence}.
Several key observations emerge from these results:

\textbf{Convergence Speed.}
Across the evaluated datasets, \modelabbr{} demonstrates faster convergence and higher validation accuracy levels in fewer epochs compared to the no-CL baseline.
This trend is notably visible in MedQA, where \modelabbr{} shows a much steeper accuracy ascent in the initial epochs, consistent with results for AG News, MNLI, MRPC, and QNLI.
In contrast, the baseline model often requires more training epochs to reach comparable performance.
The rapid convergence characteristic of \modelabbr{} reinforces its potential for achieving strong results with reduced training iterations, thereby conserving computational resources.

\textbf{Accuracy Comparison.}
In terms of peak validation accuracy, indicated by circular markers in Figure~\ref{fig:convergence}, \modelabbr{} consistently matches or surpasses the performance of the baseline across the majority of the evaluated tasks.
Notably, \modelabbr{} achieves distinctly higher peak accuracy on MedQA, AG News, RTE, and MNLI.
On the remaining tasks (QQP, SST-2, MRPC, and QNLI), \modelabbr{} achieves slightly improved peak results.

\textbf{Data Efficiency.}
The percentage of training data utilized by \modelabbr{} during fine-tuning (represented by the red dotted line in Figure~\ref{fig:convergence}) offers critical insights into its data efficiency.
A consistent observation across most tasks, including AG News and MedQA, is that \modelabbr{} does not initially process the entire training dataset.
In the early epochs, typically around 40-60\% of the data is actively used (e.g., AG News, MedQA, MNLI, MRPC; SST-2 starts lower).
As the fine-tuning progresses and the model's capabilities enhance, \modelabbr{} progressively incorporates a larger fraction of the training data.
This gradual data introduction strategy aligns with the model's increasing capacity to effectively learn from a broader and potentially more complex range of examples.

\textbf{Data Usage and Strategic Scheduling.}
Analysis of the training trajectories reveals that \modelabbr{} frequently attains peak validation accuracy without the complete training dataset.
For instance, on MedQA, AG News, QNLI, and QQP, optimal performance is often reached when \modelabbr{} has utilized approximately 70-95\% of the available training data.
This highlights the model's proficiency in identifying and prioritizing the most informative examples for learning.
Furthermore, this behavior underscores the principle that not all training instances contribute equally or positively to model performance, particularly in the early stages.
As elaborated in our qualitative analysis (\S\ref{ssec:Quali}), some data points may be inherently more challenging or even contain labeling inaccuracies.
The DDS-MAE component of \modelabbr{} is designed to leverage information about example difficulty (e.g., as identified by IRT-AC) to strategically schedule more difficult or potentially noisy data points at later stages of training.
This approach mitigates their potential to disrupt learning when the model is less robust.
This curated data presentation contrasts with a more conventional or random ordering of training instances.

\textbf{Training Stability.}
The validation accuracy curves for \modelabbr{} generally exhibit greater stability after reaching peak performance, often with less pronounced fluctuations compared to the baseline on several tasks (e.g., MNLI, MRPC).
This suggests that \modelabbr{} may be less susceptible to overfitting, maintaining more consistent performance as training progresses.
We attribute this enhanced stability to the DDS-MAE mechanism, which dynamically selects and schedules training data based on an ongoing assessment of the model's learning state and data characteristics.
This adaptive approach can contribute to a more regularized and stable training process than methods employing a less informed or random data feeding strategy.

In summary, the DDS-MAE component integrated within \modelabbr{} offers notable advantages in terms of training efficiency, data utilization, and predictive performance across the diverse set of evaluated benchmark datasets, including AG News, MedQA, and tasks from GLUE.
It consistently demonstrates the capacity to achieve competitive or superior accuracy in most cases, often with fewer training epochs and by strategically utilizing subsets of the available training data.
Moreover, this approach tends to yield more stable training dynamics compared to the baseline.

\subsection{Extension of PUDF to Generative Tasks}
\label{ssec:gsm8k}

Thus far, we have demonstrated the ability of \modelabbr{} on several classification tasks. 
In this section, we show that \modelabbr{} can also handle novel text generation tasks.
Specifically, we apply \modelabbr{} to a difficult math question-answering dataset, GSM8K \citep{cobbe2021training}. 
To adapt to this dataset, we construct an artificial crowd using 11 SOTA large language models with varying prompting strategies. We collect responses from seven models via the Replicate API:\footnote{\url{https://replicate.com}} Claude 3.5 Sonnet~\cite{anthropic2024claude}, DeepSeek-V3~\cite{deepseekai2024deepseekv3}, Granite 3.3 8B Instruct~\cite{ibm2025granite}, Llama 3 8B Instruct~\cite{grattafiori2024llama}, Llama 3.1 405B Instruct~\cite{grattafiori2024llama}, GPT-4o-mini~\cite{openai2024gpt4o}, and GPT-5~\cite{openai2025gpt5}. Additionally, we employ four models through the Hugging Face API:\footnote{\url{https://huggingface.co}} Yi-1.5 9B Chat~\cite{young2024yi}, Gemma 2 9B IT~\cite{gemmateam2024gemma2}, Mistral 7B Instruct v0.2~\cite{jiang2023mistral}, and Qwen2.5 7B Instruct~\cite{yang2024qwen2}. For each model, we evaluate five prompting strategies: zero-shot, zero-shot with chain-of-thought (CoT) reasoning~\cite{wei2022chain}, 4-shot, 4-shot with CoT, and 8-shot with CoT. This yields a total of 55 experimental configurations (11 models $\times$ 5 strategies). For hyperparameter tuning and the hardware platform, we adopted the same settings used for Llama3.1-8B and Qwen2.5-7B on the MedQA and AG News datasets.

\begin{table}[!bht]
    \caption{%
        \label{tab:gsm8k_main}%
        Accuracy and training time results on GSM8K comparing PUDF with other 
        curriculum learning methods. Results are averaged over 5 runs with standard 
        deviations as subscripts. The best performing method for each model is in 
        \textbf{bold}; the second-best method is \underline{underlined}.%
    }
    \centering
    \small
    \begin{tabular}[t]{llll}
        \toprule
        \textbf{Model}      & 
        \textbf{Method}     & 
        \textbf{Accuracy (\%)} & 
        \textbf{Time (mins)} 
        \\
        \midrule
        Llama3.1-8B         & Baseline   & $59.28_{\pm 0.81}^{*}$             & $108.41_{\pm 7.44}^{*}$ \\
                            & d\_SL-L    & $57.39_{\pm 1.01}^{*}$             & $\mathbf{52.10}_{\pm 2.89}$ \\
                            & d\_SL-R    & $59.27_{\pm 0.68}^{*}$             & $62.38_{\pm 1.56}$ \\
                            & d\_WR-L    & $58.45_{\pm 0.66}^{*}$             & $55.97_{\pm 1.50}$ \\
                            & d\_WR-R    & $58.56_{\pm 0.89}^{*}$             & $57.49_{\pm 2.85}$ \\
                            & SPL        & $60.06_{\pm 0.77}^{*}$             & $95.89_{\pm 2.62}^{*}$ \\
                            & TT         & $60.64_{\pm 0.72}^{*}$             & $115.69_{\pm 9.98}^{*}$ \\
                            & RL         & $\underline{60.99}_{\pm 0.64}^{*}$ & $145.24_{\pm 5.04}^{*}$ \\
                            & PUDF       & $\mathbf{61.72}_{\pm 0.44}$        & $\underline{78.66}_{\pm 1.71}^{*}$ \\
        \cmidrule{1-4}
        
        Qwen2.5-7B          & Baseline   & $74.10_{\pm 0.73}^{*}$             & $146.59_{\pm 5.81}^{*}$ \\
                            & d\_SL-L    & $67.88_{\pm 1.71}^{*}$             & $\mathbf{63.14}_{\pm 1.30}$ \\
                            & d\_SL-R    & $70.85_{\pm 0.76}^{*}$             & $73.29_{\pm 1.75}$ \\
                            & d\_WR-L    & $73.89_{\pm 0.78}^{*}$             & $90.83_{\pm 1.89}^{*}$ \\
                            & d\_WR-R    & $72.76_{\pm 0.58}^{*}$             & $83.17_{\pm 1.62}$ \\
                            & SPL        & $74.15_{\pm 0.76}^{*}$             & $128.83_{\pm 2.50}^{*}$ \\
                            & TT         & $75.89_{\pm 0.99}$                 & $137.24_{\pm 3.16}^{*}$ \\
                            & RL         & $\underline{76.18}_{\pm 0.61}$     & $177.16_{\pm 2.61}^{*}$ \\
                            & PUDF       & $\mathbf{76.70}_{\pm 0.37}$        & $\underline{85.05}_{\pm 1.68}^{*}$ \\
        \bottomrule
    \end{tabular}
    
    \vspace{2mm}
    {\raggedright \footnotesize 
        $^*$For accuracy: indicates significantly lower than the \textbf{best accuracy} 
        for that model. For training time: indicates significantly higher than the 
        \textbf{fastest time} for that model (Welch's one-tailed t-test with Benjamini-Hochberg correction, $\alpha < 0.05$).
    }
\end{table}

\begin{table}[!bht]
    \caption{%
        \label{tab:gsm8k_ablation}%
        Results of ablation study on GSM8K for Qwen2.5-7B. DM: Difficulty Metric; 
        TS: Training Schedule. The best performing variant is in \textbf{bold}; 
        the second-best variant is \underline{underlined}.%
    }
    \centering
    \small
    \begin{tabular}[t]{p{1.75cm}lll}
        \toprule
        \textbf{Metric}     & 
        \textbf{DM}         & 
        \textbf{TS}         & 
        \textbf{GSM8K} 
        \\
        \midrule
        
        \multirow{5}{*}{Accuracy (\%)} 
                            & d\_SL      & DDS-MAE        & $\underline{73.89}_{\pm 0.80}^{*}$ \\
                            & d\_WR      & DDS-MAE        & ${73.68}_{\pm 0.48}^{*}$ \\
                            & IRT-AC     & Root           & $68.95_{\pm 1.41}^{*}$ \\
                            & IRT-AC     & Linear         & $71.52_{\pm 1.20}^{*}$ \\
                            & IRT-AC     & DDS-MAE (PUDF) & $\mathbf{76.70}_{\pm 0.37}$ \\
        \midrule
        
        \multirow{5}{*}{\parbox{1.75cm}{Training Time\newline (minutes)}} 
                            & d\_SL      & DDS-MAE        & $75.32_{\pm 1.90}^{*}$ \\
                            & d\_WR      & DDS-MAE        & ${71.71}_{\pm 1.61}^{*}$ \\
                            & IRT-AC     & Root           & $\underline{57.96}_{\pm 1.91}^{*}$ \\
                            & IRT-AC     & Linear         & $\mathbf{48.64}_{\pm 2.19}$ \\
                            & IRT-AC     & DDS-MAE (PUDF) & $85.05_{\pm 1.68}^{*}$ \\
        \bottomrule
    \end{tabular}
    
    \vspace{2mm}
    {\raggedright \footnotesize 
        $^*$For accuracy: indicates significantly lower than the \textbf{best accuracy}. 
        For training time: indicates significantly higher than the \textbf{fastest time} 
        (Welch's one-tailed t-test, $\alpha < 0.05$).
    }
\end{table}

Table~\ref{tab:gsm8k_main} demonstrates that \modelabbr{} successfully extends to generative mathematical reasoning tasks, achieving the best performance on both model architectures. For Llama3.1-8B, \modelabbr{} attains 61.72\% accuracy, significantly outperforming all baselines, including the strongest competitor, RL (60.99\%). On Qwen2.5-7B, \modelabbr{} achieves the highest numerical accuracy at 76.70\%, though the improvement over strong competitors RL (76.18\%) and TT (75.89\%) is not statistically significant ($p \approx 0.07$). However, in both cases, \modelabbr{} is significantly faster than these high-performing competitors (RL and TT).
Notably, heuristic-based CL methods (d\_SL-L, d\_SL-R, d\_WR-L, d\_WR-R) consistently underperform compared to learnable difficulty estimation approaches, with d\_SL-L achieving only 57.39\% on Llama3.1-8B and 67.88\% on Qwen2.5-7B. These results validate that \modelabbr{}'s learnable IRT-based difficulty metric, combined with adaptive pacing through DDS-MAE, effectively captures the complexity of mathematical reasoning tasks where human-annotated difficulty signals are unavailable. Furthermore, \modelabbr{} demonstrates competitive training efficiency, requiring substantially less time than RL (145.24 and 177.16 minutes for the two models) while maintaining superior or competitive accuracy.

The ablation study in Table~\ref{tab:gsm8k_ablation} reveals that both the difficulty metric and training schedule components are essential to \modelabbr{}'s success. When using the same DDS-MAE schedule but replacing IRT-AC with deterministic metrics, d\_SL + DDS-MAE achieves 73.89\% and d\_WR + DDS-MAE reaches 73.68\%, both significantly lower than \modelabbr{}'s 76.70\%. Conversely, maintaining the IRT-AC difficulty metric but substituting DDS-MAE with simpler schedules (Root or Linear pacing) yields even worse results of 68.95\% and 71.52\%, respectively.
Notably, these results are worse than those achieved by simple heuristic methods in Table~\ref{tab:gsm8k_main} (e.g., $d_{WR}$-L at 73.89\%), confirming that a naive application of the IRT-AC metric without its co-designed DDS-MAE scheduler can actually be detrimental to performance.
The training time analysis further illuminates this trade-off: while the Linear schedule completes fastest at 48.64 minutes, it scores 5.18 percentage points lower in accuracy compared to \modelabbr{}. These findings underscore that \modelabbr{}'s effectiveness stems from the synergistic integration of learnable difficulty assessment and dynamic data scheduling, with neither component alone sufficient to achieve optimal performance on complex generative reasoning tasks.

\section{Further Analyses: Exploring the IRT-AC}
\label{sec:exploration}

This section presents an in-depth analysis of IRT-AC, which estimates the difficulty value for each data instance.
We look at the properties of the learned example difficulties to demonstrate their calibration with expected results and further demonstrate IRT-AC as a difficulty estimation mechanism with potential benefits independent of \modelabbr{}.


\subsection{Distribution of Difficulty}
\label{sssec:diffdist}

\begin{figure}[ht]
    \centering
    \includegraphics[trim={0.18in 0.1in 0.1in 0.02in}, clip, scale=0.22]{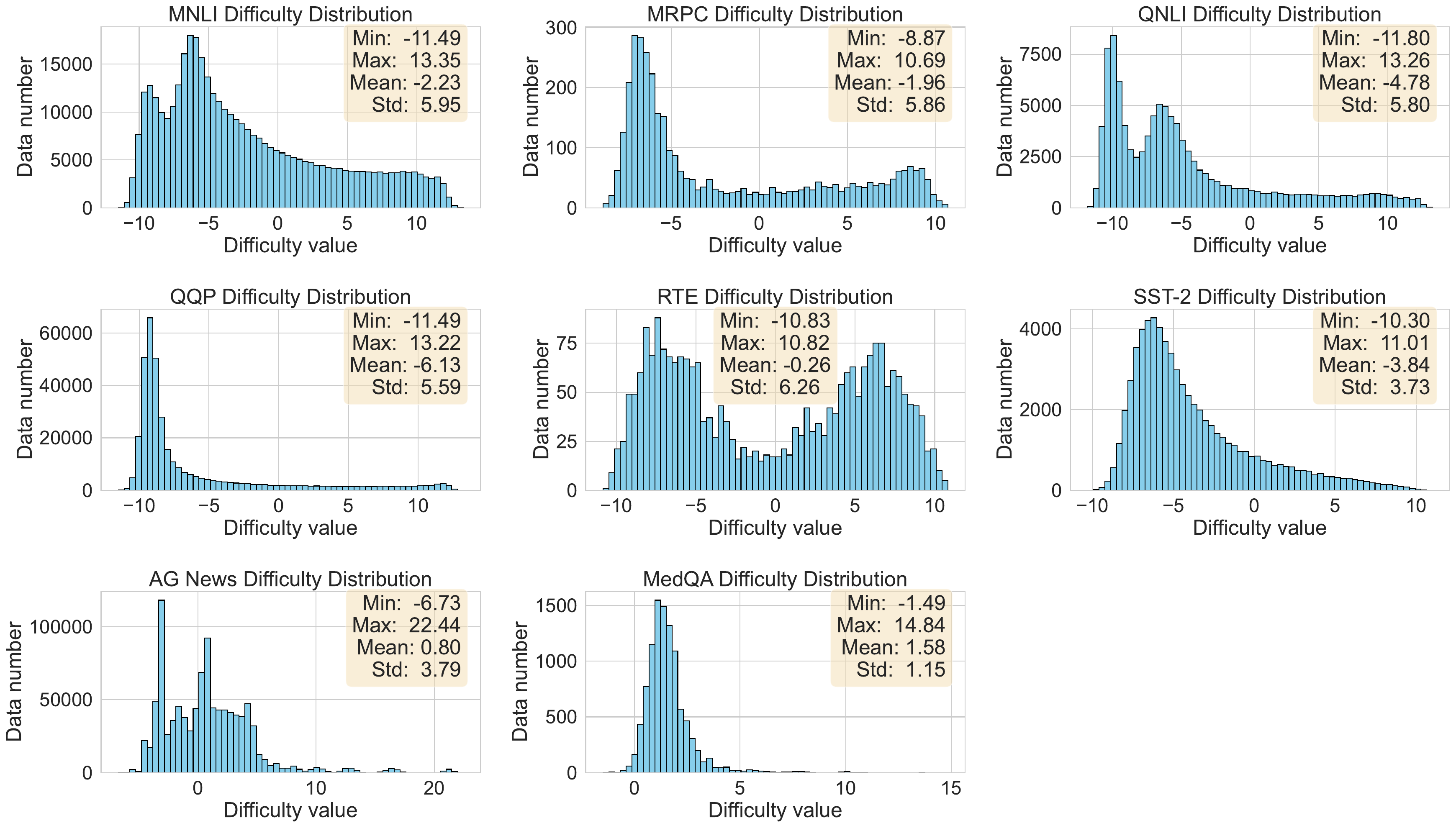}
    \caption{IRT-AC generated difficulty distributions for the GLUE benchmark, AG News, and MedQA datasets.}
    \label{fig:difficulty_distributions}
\end{figure}

Figure~\ref{fig:difficulty_distributions} displays the difficulty distributions estimated by our IRT-AC model for instances across the MedQA, AG News, and GLUE benchmark datasets. A notable characteristic across these diverse datasets is that the difficulty scores generally form distributions approximating a Gaussian profile. This observation suggests that a majority of instances in these benchmarks tend to cluster around a central difficulty level, with fewer examples at the extremes of being excessively easy or prohibitively challenging.
Analyzing the mean difficulty values provides insights into the relative challenge posed by each dataset. For instance, datasets such as QQP (mean: -6.13), QNLI (mean: -4.78), and SST-2 (mean: -3.84) have low mean difficulties, indicating they are, on average, less challenging. 
Conversely, datasets like MedQA (mean: 1.58), AG News (mean: 0.80), and RTE (mean: -0.26) present higher mean difficulty values, suggesting that they are comparatively more challenging.
Crucially, these IRT-AC derived difficulty metrics show a strong correspondence with empirical model performance reported in Table~\ref{tab:mainResult} (and Appendix Table \ref{tab:appendixResult}). 
Tasks with lower mean difficulty scores consistently achieve higher accuracies, whereas those identified as more difficult (e.g., MedQA, AG News, RTE) tend to yield lower accuracy scores. 
This observed trend between our estimated difficulties and actual model outcomes further supports the reliability and validity of the IRT-AC framework in quantifying task and instance-level challenge.

\subsection{Qualitative Analysis of IRT-AC Difficulty Scores}
\label{ssec:Quali}

\begin{table}[!b]
  \centering
  \footnotesize
  \caption{Top 5 hardest and easiest questions from the MedQA dataset. Correct answer for each is \textbf{in bold}.}
  \label{tab:medqa_case_study}

  \begin{tabular}{p{\textwidth}} 
  \toprule
    \bf (a) Top 5 Hardest Questions \\
    \midrule
      Q: A 35-year-old man is brought to the emergency department by his wife because of a 1-week history of progressive confusion, myalgia, and nausea. His wife says that he first reported headaches and fatigue 10 days ago, and since then ``he has not been himself.'' He has refused to drink any liquids for the last day. Two months ago, he helped his neighbor remove a raccoon’s den from her backyard. He appears agitated. His temperature is 100.8°F (38.2°C). Examination shows excessive drooling. Muscle tone and deep tendon reflexes are increased bilaterally. Administration of which of the following is most likely to have prevented this patient’s condition?\newline
        A) RNA-dependent DNA polymerase inhibitor~
        \textbf{B) Chemically-inactivated virus}\newline
        C) Live attenuated vaccine~
        D) Immunoglobulin against a bacterial protein 
    \\[0.8em]
    \midrule
    
      Q: Physical exam of a 15-year-old female reveals impetigo around her mouth. A sample of the pus is taken and cultured. Growth reveals gram-positive cocci in chains that are bacitracin sensitive. Which of the following symptoms would be concerning for a serious sequelae of this skin infection? \newline
        A) Fever~
        B) Myocarditis~
        \textbf{C) Hematuria}~
        D) Chorea
    \\[0.8em]
    \midrule
    
      Q: A 40-year-old man comes to the physician for a follow-up examination. He feels well. He has no urinary urgency, increased frequency, dysuria, or gross hematuria. He has a history of recurrent urinary tract infections; his last UTI was treated with ciprofloxacin 3 months ago. Exam and labs are unremarkable except trace blood on UA. Cystoscopy is normal. Which of the following is the most appropriate next step in management?\newline
        A) Transrectal ultrasound~
        B) Voided urine cytology~
        C) Reassurance~
        \textbf{D) CT urography}
    \\[0.8em]
    \midrule
      Q: A 4-week-old boy is brought to the ED with a 2-day history of projectile vomiting after feeding. Parents report normal development until 1 week ago when he began to eat less. Exam: palpable “olive” in RUQ, non-bilious vomiting. Which of the following is associated with the most likely cause of this patient’s symptoms?\newline
        A) Chloride transport defect~
        B) Failure of neural crest migration\newline
        \textbf{C) Nitric oxide synthase deficiency}~
        D) Recanalization defect
    \\[0.8em]
    \midrule
      Q: A 54-year-old man presents with 4 months of foul-smelling diarrhea, weight loss, and fatigue after returning from Bangladesh. Labs: macrocytic anemia, fecal fat 22 g/day, TTG-IgA negative, stool studies negative. What is the most appropriate next step in diagnosis?\newline
        A) CT scan of the abdomen~
        B) Schilling test~
        \textbf{C) Enteroscopy}~
        D) PAS-stained biopsy of small bowel
    \\
    \bottomrule
  \end{tabular}

  \vspace{1em}

  \begin{tabular}{p{\textwidth}} 
    \toprule
    \bf (b) Top 5 Easiest Questions \\
    \midrule
      Q: A 28-year-old Caucasian woman presents to a local walk-in clinic with pruritus and a salmon-colored scaling patch on her back. She had a cold two weeks ago and her lesion has enlarged. Exam: generalized exanthem with collarette scale. What is the best next step in management?\newline
        \textbf{A) Pruritus control and reassurance}~
        B) Systemic steroid therapy~
        C) Topical steroid therapy~
        D) Phototherapy
    \\[0.8em]
    \midrule
      Q: A 16-year-old man after a camping trip has 5 days of flatulence, nausea, and greasy, foul-smelling diarrhea. No blood or urgency. Exam: mild diffuse abdominal tenderness. What is he most likely to report about his camping activities?\newline
        \textbf{A) Collecting water from a stream, without boiling or chemical treatment}~
        B) This has been going on for months~
        C) Camped as a side excursion from a cruise ship~
        D) Camped in Mexico
    \\[0.8em]
    \midrule
      Q: A 21-year-old female college student is brought by roommates who note hair-pulling and biopsy shows traumatic alopecia. What is the single most appropriate treatment?\newline
        \textbf{A) Cognitive-behavior therapy or behavior modification}~
        B) Clomipramine~
        C) Venlafaxine~
        D) Electroconvulsive therapy
    \\[0.8em]
    \midrule
      Q: A 55-year-old woman with type 1 diabetes has 3 months of urinary dribbling and elevated post-void residual. Which intervention is most likely to benefit?\newline
        \textbf{A) Intermittent catheterization}~
        B) Amitriptyline therapy~
        C) Prazosin therapy~
        D) Oxybutynin therapy
    \\[0.8em]
    \midrule
      Q: A 6-year-old girl with 3 days of malaise, mouth sores, and vesicles on hands and feet. What is the next best step in management?\newline
        \textbf{A) Supportive care}~
        B) Aspirin~
        C) Corticosteroids~
        D) Penicillin
    \\
    \bottomrule
  \end{tabular}
\end{table}

To provide qualitative insights into the difficulty scores from our IRT-AC Difficulty Model, this section analyzes selected examples. We focus on the MedQA dataset, presenting the top five hardest and easiest questions as determined by our model (Table \ref{tab:medqa_case_study}). 
For broader context, examples from the GLUE benchmark and AG News datasets are available in Appendix \ref{ssec:easyHardExamples}. 
The questions identified as ``hardest'' by IRT-AC (Table \ref{tab:medqa_case_study}a) generally require sophisticated reasoning and specialized knowledge. 
Many require multi-step inference, such as deducing a condition and then recalling a specific preventative measure or underlying cause from highly technical options (e.g., Questions 1, 4 in Table \ref{tab:medqa_case_study}a). 
Others involve complex clinical decision-making regarding optimal diagnostic or management steps from nuanced alternatives (e.g., Questions 3, 5). 
These questions often present intricate scenarios and necessitate fine discrimination among medically specific options.
Conversely, questions rated ``easiest'' (Table \ref{tab:medqa_case_study}b) typically feature more straightforward scenarios with distinct clinical cues for common conditions (e.g., Questions 1, 5 in Table \ref{tab:medqa_case_study}b), leading to relatively direct conclusions about management or treatment. 
The correct answers frequently align with general medical principles or clearly address the primary issue, while distractors often appear less plausible, thereby reducing the need for deep, specialized knowledge or complex inferential chains.
These qualitative observations suggest that the IRT-AC model effectively discerns varying levels of question complexity, associating higher difficulty scores with tasks requiring more specialized knowledge, multi-step reasoning, and finer discrimination among options.
This also highlights the disparity between examples in a particular dataset and further strengthens the conceptual motivation for CL generally and \modelabbr{} specifically.

\begin{figure}[ht]
    \centering
    \includegraphics[scale=0.31]{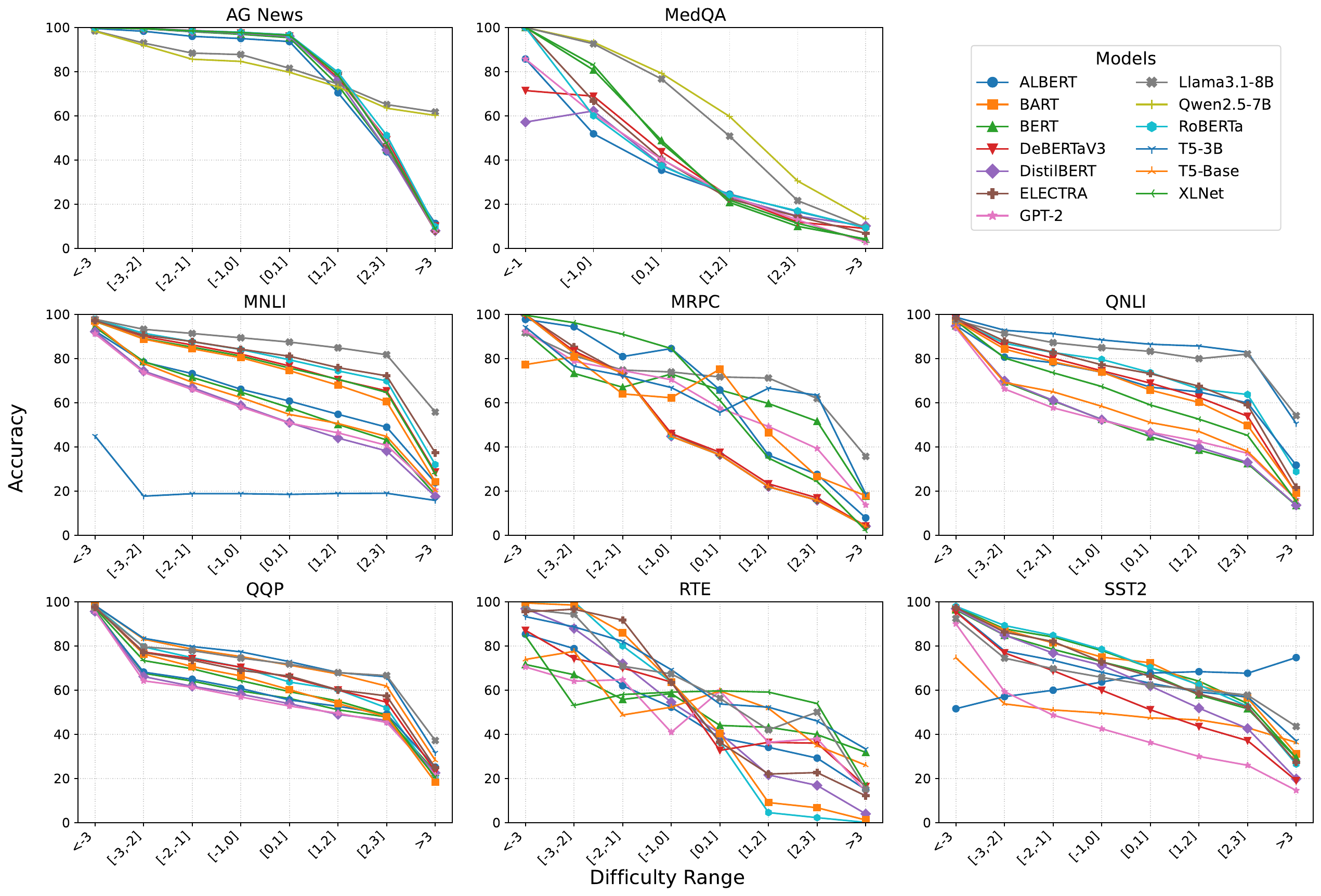} 
    \caption{Accuracy of diverse language models (comprising the Artificial Crowd) across IRT-AC difficulty bins for GLUE, AG News, and MedQA datasets. The legend details the specific LLMs utilized.}
    \label{fig:accuracy_across_difficulty_bins}
\end{figure}

\subsection{Artificial Crowd Accuracy Across Difficulty Bins}
\label{sssec:accDiffBins}

This section examines the relationship between IRT-AC derived example difficulty and the empirical performance of a diverse set of LLMs. These LLM crowd models provide multiple perspectives on how accuracy varies with data difficulty. 
Figure~\ref{fig:accuracy_across_difficulty_bins} illustrates the accuracy of each LLM within this crowd across binned difficulty levels, ranging from easiest (difficulty $< -3$) to hardest (difficulty $> 3$), for the GLUE, AG News, and MedQA datasets.
We observe a consistent inverse correlation between example difficulty and model accuracy; as the IRT-AC assessed difficulty of data instances increases, the evaluation accuracy of the LLMs systematically decreases. 
This trend holds across all examined datasets, demonstrating the link between higher difficulty scores and lower empirical success rates. 
This consistent behavior across a spectrum of models and data types strongly validates the IRT-AC scores, confirming their efficacy in capturing meaningful signal of task and instance-level challenge that directly predict model performance.

\begin{figure}[!bht]
    \centering
    
    \includegraphics[scale=0.27]{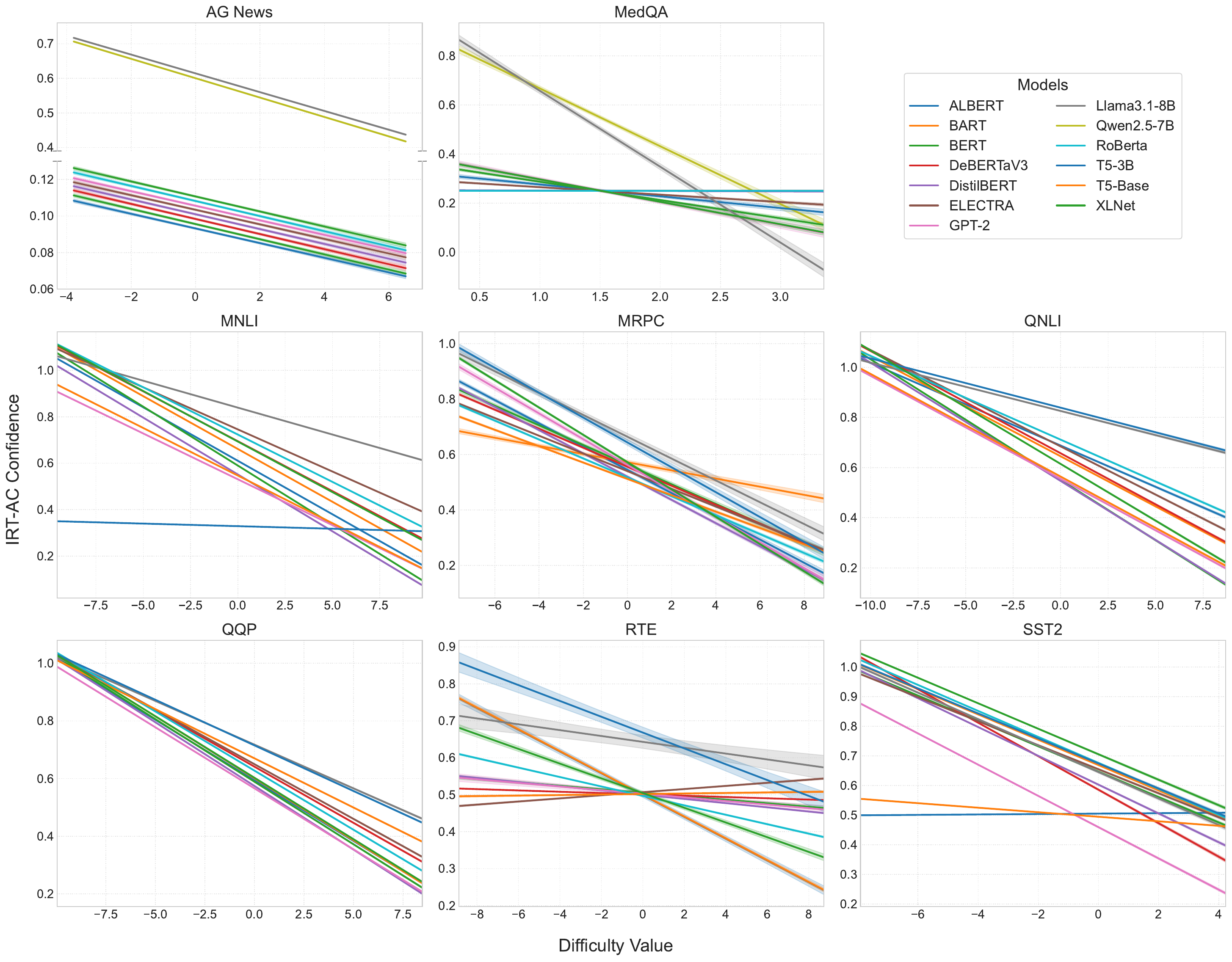}
    \caption{Model confidence in relation to IRT-AC example difficulty across GLUE, AG News, and MedQA datasets. Confidence is defined as the model's output probability for the correct label.}
    \label{fig:irt_ac_confidence}
\end{figure}

\subsection{Model Confidence in Relation to Example Difficulty}
\label{sssec:confGenDiff}

This analysis investigates model confidence, quantified as the probability assigned by each model to the true label, relative to IRT-AC derived example difficulty (Figure~\ref{fig:irt_ac_confidence}). 
We use Ordinary Least Squares (OLS) regression to estimate the relationship between difficulty and model confidence, focusing on examples within the 5th to 95th difficulty percentile for each task across the GLUE, AG News, and MedQA datasets. 
The predominant finding across most datasets and models is an inverse correlation: model confidence generally decreases as the IRT-AC assessed difficulty of instances increases. 
This suggests a meaningful alignment between our difficulty metric and the models' internal assessments of certainty.

While this inverse relationship is broadly consistent, Figure~\ref{fig:irt_ac_confidence} also reveals notable dataset- and model-specific variations. 
On AG News, for instance, larger and more recent models such as Llama3.1-8B and Qwen2.5-7B tend to maintain markedly higher confidence in the correct label across the difficulty spectrum compared to several other LLMs. 
This sustained confidence could be attributed to their enhanced representational power and potentially better calibration on this news classification task. Conversely, MedQA typically shows a steeper confidence decline for most models, underscoring its inherent challenge. 
Other datasets, like RTE, exhibit more varied confidence trends, with some models deviating from a clear downward slope, possibly reflecting task-specific complexities or differences in model calibration. 
Despite these variations, the overarching tendency of diminishing confidence on examples identified as harder by IRT-AC further substantiates the validity of our difficulty scores as indicators of task and instance-level challenge.

\subsection{AC Ablation}
\label{ssec:ac_ablation}

In this section, we investigate the importance of the construction of the IRT-AC and its effect on estimated difficulties.
Specifically, we randomly sample a subsection of our AC models and re-fit our IRT models to determine whether we can 
recover the learned difficulty parameters with a smaller crowd size. 
We calculate Pearson and Spearman correlations between difficulty estimates from our sampled crowd and difficulty estimates from our full IRT-AC (Table \ref{tab:ac_ablation}.
With only 5 models, correlations vary widely, indicating that there is insufficient information in the response patterns to accurately estimate the latent difficulty parameters. 
As the crowd size increases, the correlations improve. 
With a crowd size of 30, correlations are consistently high. 
These results indicate the importance of collecting enough data from a variety of models for IRT-AC. 
The estimated difficulty and ability parameters are population dependent, so ensuring a representative sample of models will improve IRT-AC estimations. 
If needed, the IRT-AC can be updated with response patterns from new SOTA models to reflect the updated state of overall model performance. 

\begin{table}[ht]
\centering
\small
\caption{Pearson and Spearman correlation coefficients when comparing IRT difficulty values as estimated from a subsection of AC models to the full AC estimates. All correlations are significantly different than 0 ($p < 0.01$).}
\label{tab:ac_ablation}

\begin{tabular}{lrrrrrrrrr}
  \toprule
  & \bf AC & \bf AG News & \bf MedQA & \bf MNLI & \bf MRPC & \bf QNLI & \bf QQP & \bf RTE & \bf SST-2 \\ 
  \midrule
 \multirow{4}{*}{Pearson} & 5 & 0.70 & 0.40 & 0.53 & 0.79 & 0.63 & 0.45 & 0.71 & 0.29 \\ 
    & 15 & 0.72 & 0.56 & 0.72 & 0.93 & 0.72 & 0.85 & 0.70 & 0.71 \\ 
    & 30 & 0.93 & 0.82 & 0.92 & 0.93 & 0.91 & 0.83 & 0.82 & 0.88 \\ 
    & 45 & 0.97 & 0.91 & 0.97 & 0.97 & 0.95 & 0.92 & 0.97 & 0.90 \\ 
    \cmidrule{2-10}
   \multirow{4}{*}{Spearman} & 5 & 0.73 & 0.48 & 0.72 & 0.78 & 0.64 & 0.46 & 0.86 & 0.43 \\ 
    & 15 & 0.95 & 0.78 & 0.94 & 0.87 & 0.89 & 0.67 & 0.93 & 0.66 \\ 
    & 30 & 0.97 & 0.91 & 0.96 & 0.88 & 0.94 & 0.84 & 0.97 & 0.87 \\ 
    & 45 & 0.98 & 0.94 & 0.97 & 0.92 & 0.97 & 0.81 & 0.97 & 0.87 \\ 
   \bottomrule
\end{tabular}
\end{table}

\section{Limitations}
\label{sec:limitations}

This work is not without limitations. 
One potential issue with \modelabbr~is the chance of a high variance model, due to the additional step of estimating model ability during training.
However, in our results, we find that variance in terms of output performance is low for \modelabbr~(\S\ref{ssec:LM models}).
We can infer that the ability estimation process is relatively stable.
That is, the example difficulties estimated from IRT are stable enough that ability estimates align with the current state of the model, as indicated by the regular progression through the curriculum and increasing training and validation accuracy performance. 
Our results show that adding this step does not lead to a higher variance model; in certain cases, \modelabbr~has lower variance than the baseline and competence-based frameworks.

For \modelabbr, there is a potentially significant cost associated with estimating $\theta_e$. 
Estimating $\theta_e$ requires an additional forward pass through the training dataset to gather the labels for scoring as well as MLE estimation.
For large datasets, this can effectively double the number of forward passes during training.
To alleviate the extra cost, we sample from the training set before our first epoch, and use this down-sampled subset as our ability estimation set.
As most examples have difficulty values between $-3$ and $3$, the full training set isn't necessary for estimating $\theta_e$.
Identifying the optimal number of examples needed to estimate ability is left for future work.

Another computational cost to \modelabbr{} involves IRT-AC, specifically response pattern generation. 
Collecting response patterns needed for difficulty estimation involves fine-tuning multiple LLMs, which can be costly in terms of runtime. 
For example, as shown in Table \ref{tab:pudf_pipeline_time_comparison}, IRT-AC computation time accounts for 12.7\% to 33.2\% of the total \modelabbr{} runtime. 
While the total runtime is less than a standard training runtime, this cost is not negligible, especially when considering larger, computationally expensive models being used. 
However, we do note several mitigating factors for future implementations to offset some of these costs.
First, our analyses show that these costs can be reduced by fine-tuning models in parallel.
Second, pre-trained models can be used without further fine-tuning; we can directly run inference for less costly response pattern collection.
For example, future work looking at inference time techniques such as few-shot learning and chain-of-thought for IRT-AC may find methods that reduce the computation burden further.
Third, the large number of leaderboards that are available for tracking LLM progress are a potential source of response pattern data for IRT-AC in new domains~\citep{rodriguez2021evaluation}.
Lastly, IRT-AC is not required for every run of \modelabbr. 
Once the difficulties have been estimated via IRT-AC, those learned values are valid for training any subsequent model using \modelabbr.
Therefore, the IRT-AC cost can be amortized across multiple fine-tuning runs for different models, making the overall computational burden less. 
This can also act as an encouragement to future researchers to record and make available instance-level response patterns or learned difficulty parameters so that there is a shared pool of responses for difficulty estimation.
For example, the IRT-AC response patterns for our benchmarking datasets\footnote{\url{https://huggingface.co/datasets/lalor/response-patterns}} can be a foundation for future work where new models are added to the AC and novel fine-tuning strategies leverage the pre-existing AC. 
Relatedly, the size and composition of IRT-AC should be sufficiently large and diverse to ensure that the learned difficulty parameters reflect variations in responses due to differences in latent ability. 
Otherwise, responses may be too homogeneous and therefore cannot capture the variation needed to ensure accurate difficulty estimations and appropriate scheduling in \modelabbr{}.

\section{Conclusion}
\label{sec:conclusion}

In this paper, we introduce \modelabbr{}, a novel \modelname{}.  
By combining IRT-AC for data difficulty measurement and DDS-MAE for dynamic training scheduling, \modelabbr{} offers a theoretically grounded and automated approach to CL. 
Our extensive experiments on a robust benchmark of datasets and comparison CL methods demonstrate that \modelabbr{} consistently improves both accuracy and training efficiency across multiple pre-trained language models and tasks, outperforming SOTA CL methods. 
The success of \modelabbr{} opens up promising directions for future research, including applications in other domains, such as computer vision and multimodal domains.

This work validates and supports the existing literature on curriculum learning.
Our results confirm that curriculum learning frameworks for supervised learning can lead to faster convergence or better local minima, as measured by test set performance \citep{bengio_curriculum_2009}.
We have shown that by replacing a heuristic for difficulty with a theoretically-based, learned difficulty value for training examples, static curriculum learning frameworks can be improved.
Probing the model's ability allows for data to be selected for training that is appropriate for the model and is not rigidly tied to a heuristic schedule.

By using \modelabbr,~a curriculum can adapt during training according to the estimated ability of the model.
\modelabbr~adds or removes training data based not on a fixed step schedule but rather by probing the model at each epoch and using the estimated ability to match data to the model.
This way, if a model has a high estimated ability early in training, then more data can be added to the training set more quickly, and learning isn't artificially slowed down due to the curriculum schedule.
It also allows for the possibility of a smaller dataset at later stages, if model performance decreases.

The \modelabbr{} framework significantly advances the state of CL and its application to NLP. 
By incorporating psychological principles through IRT and leveraging dynamic data selection strategies, \modelabbr{} offers a theoretically robust and adaptable approach to CL. 
This framework improves traditional heuristic-based methods and current CL methods, providing a explainable and modular system for dynamically aligning training data with the evolving capabilities of the model. 
\modelabbr{} demonstrates its effectiveness by optimizing the fine-tuning process for LLMs, improving performance metrics such as accuracy and training time across a range of tasks. 
Its dynamic scheduling mechanism reduces the reliance on static curriculum schedules, facilitating more efficient training without imposing additional computational overheads. 
\modelabbr{} can facilitate advancing CL research and enhancing the practical deployment of pre-trained language models in diverse applications~\citep{yang2022fpc,chaudhury2024dacl,liu2024curriculum}.

There are several avenues for future work.
Even though it is dynamic, \modelabbr~employs a simple curriculum schedule: only include examples where difficulty is less than or equal to estimated ability.
However, being able to estimate ability on the fly with \modelabbr~suggests the following research question: what is the best way to build a curriculum, knowing example difficulty and model ability?
It may be the case that only data with difficulty within a range of ability (higher and lower) is better, or that the training set shifts as the model improves.
Future research could also investigate the applicability of the 85\% rule of~\citep{Wilson255182} for curriculum design in LLMs.


\modelabbr~can also be adapted to more traditional information retrieval tasks, such as learning to rank and online judging for training high-ability systems and ordering examples according to learned difficulty. 
In particular, with a 1PL IRT model, the intuitive link between $\theta$ and $b$ allows for inherently explainable training mechanisms.
An example is only included in training if its difficulty is lower than the model's estimated ability at that point in time. 
This can be easily explained to model stakeholders and compared with standardized tests for humans, where questions are selected based on human-estimated ability.


\begin{acknowledgments}
	The authors would like to thank Hao Wu and Hadi Amiri for their helpful conversations with regards to this work. 
	This work was supported in part by LM012817 from the National Institutes of Health, I01HX003969 from VA Health Systems Research, and IIS-2403438 from the National Science Foundation. 
	This work was also supported in part by the Center for Intelligent Information Retrieval at UMass Amherst, and the	Center for Research Computing, the Human-centered Analytics Lab, and the Mendoza College of Business at the University of Notre Dame.
	The contents of this paper do not represent the views of CIIR, NIH, NSF, VA, the University of Notre Dame, the University of Massachusetts, or the United States Government.
\end{acknowledgments}

\bibliography{ref}

@inproceedings{lalor-yu-2020-dynamic,
    title = "Dynamic Data Selection for Curriculum Learning via Ability Estimation",
    author = "Lalor, John P.  and
      Yu, Hong",
    booktitle = "Findings of the Association for Computational Linguistics: EMNLP 2020",
    month = nov,
    year = "2020",
    address = "Online",
    publisher = "Association for Computational Linguistics",
    url = "https://www.aclweb.org/anthology/2020.findings-emnlp.48",
    doi = "10.18653/v1/2020.findings-emnlp.48",
    pages = "545--555",
}

@inproceedings{cook2025no,
  title={No Simple Answer to Data Complexity: An Examination of Instance-Level Complexity Metrics for Classification Tasks},
  author={Cook, Ryan A and Lalor, John P and Abbasi, Ahmed},
  booktitle={Proceedings of the 2025 Conference of the Nations of the Americas Chapter of the Association for Computational Linguistics: Human Language Technologies (Volume 1: Long Papers)},
  pages={2553--2573},
  year={2025}
}

@misc{Iyer_Dandekar_Csernai_2017, 
	title={First Quora Dataset Release: Question Pairs}, 
	url={https://quoradata.quora.com/First-Quora-Dataset-Release-Question-Pairs}, 
	journal={Data at Quora}, 
	author={Iyer, Shankar and Dandekar, Nikhil and Csernai, Kornél}, 
	year={2017}
}

@article{ormerod2024kitchen,
  title={How is a “kitchen chair” like a “farm horse”? Exploring the representation of noun-noun compound semantics in transformer-based language models},
  author={Ormerod, Mark and del Rinc{\'o}n, Jes{\'u}s Mart{\'\i}nez and Devereux, Barry},
  journal={Computational Linguistics},
  volume={50},
  number={1},
  pages={49--81},
  year={2024},
  publisher={MIT Press One Broadway, 12th Floor, Cambridge, Massachusetts 02142, USA~…}
}

@article{benjamini1995controlling,
  title={Controlling the false discovery rate: a practical and powerful approach to multiple testing},
  author={Benjamini, Yoav and Hochberg, Yosef},
  journal={Journal of the Royal statistical society: series B (Methodological)},
  volume={57},
  number={1},
  pages={289--300},
  year={1995},
  publisher={Wiley Online Library}
}

@inproceedings{yang2022fpc,
  title={Fpc: fine-tuning with prompt curriculum for relation extraction},
  author={Yang, Sicheng and Song, Dandan},
  booktitle={Proceedings of the 2nd Conference of the Asia-Pacific Chapter of the Association for Computational Linguistics and the 12th International Joint Conference on Natural Language Processing (Volume 1: Long Papers)},
  pages={1065--1077},
  year={2022}
}

@inproceedings{chaudhury2024dacl,
  title={DACL: Disfluency augmented curriculum learning for fluent text generation},
  author={Chaudhury, Rohan and Teleki, Maria and Dong, Xiangjue and Caverlee, James},
  booktitle={Proceedings of the 2024 Joint International Conference on Computational Linguistics, Language Resources and Evaluation (LREC-COLING 2024)},
  pages={4311--4321},
  year={2024}
}

@inproceedings{liu2024curriculum,
  title={Curriculum Consistency Learning for Conditional Sentence Generation},
  author={Liu, Liangxin and Liu, Xuebo and Lian, Lian and Cheng, Shengjun and Rao, Jun and Yu, Tengfei and Deng, Hexuan and Zhang, Min},
  booktitle={Proceedings of the 2024 Conference on Empirical Methods in Natural Language Processing},
  pages={13865--13881},
  year={2024}
}

@inproceedings{swayamdipta2020dataset,
  title={Dataset Cartography: Mapping and Diagnosing Datasets with Training Dynamics},
  author={Swayamdipta, Swabha and Schwartz, Roy and Lourie, Nicholas and Wang, Yizhong and Hajishirzi, Hannaneh and Smith, Noah A and Choi, Yejin},
  booktitle={Proceedings of the 2020 Conference on Empirical Methods in Natural Language Processing (EMNLP)},
  pages={9275--9293},
  year={2020}
}

@inproceedings{rodriguez2021evaluation,
  title={Evaluation examples are not equally informative: How should that change NLP leaderboards?},
  author={Rodriguez, Pedro and Barrow, Joe and Hoyle, Alexander Miserlis and Lalor, John P and Jia, Robin and Boyd-Graber, Jordan},
  booktitle={Proceedings of the 59th Annual Meeting of the Association for Computational Linguistics and the 11th International Joint Conference on Natural Language Processing (Volume 1: Long Papers)},
  pages={4486--4503},
  year={2021}
}

@article{clark2020electra,
  title={Electra: Pre-training text encoders as discriminators rather than generators},
  author={Clark, Kevin and Luong, Minh-Thang and Le, Quoc V and Manning, Christopher D},
  journal={arXiv preprint arXiv:2003.10555},
  year={2020}
}

@article{lagarias1998convergence,
  title={Convergence properties of the Nelder--Mead simplex method in low dimensions},
  author={Lagarias, Jeffrey C and Reeds, James A and Wright, Margaret H and Wright, Paul E},
  journal={SIAM Journal on optimization},
  volume={9},
  number={1},
  pages={112--147},
  year={1998},
  publisher={SIAM}
}

@inproceedings{tsvetkov2016learning,
  title={Learning the Curriculum with Bayesian Optimization for Task-Specific Word Representation Learning},
  author={Tsvetkov, Yulia and Faruqui, Manaal and Ling, Wang and MacWhinney, Brian and Dyer, Chris},
  booktitle={Proceedings of the 54th Annual Meeting of the Association for Computational Linguistics (Volume 1: Long Papers)},
  pages={130--139},
  year={2016}
}

@inproceedings{zhan2021meta,
  title={Meta-curriculum learning for domain adaptation in neural machine translation},
  author={Zhan, Runzhe and Liu, Xuebo and Wong, Derek F and Chao, Lidia S},
  booktitle={Proceedings of the AAAI Conference on Artificial Intelligence},
  volume={35},  
  pages={14310--14318},
  year={2021}
}

@article{cirik2016visualizing,
  title={Visualizing and understanding curriculum learning for long short-term memory networks},
  author={Cirik, Volkan and Hovy, Eduard and Morency, Louis-Philippe},
  journal={arXiv preprint arXiv:1611.06204},
  year={2016}
}

@article{lalor2023py,
  title={py-irt: A scalable item response theory library for python},
  author={Lalor, John Patrick and Rodriguez, Pedro},
  journal={INFORMS Journal on Computing},
  volume={35},
  number={1},
  pages={5--13},
  year={2023},
  publisher={INFORMS}
}

@article{zhang2024weighted,
  title={Weighted Self-Paced Learning with Belief Functions},
  author={Zhang, Shixing and Han, Deqiang and Dezert, Jean and Yang, Yi},
  journal={Expert Systems with Applications},
  pages={124535},
  year={2024},
  publisher={Elsevier}
}

@article{efficient_transformer,
author = {Tay, Yi and Dehghani, Mostafa and Bahri, Dara and Metzler, Donald},
title = {Efficient Transformers: A Survey},
year = {2022},
issue_date = {June 2023},
publisher = {Association for Computing Machinery},
address = {New York, NY, USA},
volume = {55},
number = {6},
issn = {0360-0300},
url = {https://doi.org/10.1145/3530811},
doi = {10.1145/3530811},
journal = {ACM Comput. Surv.},
month = {dec},
articleno = {109},
numpages = {28}
}

@article{khan2022transformers,
  title={Transformers in vision: A survey},
  author={Khan, Salman and Naseer, Muzammal and Hayat, Munawar and Zamir, Syed Waqas and Khan, Fahad Shahbaz and Shah, Mubarak},
  journal={ACM computing surveys (CSUR)},
  volume={54},
  number={10s},
  pages={1--41},
  year={2022},
  publisher={ACM New York, NY}
}

@inproceedings{wang2019glue,
  title={{GLUE}: A Multi-Task Benchmark and Analysis Platform for Natural Language Understanding},
  author={Wang, Alex and Singh, Amanpreet and Michael, Julian and Hill, Felix and Levy, Omer and Bowman, Samuel R.},
  booktitle={Proceedings of the International Conference on Learning Representations (ICLR)},
  year={2019}
}

@article{vaswani2017attention,
  title={Attention is all you need},
  author={Vaswani, Ashish and Shazeer, Noam and Parmar, Niki and Uszkoreit, Jakob and Jones, Llion and Gomez, Aidan N and Kaiser, {\L}ukasz and Polosukhin, Illia},
  journal={Advances in neural information processing systems},
  volume={30},
  year={2017}
}

@article{cobbe2021training,
  title={Training verifiers to solve math word problems},
  author={Cobbe, Karl and Kosaraju, Vineet and Bavarian, Mohammad and Chen, Mark and Jun, Heewoo and Kaiser, Lukasz and Plappert, Matthias and Tworek, Jerry and Hilton, Jacob and Nakano, Reiichiro and others},
  journal={arXiv preprint arXiv:2110.14168},
  year={2021}
}

@inproceedings{kenton2019bert,
  title={BERT: Pre-training of Deep Bidirectional Transformers for Language Understanding},
  author={Kenton, Jacob Devlin Ming-Wei Chang and Toutanova, Lee Kristina},
  booktitle={Proceedings of NAACL-HLT},
  pages={4171--4186},
  year={2019}
}

@article{bai2022exploiting,
  title={Exploiting diverse information in pre-trained language model for multi-choice machine reading comprehension},
  author={Bai, Ziwei and Liu, Junpeng and Wang, Meiqi and Yuan, Caixia and Wang, Xiaojie},
  journal={Applied Sciences},
  volume={12},
  number={6},
  pages={3072},
  year={2022},
  publisher={MDPI}
}

@article{brown2020language,
  title={Language models are few-shot learners},
  author={Brown, Tom and Mann, Benjamin and Ryder, Nick and Subbiah, Melanie and Kaplan, Jared D and Dhariwal, Prafulla and Neelakantan, Arvind and Shyam, Pranav and Sastry, Girish and Askell, Amanda and others},
  journal={Advances in neural information processing systems},
  volume={33},
  pages={1877--1901},
  year={2020}
}

@techreport{radford2019language,
  title={Language Models are Unsupervised Multitask Learners},
  author={Radford, Alec and Wu, Jeff and Child, Rewon and Luan, David and Amodei, Dario and Sutskever, Ilya},
  year={2019},
  institution={OpenAI},
  type={Technical Report}
}

@inproceedings{senguptagood,
  title={A Good Learner can Teach Better: Teacher-Student Collaborative Knowledge Distillation},
  author={Sengupta, Ayan and Dixit, Shantanu and Akhtar, Md Shad and Chakraborty, Tanmoy},
  booktitle={The Twelfth International Conference on Learning Representations},
  year={2023}
}

@inproceedings{ouyang2023unsupervised,
  title={Unsupervised Aspect Term Extraction by Integrating Sentence-level Curriculum Learning with Token-level Self-paced Learning},
  author={Ouyang, Jihong and Yang, Zhiyao and Xuan, Chang and Wang, Bing and Wang, Yiyuan and Li, Ximing},
  booktitle={Proceedings of the 32nd ACM International Conference on Information and Knowledge Management},
  pages={1982--1991},
  year={2023}
}

@inproceedings{mohiuddin-etal-2022-data,
    title = "Data Selection Curriculum for Neural Machine Translation",
    author = "Mohiuddin, Tasnim  and
      Koehn, Philipp  and
      Chaudhary, Vishrav  and
      Cross, James  and
      Bhosale, Shruti  and
      Joty, Shafiq",
    editor = "Goldberg, Yoav  and
      Kozareva, Zornitsa  and
      Zhang, Yue",
    booktitle = "Findings of the Association for Computational Linguistics: EMNLP 2022",
    month = dec,
    year = "2022",
    address = "Abu Dhabi, United Arab Emirates",
    publisher = "Association for Computational Linguistics",
    url = "https://aclanthology.org/2022.findings-emnlp.113",
    doi = "10.18653/v1/2022.findings-emnlp.113",
    pages = "1569--1582"
}

@article{yao2007early,
  title={On early stopping in gradient descent learning},
  author={Yao, Yuan and Rosasco, Lorenzo and Caponnetto, Andrea},
  journal={Constructive Approximation},
  volume={26},
  number={2},
  pages={289--315},
  year={2007},
  publisher={Springer}
}

@inproceedings{xu2020curriculum,
  title={Curriculum learning for natural language understanding},
  author={Xu, Benfeng and Zhang, Licheng and Mao, Zhendong and Wang, Quan and Xie, Hongtao and Zhang, Yongdong},
  booktitle={Proceedings of the 58th Annual Meeting of the Association for Computational Linguistics},
  pages={6095--6104},
  year={2020}
}

@inproceedings{weinshall2018curriculum,
  title={Curriculum learning by transfer learning: Theory and experiments with deep networks},
  author={Weinshall, Daphna and Cohen, Gad and Amir, Dan},
  booktitle={International conference on machine learning},
  pages={5238--5246},
  year={2018},
  organization={PMLR}
}

@article{manela2022curriculum,
  title={Curriculum learning with hindsight experience replay for sequential object manipulation tasks},
  author={Manela, Binyamin and Biess, Armin},
  journal={Neural Networks},
  volume={145},
  pages={260--270},
  year={2022},
  publisher={Elsevier}
}

@inproceedings{burduja2021unsupervised,
  title={Unsupervised medical image alignment with curriculum learning},
  author={Burduja, Mihail and Ionescu, Radu Tudor},
  booktitle={2021 IEEE International Conference on Image Processing (ICIP)},
  pages={3787--3791},
  year={2021},
  organization={IEEE}
}

@inproceedings{zhao2021automatic,
  title={Automatic curriculum learning with over-repetition penalty for dialogue policy learning},
  author={Zhao, Yangyang and Wang, Zhenyu and Huang, Zhenhua},
  booktitle={Proceedings of the AAAI Conference on Artificial Intelligence},
  pages={14540--14548},
  year={2021}
}

@article{liu2022competence,
  title={Competence-based multimodal curriculum learning for medical report generation},
  author={Liu, Fenglin and Ge, Shen and Zou, Yuexian and Wu, Xian},
  journal={arXiv preprint arXiv:2206.14579},
  year={2022}
}

@article{soviany2021curriculum,
  title={Curriculum self-paced learning for cross-domain object detection},
  author={Soviany, Petru and Ionescu, Radu Tudor and Rota, Paolo and Sebe, Nicu},
  journal={Computer Vision and Image Understanding},
  volume={204},
  pages={103166},
  year={2021},
  publisher={Elsevier}
}

@article{zhang2021flexmatch,
  title={Flexmatch: Boosting semi-supervised learning with curriculum pseudo labeling},
  author={Zhang, Bowen and Wang, Yidong and Hou, Wenxin and Wu, Hao and Wang, Jindong and Okumura, Manabu and Shinozaki, Takahiro},
  journal={Advances in Neural Information Processing Systems},
  volume={34},
  pages={18408--18419},
  year={2021}
}

@article{milano2021automated,
  title={Automated curriculum learning for embodied agents a neuroevolutionary approach},
  author={Milano, Nicola and Nolfi, Stefano},
  journal={Scientific reports},
  volume={11},
  number={1},
  pages={8985},
  year={2021},
  publisher={Nature Publishing Group UK London}
}

@inproceedings{maharana2022curriculum,
  title={On curriculum learning for commonsense reasoning},
  author={Maharana, Adyasha and Bansal, Mohit},
  booktitle={Proceedings of the 2022 Conference of the North American Chapter of the Association for Computational Linguistics: Human Language Technologies},
  pages={983--992},
  year={2022}
}

@inproceedings{wan-etal-2020-self,
    title = "Self-Paced Learning for Neural Machine Translation",
    author = "Wan, Yu  and
      Yang, Baosong  and
      Wong, Derek F.  and
      Zhou, Yikai  and
      Chao, Lidia S.  and
      Zhang, Haibo  and
      Chen, Boxing",
    editor = "Webber, Bonnie  and
      Cohn, Trevor  and
      He, Yulan  and
      Liu, Yang",
    booktitle = "Proceedings of the 2020 Conference on Empirical Methods in Natural Language Processing (EMNLP)",
    month = nov,
    year = "2020",
    address = "Online",
    publisher = "Association for Computational Linguistics",
    url = "https://aclanthology.org/2020.emnlp-main.80",
    doi = "10.18653/v1/2020.emnlp-main.80",
    pages = "1074--1080"
}

@inproceedings{he2022debertav3,
  title={DeBERTaV3: Improving DeBERTa using ELECTRA-Style Pre-Training with Gradient-Disentangled Embedding Sharing},
  author={He, Pengcheng and Gao, Jianfeng and Chen, Weizhu},
  booktitle={The Eleventh International Conference on Learning Representations},
  year={2022}
}

@article{2020t5,
  author  = {Colin Raffel and Noam Shazeer and Adam Roberts and Katherine Lee and Sharan Narang and Michael Matena and Yanqi Zhou and Wei Li and Peter J. Liu},
  title   = {Exploring the Limits of Transfer Learning with a Unified Text-to-Text Transformer},
  journal = {Journal of Machine Learning Research},
  year    = {2020},
  volume  = {21},
  number  = {140},
  pages   = {1-67},
  url     = {http://jmlr.org/papers/v21/20-074.html}
}

@book{de2013theory,
	title={The theory and practice of item response theory},
	author={De Ayala, Rafael Jaime},
	year={2013},
	publisher={Guilford Publications}
}

@inproceedings{DBLP:conf/edm/WuDDPG20,
	author    = {Mike Wu and
	Richard L. Davis and
	Benjamin W. Domingue and
	Chris Piech and
	Noah D. Goodman},
	editor    = {Anna N. Rafferty and
	Jacob Whitehill and
	Crist{\'{o}}bal Romero and
	Violetta Cavalli{-}Sforza},
	title     = {Variational Item Response Theory: Fast, Accurate, and Expressive},
	booktitle = {Proceedings of the 13th International Conference on Educational Data
	Mining, {EDM} 2020, Fully virtual conference, July 10-13, 2020},
	publisher = {International Educational Data Mining Society},
	year      = {2020},
	url       = {https://educationaldatamining.org/files/conferences/EDM2020/papers/paper\_22.pdf},
	bibsource = {dblp computer science bibliography, https://dblp.org}
}

@article{elman1993learning,
	title={Learning and development in neural networks: The importance of starting small},
	author={Elman, Jeffrey L},
	journal={Cognition},
	volume={48},
	number={1},
	pages={71--99},
	year={1993},
	publisher={Elsevier}
}

@article {Wilson255182,
	author = {Wilson, Robert C. and Shenhav, Amitai and Straccia, Mark and Cohen, Jonathan D.},
	title = {The Eighty Five Percent Rule for Optimal Learning},
	elocation-id = {255182},
	year = {2019},
	doi = {10.1038/s41467-019-12552-4},
	journal = {Nature Communications}
}

@inproceedings{williams2018broad,
	author    = {Williams, Adina and Nangia, Nikita and Bowman, Samuel R.},
	title = {A Broad-Coverage Challenge Corpus for Sentence Understanding through Inference},
	booktitle = {Proceedings of NAACL-HLT},
	year = 2018
}

@inproceedings{rajpurkar2016squad,
	author = {Rajpurkar, Pranav and Zhang, Jian and Lopyrev, Konstantin and Liang, Percy},
	title = {{SQ}u{AD}: 100,000+ Questions for Machine Comprehension of Text},
	booktitle = {Proceedings of EMNLP},
	year = {2016},
	publisher = {Association for Computational Linguistics},
	pages = {2383--2392},
	location = {Austin, Texas},
}

@inproceedings{dolan2005automatically,
	title={Automatically constructing a corpus of sentential paraphrases},
	author={Dolan, William B and Brockett, Chris},
	booktitle={Proceedings of the International Workshop on Paraphrasing},
	year={2005}
}

@inproceedings{bentivogli2009fifth,
  title={The Fifth {PASCAL} Recognizing Textual Entailment Challenge},
  author={Bentivogli, Luisa and Dagan, Ido and Dang, Hoa Trang and Giampiccolo, Danilo and Magnini, Bernardo},
  booktitle={Proceedings of the Text Analysis Conference (TAC)},
  year={2009}
}

@inproceedings{jiang2015self,
  title={Self-paced curriculum learning},
  author={Jiang, Lu and Meng, Deyu and Zhao, Qian and Shan, Shiguang and Hauptmann, Alexander},
  booktitle={Proceedings of the AAAI Conference on Artificial Intelligence},
  volume={29},
  year={2015}
}

@article{kumar2010self,
  title={Self-paced learning for latent variable models},
  author={Kumar, M and Packer, Benjamin and Koller, Daphne},
  journal={Advances in neural information processing systems},
  volume={23},
  year={2010}
}

@article{wang2021survey,
  title={A survey on curriculum learning},
  author={Wang, Xin and Chen, Yudong and Zhu, Wenwu},
  journal={IEEE transactions on pattern analysis and machine intelligence},
  volume={44},
  number={9},
  pages={4555--4576},
  year={2021},
  publisher={IEEE}
}

@article{wei2016stc,
  title={Stc: A simple to complex framework for weakly-supervised semantic segmentation},
  author={Wei, Yunchao and Liang, Xiaodan and Chen, Yunpeng and Shen, Xiaohui and Cheng, Ming-Ming and Feng, Jiashi and Zhao, Yao and Yan, Shuicheng},
  journal={IEEE transactions on pattern analysis and machine intelligence},
  volume={39},
  number={11},
  pages={2314--2320},
  year={2016},
  publisher={IEEE}
}

@inproceedings{spitkovsky2010baby,
  title={From baby steps to leapfrog: How “less is more” in unsupervised dependency parsing},
  author={Spitkovsky, Valentin I and Alshawi, Hiyan and Jurafsky, Dan},
  booktitle={Human Language Technologies: The 2010 Annual Conference of the North American Chapter of the Association for Computational Linguistics},
  pages={751--759},
  year={2010}
}

@inproceedings{bowman2015large,
  title={A large annotated corpus for learning natural language inference},
  author={Bowman, Samuel and Angeli, Gabor and Potts, Christopher and Manning, Christopher D},
  booktitle={Proceedings of the 2015 Conference on Empirical Methods in Natural Language Processing},
  pages={632--642},
  year={2015}
}

@inproceedings{bengio_curriculum_2009,
	title = {Curriculum learning},
	url = {http://dl.acm.org/citation.cfm?id=1553380},
	urldate = {2017-02-23},
	booktitle = {Proceedings of the 26th annual international conference on machine learning},
	publisher = {ACM},
	author = {Bengio, Yoshua and Louradour, Jérôme and Collobert, Ronan and Weston, Jason},
	year = {2009},
	pages = {41--48}
}

@article{hochreiter_long_1997,
	title = {Long {Short}-{Term} {Memory}},
	volume = {9},
	issn = {0899-7667},
	number = {8},
	journal = {Neural Computation},
	author = {Hochreiter, S and Schmidhuber, J},
	month = nov,
	year = {1997},
	pages = {1735--1780}
}

@article{yuan2019adversarial,
  title={Adversarial examples: Attacks and defenses for deep learning},
  author={Yuan, Xiaoyong and He, Pan and Zhu, Qile and Li, Xiaolin},
  journal={IEEE transactions on neural networks and learning systems},
  volume={30},
  number={9},
  pages={2805--2824},
  year={2019},
  publisher={IEEE}
}

@article{bock_marginal_1981,
	title = {Marginal maximum likelihood estimation of item parameters: {Application} of an {EM} algorithm},
	volume = {46},
	number = {4},
	journal = {Psychometrika},
	author = {Bock, R Darrell and Aitkin, Murray},
	year = {1981},
	pages = {443--459}
}

@inproceedings{socher_recursive_2013,
	address = {Seattle, Washington, USA},
	title = {Recursive {Deep} {Models} for {Semantic} {Compositionality} {Over} a {Sentiment} {Treebank}},
	url = {http://aclweb.org/anthology/D13-1170},
	booktitle = {Proceedings of the 2013 {Conference} on {Empirical} {Methods} in {Natural} {Language} {Processing}},
	publisher = {Association for Computational Linguistics},
	author = {Socher, Richard and Perelygin, Alex and Wu, Jean and Chuang, Jason and Manning, D. Christopher and Ng, Andrew and Potts, Christopher},
	year = {2013},
	pages = {1631--1642}
}

@inproceedings{nagatsuka2021pre,
  title={Pre-training a BERT with curriculum learning by increasing block-size of input text},
  author={Nagatsuka, Koichi and Broni-Bediako, Clifford and Atsumi, Masayasu},
  booktitle={Proceedings of the International Conference on Recent Advances in Natural Language Processing (RANLP 2021)},
  pages={989--996},
  year={2021}
}

@article{lee2022efficient,
  title={Efficient pre-training of masked language model via concept-based curriculum masking},
  author={Lee, Mingyu and Park, Jun-Hyung and Kim, Junho and Kim, Kang-Min and Lee, SangKeun},
  journal={arXiv preprint arXiv:2212.07617},
  year={2022}
}

@inproceedings{hacohen2019power,
  title={On the power of curriculum learning in training deep networks},
  author={Hacohen, Guy and Weinshall, Daphna},
  booktitle={International conference on machine learning},
  pages={2535--2544},
  year={2019},
  organization={PMLR}
}

@article{he2020deberta,
  title={Deberta: Decoding-enhanced bert with disentangled attention},
  author={He, Pengcheng and Liu, Xiaodong and Gao, Jianfeng and Chen, Weizhu},
  journal={arXiv preprint arXiv:2006.03654},
  year={2020}
}

@article{hoffman2013stochastic,
  title={Stochastic variational inference},
  author={Hoffman, Matthew D and Blei, David M and Wang, Chong and Paisley, John},
  journal={Journal of Machine Learning Research},
  year={2013}
}

@article{sanh2019distilbert,
  title={DistilBERT, a distilled version of BERT: smaller, faster, cheaper and lighter},
  author={Sanh, Victor and Debut, Lysandre and Chaumond, Julien and Wolf, Thomas},
  journal={arXiv preprint arXiv:1910.01108},
  year={2019}
}

@article{lewis2019bart,
  title={Bart: Denoising sequence-to-sequence pre-training for natural language generation, translation, and comprehension},
  author={Lewis, Mike and Liu, Yinhan and Goyal, Naman and Ghazvininejad, Marjan and Mohamed, Abdelrahman and Levy, Omer and Stoyanov, Ves and Zettlemoyer, Luke},
  journal={arXiv preprint arXiv:1910.13461},
  year={2019}
}

@article{lan2019albert,
  title={Albert: A lite bert for self-supervised learning of language representations},
  author={Lan, Zhenzhong and Chen, Mingda and Goodman, Sebastian and Gimpel, Kevin and Sharma, Piyush and Soricut, Radu},
  journal={arXiv preprint arXiv:1909.11942},
  year={2019}
}

@article{yang2019xlnet,
  title={Xlnet: Generalized autoregressive pretraining for language understanding},
  author={Yang, Zhilin and Dai, Zihang and Yang, Yiming and Carbonell, Jaime and Salakhutdinov, Russ R and Le, Quoc V},
  journal={Advances in neural information processing systems},
  volume={32},
  year={2019}
}

@article{liu2019roberta,
  title={Roberta: A robustly optimized bert pretraining approach},
  author={Liu, Yinhan and Ott, Myle and Goyal, Naman and Du, Jingfei and Joshi, Mandar and Chen, Danqi and Levy, Omer and Lewis, Mike and Zettlemoyer, Luke and Stoyanov, Veselin},
  journal={arXiv preprint arXiv:1907.11692},
  year={2019}
}

@article{jordan1999introduction,
  title={An introduction to variational methods for graphical models},
  author={Jordan, Michael I and Ghahramani, Zoubin and Jaakkola, Tommi S and Saul, Lawrence K},
  journal={Machine learning},
  volume={37},
  pages={183--233},
  year={1999},
  publisher={Springer}
}

@book{baker_item_2004,
	title = {Item {Response} {Theory}: {Parameter} {Estimation} {Techniques}, {Second} {Edition}},
	isbn = {978-0-8247-5825-7},
	shorttitle = {Item {Response} {Theory}},
	language = {en},
	publisher = {CRC Press},
	author = {Baker, Frank B. and Kim, Seock-Ho},
	month = jul,
	year = {2004}
}

@inproceedings{zhao2020reinforced,
  title={Reinforced curriculum learning on pre-trained neural machine translation models},
  author={Zhao, Mingjun and Wu, Haijiang and Niu, Di and Wang, Xiaoli},
  booktitle={Proceedings of the AAAI Conference on Artificial Intelligence},
  volume={34},  
  pages={9652--9659},
  year={2020}
}

@inproceedings{kocmi2017curriculum,
  title={Curriculum Learning and Minibatch Bucketing in Neural Machine Translation},
  author={Kocmi, Tom and Bojar, Ond{\v{r}}ej},
  booktitle={Proceedings of the International Conference Recent Advances in Natural Language Processing, RANLP 2017},
  pages={379--386},
  year={2017}
}

@article{kumar2019reinforcement,
  title={Reinforcement learning based curriculum optimization for neural machine translation},
  author={Kumar, Gaurav and Foster, George and Cherry, Colin and Krikun, Maxim},
  journal={arXiv preprint arXiv:1903.00041},
  year={2019}
}

@inproceedings{graves2017automated,
  title={Automated curriculum learning for neural networks},
  author={Graves, Alex and Bellemare, Marc G and Menick, Jacob and Munos, Remi and Kavukcuoglu, Koray},
  booktitle={international conference on machine learning},
  pages={1311--1320},
  year={2017},
  organization={Pmlr}
}

@article{bingham2018pyro,
	author = {Bingham, Eli and Chen, Jonathan P. and Jankowiak, Martin and Obermeyer, Fritz and
	Pradhan, Neeraj and Karaletsos, Theofanis and Singh, Rohit and Szerlip, Paul and
	Horsfall, Paul and Goodman, Noah D.},
	title = {{Pyro: Deep Universal Probabilistic Programming}},
	journal = {Journal of Machine Learning Research},
	year = {2018}
}

@article{freund1997decision,
  title={A decision-theoretic generalization of on-line learning and an application to boosting},
  author={Freund, Yoav and Schapire, Robert E},
  journal={Journal of computer and system sciences},
  volume={55},
  number={1},
  pages={119--139},
  year={1997},
  publisher={Elsevier}
}

@inproceedings{platanios_competence-based_2019,
	address = {Minneapolis, Minnesota},
	title = {Competence-based {Curriculum} {Learning} for {Neural} {Machine} {Translation}},
	url = {https://www.aclweb.org/anthology/N19-1119},
	urldate = {2019-07-31},
	booktitle = {Proceedings of the 2019 {Conference} of the {North} {American} {Chapter} of the {Association} for {Computational} {Linguistics}: {Human} {Language} {Technologies}, {Volume} 1 ({Long} and {Short} {Papers})},
	publisher = {Association for Computational Linguistics},
	author = {Platanios, Emmanouil Antonios and Stretcu, Otilia and Neubig, Graham and Poczos, Barnabas and Mitchell, Tom},
	month = jun,
	year = {2019},
	pages = {1162--1172}
}

@article{natesan_bayesian_2016,
	title = {Bayesian {Prior} {Choice} in {IRT} {Estimation} {Using} {MCMC} and {Variational} {Bayes}},
	volume = {7},
	issn = {1664-1078},
	url = {https://www.ncbi.nlm.nih.gov/pmc/articles/PMC5037236/},
	doi = {10.3389/fpsyg.2016.01422},
	urldate = {2019-07-31},
	journal = {Frontiers in Psychology},
	author = {Natesan, Prathiba and Nandakumar, Ratna and Minka, Tom and Rubright, Jonathan D.},
	month = sep,
	year = {2016},
	pmid = {27729878},
	pmcid = {PMC5037236}
}

@inproceedings{lalor_learning_2019,
	title = {Learning {Latent} {Parameters} without {Human} {Response} {Patterns}: {Item} {Response} {Theory} with {Artificial} {Crowds}},
	volume = {2019},
	booktitle = {Proceedings of the {Conference} on {Empirical} {Methods} in {Natural} {Language} {Processing}. {Conference} on {Empirical} {Methods} in {Natural} {Language} {Processing}},
	publisher = {Association for Computational Linguistics},
	author = {Lalor, John P. and Wu, Hao and Yu, Hong},
	year = {2019}
}

@book{rasch_studies_1960,
	address = {Oxford, England},
	series = {Studies in mathematical psychology: {I}. {Probabilistic} models for some intelligence and attainment tests},
	title = {Studies in mathematical psychology: {I}. {Probabilistic} models for some intelligence and attainment tests},
	shorttitle = {Studies in mathematical psychology},
	publisher = {Nielsen \& Lydiche},
	author = {Rasch, Georg},
	year = {1960}
}

@inproceedings{shrivastava_training_2016,
	title = {Training region-based object detectors with online hard example mining},
	booktitle = {Proceedings of the {IEEE} conference on computer vision and pattern recognition},
	author = {Shrivastava, Abhinav and Gupta, Abhinav and Girshick, Ross},
	year = {2016},
	pages = {761--769}
}

@article{jin2020disease,
  title={What Disease does this Patient Have? A Large-scale Open Domain Question Answering Dataset from Medical Exams},
  author={Jin, Di and Pan, Eileen and Oufattole, Nassim and Weng, Wei-Hung and Fang, Hanyi and Szolovits, Peter},
  journal={arXiv preprint arXiv:2009.13081},
  year={2020}
}

@article{zhang2015character,
  title={Character-level convolutional networks for text classification},
  author={Zhang, Xiang and Zhao, Junbo and LeCun, Yann},
  journal={Advances in neural information processing systems},
  volume={28},
  year={2015}
}

@article{yang2024qwen2,
  title={Qwen2. 5 Technical Report},
  author={Yang, An and Yang, Baosong and Zhang, Beichen and Hui, Binyuan and Zheng, Bo and Yu, Bowen and Li, Chengyuan and Liu, Dayiheng and Huang, Fei and Wei, Haoran and others},
  journal={arXiv e-prints},
  pages={arXiv--2412},
  year={2024}
}

@article{grattafiori2024llama,
  title={The llama 3 herd of models},
  author={Grattafiori, Aaron and Dubey, Abhimanyu and Jauhri, Abhinav and Pandey, Abhinav and Kadian, Abhishek and Al-Dahle, Ahmad and Letman, Aiesha and Mathur, Akhil and Schelten, Alan and Vaughan, Alex and others},
  journal={arXiv preprint arXiv:2407.21783},
  year={2024}
}

@article{hu2022lora,
  title={Lora: Low-rank adaptation of large language models.},
  author={Hu, Edward J and Shen, Yelong and Wallis, Phillip and Allen-Zhu, Zeyuan and Li, Yuanzhi and Wang, Shean and Wang, Lu and Chen, Weizhu and others},
  journal={ICLR},
  volume={1},
  number={2},
  pages={3},
  year={2022}
}

@article{loshchilov2017decoupled,
  title={Decoupled weight decay regularization},
  author={Loshchilov, Ilya and Hutter, Frank},
  journal={arXiv preprint arXiv:1711.05101},
  year={2017}
}

@article{dettmers2023qlora,
  title={Qlora: Efficient finetuning of quantized llms},
  author={Dettmers, Tim and Pagnoni, Artidoro and Holtzman, Ari and Zettlemoyer, Luke},
  journal={Advances in neural information processing systems},
  volume={36},
  pages={10088--10115},
  year={2023}
}

@misc{anthropic2024claude,
  author = {Anthropic},
  title = {Introducing Claude 3.5 Sonnet},
  year = {2024},
  url = {https://www.anthropic.com/news/claude-3-5-sonnet},
  note = {Accessed: 2024-06-20}
}

@article{deepseekai2024deepseekv3,
  title = {{DeepSeek-V3} Technical Report},
  author = {{DeepSeek-AI}},
  journal = {arXiv preprint arXiv:2412.19437},
  year = {2024},
  url = {https://arxiv.org/abs/2412.19437}
}

@misc{openai2024gpt4o,
  author = {OpenAI},
  title = {{GPT-4o} System Card},
  year = {2024},
  url = {https://openai.com/index/gpt-4o-system-card/},
  note = {Accessed: 2024-08-08}
}

@article{young2024yi,
  title = {Yi: Open Foundation Models by 01.{AI}},
  author = {Young, Alex and Chen, Bei and Li, Chao and Huang, Chengen and Zhang, Ge and Zhang, Guanwei and Li, Heng and Zhu, Jiangcheng and Chen, Jianqun and Chang, Jing and others},
  journal = {arXiv preprint arXiv:2403.04652},
  year = {2024},
  url = {https://arxiv.org/abs/2403.04652}
}

@article{gemmateam2024gemma2,
  title = {Gemma 2: Improving Open Language Models at a Practical Size},
  author = {{Gemma Team} and Rivi\`ere, Morgane and Pathak, Shreya and Sessa, Pier Giuseppe and Hardin, Cassidy and Bhupatiraju, Surya and Hussenot, L\'eonard and Mesnard, Thomas and Shahriari, Bobak and Ram\'e, Alexandre and others},
  journal = {arXiv preprint arXiv:2408.00118},
  year = {2024},
  url = {https://arxiv.org/abs/2408.00118}
}

@article{jiang2023mistral,
  title = {Mistral 7{B}},
  author = {Jiang, Albert Q. and Sablayrolles, Alexandre and Mensch, Arthur and Bamford, Chris and Chaplot, Devendra Singh and Casas, Diego de las and Bressand, Florian and Lengyel, Gianna and Lample, Guillaume and Saulnier, Lucile and others},
  journal = {arXiv preprint arXiv:2310.06825},
  year = {2023},
  url = {https://arxiv.org/abs/2310.06825}
}

@inproceedings{wei2022chain,
  title = {Chain-of-Thought Prompting Elicits Reasoning in Large Language Models},
  author = {Wei, Jason and Wang, Xuezhi and Schuurmans, Dale and Bosma, Maarten and Ichter, Brian and Xia, Fei and Chi, Ed and Le, Quoc V. and Zhou, Denny},
  booktitle = {Advances in Neural Information Processing Systems},
  volume = {35},
  pages = {24824--24837},
  year = {2022}
}

@misc{openai2025gpt5,
  author = {OpenAI},
  title = {{GPT-5} System Card},
  year = {2025},
  url = {https://cdn.openai.com/gpt-5-system-card.pdf},
  note = {Accessed: 2025-11-02}
}

@misc{ibm2025granite,
  author = {{IBM}},
  title = {ibm-granite/granite-3.3-8b-instruct},
  year = {2025},
  publisher = {Hugging Face},
  url = {https://huggingface.co/ibm-granite/granite-3.3-8b-instruct},
  note = {Accessed: 2025-11-02}
}

\newpage
\appendix
\section{Appendix}
\subsection{Original Formulation of Artificial Crowd}
\label{ssec:acOld}

\tikzstyle{decision} = [diamond, draw, fill=blue!20, 
text width=4.5em, text badly centered, node distance=3cm, inner sep=0pt]
\tikzstyle{block} = [rectangle, draw, fill=blue!20, 
text width=5em, text centered, rounded corners, minimum height=4em]
\tikzstyle{line} = [draw, -latex']
\tikzstyle{cloud} = [draw, ellipse,fill=red!20, 
minimum height=2em, text width=5em, text centered]
\begin{figure}[h]
	\centering 
	\hspace{7em}
	\begin{tikzpicture}[node distance = 0.35cm and 0.75cm, auto]
		\node [cloud] (bank) {Training set};
		\node [block, right= of bank] (rps) {Sample,\\Add noise} ;
		\node [block, right= of rps]  (irt) {Train DNN};
		\node [cloud, above= of irt] (dataset) {Full dataset};
		\node [cloud, right= of irt] (test) {Output \\response \\pattern};
		
		\path [line] (bank) -- (rps);
		\path [line] (rps) -- (irt);
		\path [line] (dataset) -- (test);
		\path [line] (irt) -- (test);
	\end{tikzpicture}
	\caption{Response pattern construction for IRT model fitting with artificial crowds from our prior work \citep{lalor-yu-2020-dynamic}.}
	\label{fig:artificialcrowd}
\end{figure}

\subsection{Main Results for GLUE Tasks}

\begin{figure}[htb]
    \centering
    \includegraphics [width=\textwidth] {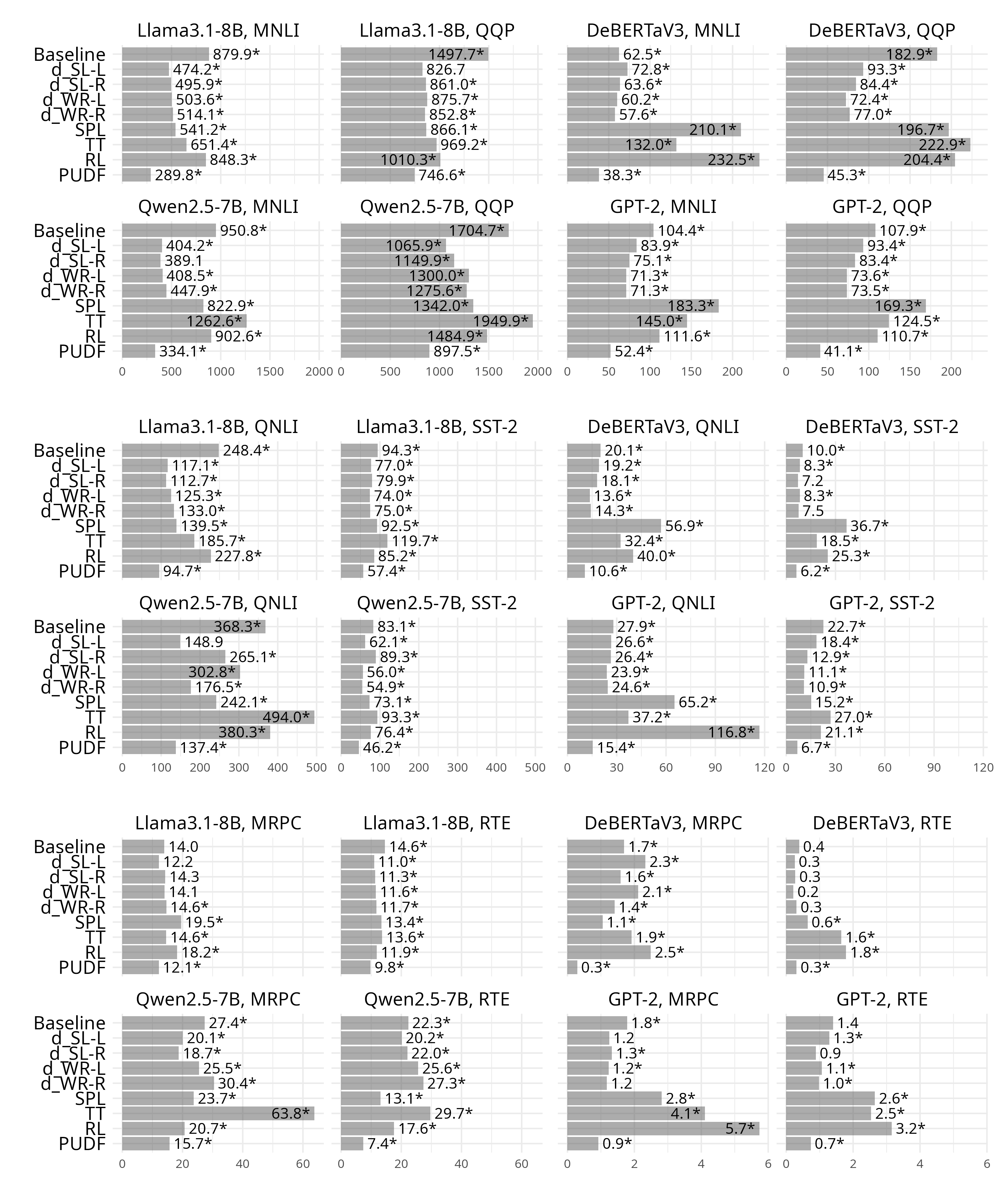}
    \caption{Comparing training time between \modelabbr{} and other CL methods for the GLUE datasets. All runtimes are measured in minutes.\\
    $*$Indicates that the runtime is significantly longer than \modelabbr{} (Welch's single-tailed t-test with Benjamini-Hochberg correction, $\alpha < 0.05$).}
    \label{fig:traintimeCLGLUE}
\end{figure}

In this section, we report the detailed results across each GLUE task, which have been aggregated in our main results. 
Table \ref{tab:appendixResult} and Figure \ref{fig:traintimeCLGLUE} show the classification performance and runtime comparison, respectively, for our benchmarking. 
\modelabbr{} consistently enhances both accuracy and training efficiency across multiple LLMs on the benchmark datasets. 
These results highlight \modelabbr{}'s effectiveness in enhancing LLM performance and efficiency across NLP tasks, with particularly strong benefits in reducing computational demands while maintaining or improving accuracy.
In almost all cases, the standard deviation of \modelabbr{} is low enough to suggest that our improved performance is consistently higher than the benchmarks. 
The RTE dataset reveals high standard deviations, likely due to the relatively smaller data size compared to other datasets.
We see similarly consistent results regarding training time. 
This indicates that \modelabbr{} can effectively reduce training time across a wide range of scenarios. 
Overall, based on these results, we can summarize that \modelabbr{} can improve model accuracy and reduce training time across most scenarios.


\begin{landscape}

	\begin{table}

		\caption{\label{tab:appendixResult}
			Mean and standard deviation accuracy results comparing PUDF with other CL Methods for the GLUE datasets over 5 runs. Best performing method for each model is in \textbf{bold}; the second best model is \underline{underlined}.\\
		}
		\centering
		\footnotesize
		\begin{tabular}[t]{llllllll}
			\toprule
			\textbf{Model} & \textbf{Method} & \textbf{MNLI}                   & \textbf{MRPC}                   & \textbf{QNLI}                   & \textbf{QQP}                & \textbf{RTE}                & \textbf{SST2}               \\
			\midrule
			DeBERTaV3      & Baseline        & $\underline{90.26}_{\pm 0.04}$  & $86.39_{\pm 1.28}$              & $93.43_{\pm 0.28}$              & $92.00_{\pm 0.11}$          & $80.24_{\pm 0.91}$          & $95.58_{\pm 0.18}$          \\
			               & d\_SL-L         & $89.45_{\pm 0.05}$              & $85.86_{\pm 1.37}$              & $92.84_{\pm 0.12}$              & $92.08_{\pm 0.18}$          & $74.52_{\pm 0.46}$          & $93.99_{\pm 0.28}$          \\
			               & d\_SL-R         & $89.78_{\pm 0.09}$              & $85.86_{\pm 1.83}$              & $92.72_{\pm 0.21}$              & $\underline{92.17}_{\pm 0.19}$& $75.96_{\pm 1.85}$        & $94.98_{\pm 0.38}$          \\
			               & d\_WR-L         & $89.42_{\pm 0.04}$              & $85.57_{\pm 2.00}$              & $92.92_{\pm 0.16}$              & $92.01_{\pm 0.15}$          & $69.05_{\pm 1.99}$          & $94.13_{\pm 0.51}$          \\
			               & d\_WR-R         & $89.64_{\pm 0.10}$              & $86.40_{\pm 1.66}$              & $93.17_{\pm 0.21}$              & $92.08_{\pm 0.06}$          & $78.11_{\pm 0.53}$          & $94.42_{\pm 0.15}$          \\
			               & SPL             & $89.83_{\pm 0.14}$              & $86.86_{\pm 3.12}$              & $93.57_{\pm 0.10}$              & $91.44_{\pm 1.11}$          & $77.16_{\pm 1.09}$          & $95.87_{\pm 0.21}$          \\
			               & TT              & $89.09_{\pm 1.42}$              & $87.46_{\pm 2.06}$              & $93.35_{\pm 2.34}$              & $90.80_{\pm 1.94}$          & $\underline{81.13}_{\pm 0.40}$& $94.81_{\pm 0.93}$          \\
			               & RL              & $89.63_{\pm 0.7}$               & $\mathbf{89.48}_{\pm 1.15}$     & $\underline{94.04}_{\pm 0.46}$  & $90.35_{\pm 1.14}$          & $80.46_{\pm 0.93}$          & $\underline{95.96}_{\pm 0.29}$ \\
			               & PUDF            & $\mathbf{90.71}_{\pm 0.01}^{*}$ & $\underline{89.16}_{\pm 0.52}$  & $\mathbf{94.80}_{\pm 0.20}$     & $\mathbf{92.57}_{\pm 0.05}^{*}$ & $\mathbf{81.28}_{\pm 0.48}$ & $\mathbf{96.05}_{\pm 0.24}$ \\
			\cmidrule{1-8}
			GPT-2          & Baseline        & $81.66_{\pm 0.05}$              & $78.14_{\pm 0.20}$              & $87.53_{\pm 0.45}$              & $\underline{89.77}_{\pm 0.16}$ & $65.70_{\pm 0.70}$       & $91.25_{\pm 0.54}$          \\
			               & d\_SL-L         & $81.05_{\pm 0.15}$              & $75.48_{\pm 1.04}$              & $86.94_{\pm 0.37}$              & $88.98_{\pm 0.10}$          & $51.04_{\pm 3.25}$          & $\underline{91.29}_{\pm 0.10}$ \\
			               & d\_SL-R         & $81.41_{\pm 0.11}$              & $72.76_{\pm 1.00}$              & $\underline{87.62}_{\pm 0.40}$  & $89.02_{\pm 0.32}$          & $60.69_{\pm 1.38}$          & $90.49_{\pm 0.61}$          \\
			               & d\_WR-L         & $81.04_{\pm 0.18}$              & $77.64_{\pm 0.58}$              & $87.26_{\pm 0.36}$              & $88.91_{\pm 0.46}$          & $51.45_{\pm 2.49}$          & $90.35_{\pm 0.45}$          \\
			               & d\_WR-R         & $80.82_{\pm 0.14}$              & $74.53_{\pm 0.46}$              & $87.44_{\pm 0.49}$              & $88.90_{\pm 0.37}$          & $61.04_{\pm 2.16}$          & $90.51_{\pm 0.34}$          \\
			               & SPL             & $81.36_{\pm 0.16}$              & $76.86_{\pm 0.91}$              & $86.24_{\pm 0.89}$              & $\mathbf{89.82}_{\pm 0.06}^{*}$ & $56.66_{\pm 0.72}$      & $90.50_{\pm 0.22}$          \\
			               & TT              & $\underline{81.83}_{\pm 0.06}$  & $\mathbf{81.67}_{\pm 0.42}^{*}$ & $85.89_{\pm 1.08}$              & $85.47_{\pm 1.25}$          & $65.07_{\pm 0.54}$          & $90.88_{\pm 0.80}$          \\
			               & RL              & $81.27_{\pm 0.12}$              & $\underline{80.57}_{\pm 0.18}$  & $85.94_{\pm 0.94}$              & $88.06_{\pm 0.80}$          & $\mathbf{66.59}_{\pm 0.41}$ & $90.84_{\pm 1.29}$          \\
			               & PUDF            & $\mathbf{82.00}_{\pm 0.05}^{*}$ & $79.77_{\pm 0.17}$              & $\mathbf{88.15}_{\pm 0.48}^{*}$ & $89.73_{\pm 0.47}$          & $\underline{66.57}_{\pm 0.90}$& $\mathbf{91.97}_{\pm 0.09}^{*}$ \\
			\cmidrule{1-8}
			Llama3.1-8B    & Baseline        & $\underline{91.29}_{\pm 0.22}$  & $88.25_{\pm 0.11}$              & $93.86_{\pm 0.57}$              & $\underline{92.58}_{\pm 0.22}$ & $81.11_{\pm 0.97}$       & $96.46_{\pm 0.77}$          \\
			               & d\_SL-L         & $87.61_{\pm 1.14}$              & $86.09_{\pm 1.35}$              & $90.07_{\pm 2.44}$              & $90.07_{\pm 1.76}$          & $79.60_{\pm 0.90}$          & $93.24_{\pm 2.16}$          \\
			               & d\_SL-R         & $88.99_{\pm 0.78}$              & $86.82_{\pm 0.79}$              & $93.67_{\pm 0.43}$              & $90.80_{\pm 0.58}$          & $77.27_{\pm 2.31}$          & $94.67_{\pm 0.96}$          \\
			               & d\_WR-L         & $90.03_{\pm 0.34}$              & $85.84_{\pm 1.73}$              & $92.97_{\pm 0.59}$              & $91.95_{\pm 0.28}$          & $79.84_{\pm 0.67}$          & $95.03_{\pm 0.65}$          \\
			               & d\_WR-R         & $89.49_{\pm 0.55}$              & $85.53_{\pm 1.75}$              & $92.29_{\pm 0.68}$              & $91.39_{\pm 0.72}$          & $80.28_{\pm 0.77}$          & $95.58_{\pm 0.61}$          \\
			               & SPL             & $90.00_{\pm 0.14}$              & $\underline{88.45}_{\pm 0.33}$  & $\underline{94.08}_{\pm 0.36}$  & $92.00_{\pm 0.21}$          & $80.95_{\pm 0.89}$          & $96.69_{\pm 0.49}$          \\
			               & TT              & $90.33_{\pm 0.50}$              & $87.95_{\pm 0.41}$              & $92.85_{\pm 0.87}$              & $91.98_{\pm 0.32}$          & $80.81_{\pm 0.74}$          & $96.55_{\pm 0.49}$          \\
			               & RL              & $90.71_{\pm 0.28}$              & $87.24_{\pm 0.37}$              & $93.93_{\pm 0.80}$              & $92.30_{\pm 0.48}$          & $\underline{81.85}_{\pm 0.30}$& $\underline{96.95}_{\pm 0.30}$ \\
			               & PUDF            & $\mathbf{91.78}_{\pm 0.14}^{*}$ & $\mathbf{89.39}_{\pm 0.11}^{*}$ & $\mathbf{94.61}_{\pm 0.44}$     & $\mathbf{92.81}_{\pm 0.31}$ & $\mathbf{81.90}_{\pm 0.89}$ & $\mathbf{97.07}_{\pm 0.52}$ \\
			\cmidrule{1-8}
			Qwen2.5-7B     & Baseline        & $89.84_{\pm 0.93}$              & $\underline{87.61}_{\pm 0.44}$  & $93.61_{\pm 1.00}$              & $\underline{92.38}_{\pm 0.44}$ & $79.81_{\pm 0.41}$       & $95.66_{\pm 0.56}$          \\
			               & d\_SL-L         & $90.03_{\pm 0.42}$              & $84.56_{\pm 1.83}$              & $93.88_{\pm 0.20}$              & $92.06_{\pm 0.36}$          & $\mathbf{83.99}_{\pm 0.30}$ & $\underline{96.40}_{\pm 0.35}$ \\
			               & d\_SL-R         & $89.35_{\pm 0.55}$              & $85.23_{\pm 1.44}$              & $\underline{94.04}_{\pm 0.22}$  & $91.90_{\pm 0.47}$          & $80.63_{\pm 1.17}$          & $95.40_{\pm 0.41}$           \\
			               & d\_WR-L         & $90.09_{\pm 0.69}$              & $84.30_{\pm 1.62}$              & $93.32_{\pm 0.39}$              & $91.98_{\pm 0.43}$          & $82.71_{\pm 0.42}$          & $95.11_{\pm 0.17}$          \\
			               & d\_WR-R         & $89.95_{\pm 0.62}$              & $84.31_{\pm 1.58}$              & $\mathbf{94.49}_{\pm 0.41}^{*}$ & $91.84_{\pm 0.35}$          & $83.05_{\pm 0.29}$          & $95.00_{\pm 0.16}$             \\
			               & SPL             & $\underline{90.35}_{\pm 0.34}$  & $86.42_{\pm 0.77}$              & $92.91_{\pm 0.44}$              & $92.13_{\pm 0.28}$          & $76.54_{\pm 3.20}$          & $95.44_{\pm 0.22}$          \\
			               & TT              & $89.20_{\pm 0.43}$              & $85.00_{\pm 0.65}$              & $93.98_{\pm 0.25}$              & $91.65_{\pm 0.35}$          & $80.91_{\pm 0.59}$          & $95.36_{\pm 1.33}$          \\
			               & RL              & $89.97_{\pm 0.44}$              & $87.22_{\pm 0.56}$              & $93.28_{\pm 1.03}$              & $91.10_{\pm 0.92}$          & $82.53_{\pm 0.31}$          & $95.54_{\pm 0.75}$          \\
			               & PUDF            & $\mathbf{90.52}_{\pm 0.64}$     & $\mathbf{87.90}_{\pm 0.33}$     & $93.86_{\pm 0.81}$              & $\mathbf{92.70}_{\pm 0.31}$ & $\underline{83.81}_{\pm 0.25}$& $\mathbf{96.64}_{\pm 0.54}$ \\
			\bottomrule
		\end{tabular}

    {\raggedright \footnotesize $^*$Indicates that the difference between the \textbf{best accuracy} and \underline{second-best accuracy} for a dataset-model experiment is significant (Welch's single-tailed t-test, $p < 0.05$).}
	\end{table}

\end{landscape}

\subsection{Additional Qualitative Analysis on IRT-AC}
\label{ssec:easyHardExamples}

Here we replicate the qualitative analysis discussed in Section~\ref{ssec:Quali} for the other datasets in GLUE. These examples provide further insights into the IRT-AC on AG News and GLUE benchmark by presenting the top 5 hardest and easiest sentences for each dataset, along with their respective labels and difficulty scores.

\begin{table}[!bpht]
\centering
\small
\caption{Top 5 hardest and easiest examples in AG News.}
\label{tab:agnews_hardest_easiest_subtables}

\begin{tabular}{p{9cm} p{1.75cm} r} 
    \multicolumn{3}{c}{\textbf{(a) Top 5 Hardest Examples}} \\
    \toprule
     \textbf{Text} & \textbf{Category} & \textbf{Difficulty} \\
    \midrule
     \#1 Ben Roethlisberger guided the Steelers to four touchdowns following uncharacteristic New England turnovers as Pittsburgh ended the Patriots' two long win streaks with a remarkably easy 34-20 victory Sunday. & Entertainment & 22.4401 \\
    \addlinespace 
     \#2 Something big is happening in Cairo. It isn't just the release of Azzam Azzam from wrongful imprisonment as an alleged spy, or nice words from President Hosni Mubarak about Prime Minister Ariel Sharon, or the return of the Egyptian ambassador. & Entertainment & 22.3745 \\
    \addlinespace
     \#3 Chennai: Internet major Yahoo Inc will own a 25-member high-skilled software professional team in Chennai, after it acquired US-based Stata Labs. & Top Stories & 22.3580 \\
    \addlinespace
     \#4 FOR Wall Street types reluctantly dragging themselves back from the Hamptons, slightly cheaper crude oil prices gave them something to cheer about this week and an excuse to buy some oversold stocks. & Sci/Tech & 22.3431 \\
    \addlinespace
     \#5 Pakistani President Pervez Musharraf is to hold talks with President Bush in New York on the war against terror. & Europe & 22.3350 \\
    \bottomrule
\end{tabular}

\vspace{1.5em} 

\begin{tabular}{p{9cm} p{1.75cm} r} 
    \multicolumn{3}{c}{\textbf{(b) Top 5 Easiest Examples}} \\
    \toprule
    \textbf{Text} & \textbf{Category} & \textbf{Difficulty} \\
    \midrule
     \#1 AFP - Opponents of the Lao government may be plotting multiple bomb attacks in Vientiane and other areas of Laos timed to coincide with a meeting of Southeast Asian leaders the country is hosting next month, the United States said. & World & -6.7284 \\
    \addlinespace
     \#2 Picture dated 1977 from the German criminal office Bundeskriminalamt shows German Red Army Faction activist Brigitte Mohnhaupt. Comment & Europe & -6.6625 \\
    \addlinespace
     \#3 AP - The Bush administration has no plans to seek an "Iraq-style" U.N. Security Council resolution on Iran if it succeeds in efforts to have the council address that country's nuclear activities. & World & -6.6365 \\
    \addlinespace
     \#4 Gunmen seized control of a school in northwestern Pakistan, taking more than 200 schoolchildren hostage before surrendering to local elders. & World & -6.6194 \\
    \addlinespace
     \#5 DETROIT - Democratic presidential candidate John Kerry attacked the Bush administration as the "excuse presidency" on Wednesday, charging that the president stood by while jobs disappeared and the middle class lost ground. "This president has created more excuses than jobs... & World & -6.6161 \\
    \bottomrule
\end{tabular}
\end{table}

\begin{table}[!bpht]
\centering
\small
  \caption{Top 5 hardest and easiest examples in MNLI.}
  \label{tab:mnli_hardest_easiest_subtables} 
  \begin{tabular}{p{9cm} p{1.75cm} r} 

  \multicolumn{3}{c}{\textbf{(a) Top 5 Hardest Examples}} \\
  \toprule
  \textbf{Text} & \textbf{Label} & \textbf{Difficulty} \\
  \midrule

  \#1a If practical, T\&A data must be approved at the end of the last day of the pay period or later. & 1 & 13.35 \\
  \#1b T\&A data can be approved at the end of the last day &  &  \\
  \#2a Are the evaluation questions stated clearly and—? & 1 & 13.30 \\
  \#2b The evaluation questions are not stated clearly. &  &  \\
  \#3a Cook will  ". & 1 & 1.370 \\
  \#3b Will will be cooked. &  &  \\
  \#4a An audio-visual show about the tower is screened on the first platform; there are restaurants on the first and second, and a bar on the third. & 1 & 13.17 \\
  \#4b There are restaurants on the first and second platforms. &  &  \\
  \#5a yeah it’s a whole whole different culture um it’s weird down there because. & 1 & 13.09 \\
  \#5b Places that have different cultures are weird. &  &  \\
  \bottomrule
\end{tabular}

\begin{tabular}{p{9cm} p{1.75cm} r} 

  \multicolumn{3}{c}{\textbf{(b) Top 5 Easiest Examples}} \\
  \toprule
  \textbf{Text} & \textbf{Label} & \textbf{Difficulty} \\
  \midrule

  \#1a There are delightful wooden figures made as servants for the dead. & 0 & –11.49 \\
  \#1b The wooden figures are supposed to be servants for the dead. &  &  \\
  \#2a The merchandise processing fee is associated with the cost of the Customs Service’s operations. & 0 & –11.47 \\
  \#2b The merchandise processing fee is associated with the cost of the Customs Service’s operations &  &  \\
  \#3a The cakes here are also excellent. & 2 & –11.41 \\
  \#3b The cakes here are equally great. &  &  \\
  \#4a In the example above, Yellowstone National Park would be reported under a category, such as National Parks, as one of the total number of heritage assets under the auspices of the Department of the Interior; it also would be reported by the number of acres that it occupies under the stewardship land category for the Department. & 2 & –11.39 \\
  \#4b Yellowstone National Park would be reported as a heritage asset. &  &  \\
  \#5a Most pubs also serve food—they are good places to have lunch and some have dining rooms. & 2 & –11.38 \\
  \#5b Some pubs have dining rooms, and they are good places to have lunch in. &  &  \\
  \bottomrule
\end{tabular}
\end{table}

\begin{table}
\small
\centering
  \caption{Top 5 hardest and easiest examples in MRPC.}
  \label{tab:mrpc_hardest_easiest_subtables} 
\begin{tabular}{p{9cm} c r}
  \multicolumn{3}{c}{\textbf{(a) Top 5 Hardest Examples}} \\
  \toprule
  \textbf{Text} & \textbf{Label} & \textbf{Difficulty} \\
  \midrule

  \#1a “In relation to the second paper, one part of that should have been sourced to a reference work.” & 0 & 10.69 \\
  \#1b He went on: “One part of that should have been sourced to a reference work.” &  &  \\
  \#2a Seven Air Force Academy cadets face punishment for allegedly drinking with two high school girls. & 0 & 10.67 \\
  \#2b Seven Air Force Academy cadets face punishment for drinking with two high school girls in the latest incident to embarrass the academy. &  &  \\
  \#3a Defense Secretary Donald Rumsfeld is awaiting recommendations from his commanders. & 0 & 10.53 \\
  \#3b Rumsfeld is awaiting recommendations from his commanders about troop needs in Iraq. &  &  \\
  \#4a Last year, he was forced to repay \$3,000 in bar tabs that he and his staff incurred while he was labour minister but had originally billed to taxpayers. & 0 & 10.53 \\
  \#4b Last year, Stockwell was forced to repay \$3,000 in bar tabs that he and his staff rang up while he was labour minister. &  &  \\
  \#5a Most of those killed were labourers from Jharkhand and Nepal who were working at Rohtang tunnel. & 0 & 10.42 \\
  \#5b The majority of the dead were labourers from Jharkhand and Nepal. &  &  \\
  \bottomrule
\end{tabular}
\begin{tabular}{p{9cm} c r}
  \multicolumn{3}{c}{\textbf{(b) Top 5 Easiest Examples}} \\
  \toprule
  \textbf{Text} & \textbf{Label} & \textbf{Difficulty} \\
  \midrule

  \#1a She was surrounded by about 50 women who regret having abortions. & 1 & –8.87 \\
  \#1b She was surrounded by about 50 women who have had abortions but now regret doing so. &  &  \\
  \#2a A race observer sits in the passenger seat of the vehicle following the contestant car to record broken rules and track of the car’s time. & 1 & –8.79 \\
  \#2b A race observer sits in the passenger seat of the follow vehicle to record any broken rules and also keep track of the car’s time. &  &  \\
  \#3a She met Lady Mary at her Double Bay home yesterday to thank her for the donation. & 1 & –8.64 \\
  \#3b She met Lady Mary for the first time at her Double Bay home in Sydney yesterday to thank her in person for the donation. &  &  \\
  \#4a The national denomination of the Episcopal Church, with 2.3 million members, is the U.S. branch of the 77 million-member Anglican Communion. & 1 & –8.57 \\
  \#4b The Episcopal Church, with 2.3 million members, is the American branch of the worldwide Anglican Communion, which has 77 million adherents. &  &  \\
  \#5a It held elections in 1990, but refused to recognize the results after Suu Kyi’s party won. & 0 & –8.56 \\
  \#5b It held elections in 1990, but refused to recognise the results when Miss Suu Kyi’s National League for Democracy won by a landslide. &  &  \\
  \bottomrule
\end{tabular}
\end{table}

\begin{table}
\small
\centering
  \caption{Top 5 hardest and easiest examples in QNLI.}
  \label{tab:qnli_hardest_easiest_subtables} 
  \begin{tabular}{p{9cm} c r}

  \multicolumn{3}{c}{\textbf{(a) Top 5 Hardest Examples}} \\
  \toprule
  \textbf{Text} & \textbf{Label} & \textbf{Difficulty} \\
  \midrule

  \#1a Europe did not feel the need to posses territory in Africa until? & 1 & 13.26 \\
  \#1b European countries established a few coastal colonies in Africa by the mid-nineteenth century, which included Cape Colony (Great Britain), Angola (Portugal), and Algeria (France), but until the late nineteenth century Europe largely traded with free African states without feeling the need for territorial possession. &  &  \\
  \#2a What are the windows of 1990s and later pubs often made of? & 1 & 13.20 \\
  \#2b Traditionally the windows of town pubs were of smoked or frosted glass to obscure the clientele from the street but from the 1990s onwards &  &  \\
  \#3a How many nominations has American Idol received for Outstanding Reality Competition Program? & 1 & 13.19 \\
  \#3b American Idol was nominated for the Emmy's Outstanding Reality Competition Program for nine years but never won. &  &  \\
  \#4a How many species have been downgraded from endangered to threatened status? & 1 & 13.17 \\
  \#4b As of September 2012, fifty-six species have been delisted; twenty-eight due to recovery, ten due to extinction (seven of which are believed to have been extinct prior to being listed), ten due to changes in taxonomic classification practices, six due to discovery of new populations, one due to an error in the listing rule, and one due to an amendment to the Endangered Species Act specifically requiring the species delisting. &  &  \\
  \#5a What causes osmotic diarrhea? & 1 & 13.17 \\
  \#5b In healthy individuals, too much magnesium or vitamin C or undigested lactose can produce osmotic diarrhea and distention of the bowel. &  &  \\
  \bottomrule
\end{tabular}
  \begin{tabular}{p{9cm} c r}

  \multicolumn{3}{c}{\textbf{(b) Top 5 Easiest Examples}} \\
  \toprule
  \textbf{Text} & \textbf{Label} & \textbf{Difficulty} \\
  \midrule

  \#1a At approximately what time did paramedics receive the call about Kanye West's mother, Donda? & 0 & -11.80 \\
  \#1b On November 10, 2007, at approximately 7:35 pm, paramedics responding to an emergency call transported West's mother, Donda West, to the nearby Centinela Freeman Hospital in Marina del Rey, California. &  &  \\
  \#2a A Soviet double agent working for The Times in Spain was a war correspondent during what war in the late 1930s? & 0 & -11.74 \\
  \#2b Kim Philby, a Soviet double agent, was a correspondent for the newspaper in Spain during the Spanish Civil War of the late 1930s. &  &  \\
  \#3a Which line was reopened in 2009? & 0 & -11.70 \\
  \#3b In July 2009 the Glounthaune to Midleton line was reopened, with new stations at Carrigtwohill and Midleton (with future stations planned for Kilbarry, Monard, Carrigtwohill West and Blarney). &  &  \\
  \#4a How much ad revenue goes to the original uploader of the YouTube video if they're in the Partner Program? & 0 & -11.69 \\
  \#4b YouTube typically takes 45 percent of the advertising revenue from videos in the Partner Program, with 55 percent going to the uploader. &  &  \\
  \#5a How does heartwood formation occur due to its being genetically programmed? & 0 & -11.69 \\
  \#5b Heartwood formation occurs spontaneously (it is a genetically programmed process). &  &  \\
  \bottomrule
\end{tabular}
\end{table}

\begin{table}
\small
\centering
  \caption{Top 5 hardest and easiest examples in QQP.}%
  \label{tab:qqp_hardest_subtables} 

\begin{tabular}{p{9cm} c r}
  \multicolumn{3}{c}{\textbf{(a) Top 5 Hardest Examples}} \\
  \toprule
  \textbf{Text} & \textbf{Label} & \textbf{Diff.} \\
  \midrule

  \#1a What licenses are required to sell agricultural products (viz fertilizers, seeds, pesticides) online in India? & 1 & 13.22 \\
  \#1b I plan to sell agricultural products such as tulsi seeds and spices through eBay India? Do I need special licenses? &  &  \\ 
  \#2a Is Zee news a BJP owned channel? & 1 & 13.22 \\
  \#2b What is the problem with Zee News channel? &  &  \\ 
  \#3a From my boyfriend’s Facebook account, an old girlfriend is not on his friends list anymore, but still shows in messenger as an active friend. Why? & 1 & 13.08 \\
  \#3b She shows up first on his list of people that have Facebook Messenger but they aren’t friends and he says they don’t talk. Why is that? &  &  \\ 
  \#4a What are some of the best ways to market to schools? & 0 & 13.06 \\
  \#4b How do I market in schools? &  &  \\ 
  \#5a List the references and study materials for IAS civil service exams? How is this formula determined? & 1 & 13.02 \\
  \#5b Whose study materials is best for philosophy optional in civil service exam—Vision IAS or Mitra Sir? &  &  \\
  \bottomrule
\end{tabular}

\begin{tabular}{p{9cm} c r}


  \multicolumn{3}{c}{\textbf{(b) Top 5 Easiest Examples}} \\
  \toprule
  \textbf{Text} & \textbf{Label} & \textbf{Diff.} \\
  \midrule

  \#1a How do I attach ccavenue payment gateway with button on my page in WordPress? & 0 & –11.49 \\
  \#1b Is it possible to add a link button to a page builder in WordPress? &  &  \\ 
  \#2a Is there any company hiring Indians (MBA + 3 years work experience in finance) for their US/UK location? & 0 & –11.48 \\
  \#2b I am a UK citizen working for a US company on a J1 visa; can I work from UK remotely for 2–4 weeks per year? &  &  \\ 
  \#3a Two moving observers cycling around the equator in an opposite direction at 0.5C. What will they measure when exchanging light signals in each orbit? & 0 & –11.41 \\
  \#3b Find the answer—two trains 120 m and 80 m in length are running in opposite directions with velocities 42 km/h and 30 km/h; at what time will they completely cross each other? &  &  \\ 
  \#4a Do students get a seat under General without fee waiver category if they are allocated seats under General with FeeWaiver category in UPSEE counseling? & 0 & –11.41 \\
  \#4b Is a score of 533 in NEET Phase 1 good enough to get an MBBS seat in a government college in Delhi under state quota for General category? &  &  \\ 
  \#5a I am Sneha from Delhi. Due to family issues, I was not able to study after 7th class. Can I travel or settle abroad? & 0 & –11.35 \\
  \#5b I have a traveling phobia and I’m very attached to my family. How can I get over it so I can go and study abroad next year? &  &  \\
  \bottomrule
\end{tabular}
\end{table}

\begin{table}[!bpht]
\centering
\small
\caption{Top 5 hardest and easiest examples in SST-2.}
\label{tab:sst2_hardest_easiest_subtables}

\begin{tabular}{p{9cm} c r}
    \multicolumn{3}{c}{\textbf{(a) Top 5 Hardest Examples}} \\
    \toprule
    \textbf{Text} & \textbf{Label} & \textbf{Difficulty} \\
    \midrule
    \#1 a , incoherence and sub-sophomoric & Pos. & 11.01 \\
    \#2 into forced fuzziness & Pos. & 10.88 \\
    \#3 wondering why lee ’s character didn’t just go to a bank manager and save everyone the misery & Pos. & 10.77 \\
    \#4 sometimes descends into sub-tarantino cuteness & Pos. & 10.59 \\
    \#5 eerily accurate depiction of depression & Neg. & 10.43 \\
    \bottomrule
\end{tabular}

\vspace{1.5em}

\begin{tabular}{p{9cm} c r}
    \multicolumn{3}{c}{\textbf{(b) Top 5 Easiest Examples}} \\
    \toprule
    \textbf{Text} & \textbf{Label} & \textbf{Difficulty} \\
    \midrule
    \#1 useless & Neg. & -10.30 \\
    \#2 ’m giving it thumbs down due to the endlessly repetitive scenes of embarrassment & Neg. & -10.04 \\
    \#3 the rest is just an overexposed waste of film & Neg. & -9.96 \\
    \#4 the characters are more deeply thought through than in most ` right-thinking ’ films & Pos. & -9.92 \\
    \#5 this orange has some juice , but it ’s far from fresh-squeezed & Neg. & -9.87 \\
    \bottomrule
\end{tabular}
\end{table}

\begin{table}
\footnotesize
\centering
  \caption{Top 5 hardest and easiest examples in RTE.} \label{tab:rte_hardest_easiest_subtables} 

\begin{tabular}{p{13cm}}
  \toprule
  \textbf{(a) Top 5 Hardest Examples} \\
  \midrule

  
  \#1a IT must rate as the literary snub of the 20th century. T S Eliot, one of Britain’s greatest poets, rejected George Orwell’s Animal Farm for publication on the grounds of its unconvincing Trotskyite politics. Eliot, a former director of Faber and Faber, wrote his rejection in a highly critical letter in 1944, one of many private papers made available for the first time by his widow Valerie for a BBC documentary. When Orwell submitted his novel, an allegory on Stalin’s dictatorship, Eliot praised its “good writing” and “fundamental integrity.” \\
  \#1b T.S. Eliot wrote “Animal Farm.”~~~Label: 1  Difficulty: 10.82 \\
  \#2a When the last member of the Hasmonean Dynasty died in 37 B.C., Rome made Herod king of Judah. With Roman backing, Herod (37–34 B.C.) ruled on both sides of the Jordan River. After his death the Jewish kingdom was divided among his heirs and gradually absorbed into the Roman Empire. \\
  \#2b The Hasmonean Dynasty rules Jordan.~~~Label: 1  Difficulty: 10.61 \\
  \#3a Over 100 people died in a massive stampede at the Chamunda Devi temple in Jodhpur, in the western state of Rajasthan, India. One hundred sixty-eight people were killed, with just as many injured. A temple wall collapsed, causing panic among tens of thousands of gathered Hindu worshippers who then began to run for safety, causing the stampede. Several are still believed to be trapped under the rubble.  \\
  \#3b 100 died at Chamunda Devi Temple.~~~Label: 1  Difficulty: 10.58 \\
  \#4a Diets that provide recommended levels of magnesium are beneficial for bone health, but further investigation on the role of magnesium in bone metabolism and osteoporosis is needed.  \\
  \#4b Dietary intake of magnesium prevents osteoporosis.~~~Label: 1  Difficulty: 10.56 \\
  \#5a Gastrointestinal bleeding can happen as an adverse effect of non-steroidal anti-inflammatory drugs such as aspirin or ibuprofen. \\
  \#5b Aspirin prevents gastrointestinal bleeding.~~~Label: 1  Difficulty: 10.55 \\

  \midrule
\end{tabular}
\begin{tabular}{p{13cm}}

  \toprule
  \textbf{(b) Top 5 Easiest Examples} \\
  \midrule

  \#1a Nicholas Cage’s wife, Alice Kim Cage, gave birth Monday to a boy, Kal-el Coppola Cage, in New York City, said Cage’s Los Angeles–based publicist, Annett Wolf. No other details were available.  \\
  \#1b Nicholas Cage is married to Alice Kim Cage.~~~Label: 0  Difficulty: –10.83 \\
  \#2a This document declares the “irrevocable determination” of Edward VIII to abdicate. By signing this document on December 10, 1936, he gave up his right to the British throne. \\
  \#2b King Edward VIII abdicated on the 10th of December, 1936.~~~Label: 0  Difficulty: –10.46 \\
  \#3a November 9, 1989 — the day the Berlin Wall fell and the world changed forever. Not even the most astute saw it coming. As Hungary’s foreign minister in the late summer of 1989, Gyula Horn gave the order to let visiting East Germans use his country to do a 400-mile end run around the Berlin Wall, a move now seen as the beginning of the end for hard-line communism in Europe.  \\
  \#3b The Berlin Wall was torn down in 1989.~~~Label: 0  Difficulty: –10.43 \\
  \#4a The floods were exceptional since they affected an extensive area across Europe from the UK to Spain and as far east as the Black Sea coast. Economic losses amounted to EUR 9.2 bn in Germany, EUR 2.9 bn in Austria, and EUR 2.3 bn in the Czech Republic. Total economic damage exceeds EUR 15 bn.  \\
  \#4b Flooding in Europe causes major economic losses.~~~Label: 0  Difficulty: –10.38 \\
  \#5a Bush’s second term as President of the United States, which began on January 20, 2005, expired with the swearing-in of the 44th President of the United States, Barack Obama, at noon EST (UTC–5), under the provisions of the Twentieth Amendment to the United States Constitution. Bush performed his final official act this morning, welcoming Barack Obama and Michelle to the White House for coffee before the swearing-in, shortly before 10 am EST, and then accompanied them there by motorcade to attend the ceremony. Last week, Bush had made his farewells to the nation in a televised address, saying that the inauguration turns a page in race relations. “Obama’s story — his black father was from Kenya, his white mother from Kansas — represents the enduring promise of our land,” said Bush.  \\
  \#5b Barack Obama is the 44th President of the United States.~~~Label: 0  Difficulty: –10.19 \\

  \bottomrule
\end{tabular}
\end{table}

\end{document}